\documentclass{article}

\usepackage{arxiv}

\usepackage[utf8]{inputenc} 
\usepackage[T1]{fontenc}    
\usepackage{hyperref}       
\usepackage{url}            
\usepackage{booktabs}       
\usepackage{amsfonts}       
\usepackage{nicefrac}       
\usepackage{microtype}      
\usepackage{lipsum}
\usepackage{tikz}
\usepackage{graphicx}  
\usepackage[caption=false]{subfig}
\usepackage{float}
\usepackage{algorithm}
\usepackage{algorithmic}

\usepackage{siunitx}
\usepackage{booktabs,colortbl, array}
\usepackage{pgfplotstable}
\pgfplotsset{compat=1.8}

\definecolor{rulecolor}{RGB}{0,71,171}
\definecolor{tableheadcolor}{gray}{0.92}
\newcommand{\topline}{ %
        \arrayrulecolor{rulecolor}\specialrule{0.1em}{\abovetopsep}{0pt}%
        \arrayrulecolor{tableheadcolor}\specialrule{\belowrulesep}{0pt}{0pt}%
        \arrayrulecolor{black}
        }
        
\newcommand{\midtopline}{ %
        \arrayrulecolor{tableheadcolor}\specialrule{\aboverulesep}{0pt}{0pt}%
        \arrayrulecolor{rulecolor}\specialrule{\lightrulewidth}{0pt}{0pt}%
        \arrayrulecolor{white}\specialrule{\belowrulesep}{0pt}{0pt}%
        \arrayrulecolor{rulecolor}}
\newcommand{\bottomline}{ %
        \arrayrulecolor{white}\specialrule{\aboverulesep}{0pt}{0pt}%
        \arrayrulecolor{rulecolor} %
        \specialrule{\heavyrulewidth}{0pt}{\belowbottomsep}}%

\newcolumntype{C}{>{\centering\arraybackslash}m{4.0em}}
\newcolumntype{D}{>{\centering\arraybackslash}m{3.0em}}
\newcolumntype{E}{>{\centering\arraybackslash}m{4.5em}}
\newcolumntype{F}{>{\centering\arraybackslash}m{7.5em}}

\pgfplotstableset{normal/.style ={%
        header=true,
        string type,
        font=\small,
        column type=c,
        every odd row/.style={
            before row=
        },
        every head row/.style={
            before row={\topline\rowcolor{tableheadcolor}},
            after row={\midtopline}
        },
        every last row/.style={
            after row=\bottomline
        },
        col sep=&,
        row sep=\\
    }
}

\title{Engineered Self-Organization for Resilient Robot Self-Assembly with Minimal Surprise}

\author{
  Tanja Katharina Kaiser\\
  Institute of Computer Engineering\\
  University of L\"{u}beck \\
  Germany \\
  \texttt{kaiser@iti.uni-luebeck.de} \\
   \And
  Heiko Hamann\\
  Institute of Computer Engineering\\
  University of L\"{u}beck \\
  Germany \\
  \texttt{hamann@iti.uni-luebeck.de} \\
}

\begin{document}
\maketitle

\begin{abstract}
In collective robotic systems, the automatic generation of controllers for complex tasks is still a challenging problem. 
Open-ended evolution of complex robot behaviors can be a possible solution whereby an intrinsic driver for pattern formation and self-organization may prove to be important. 
We implement such a driver in collective robot systems by evolving prediction networks as world models in pair with action-selection networks. 
Fitness is given for good predictions which causes a bias towards easily predictable environments and behaviors in the form of emergent patterns, that is, environments of minimal surprise. There is no task-dependent bias or any other explicit predetermination for the different qualities of the emerging patterns.
A~careful configuration of actions, sensor models, and the environment is required to stimulate the emergence of complex behaviors.
We study self-assembly to increase the scenario's complexity for our minimal surprise approach and, at the same time, limit the complexity of our simulations to a grid world to manage the feasibility of this approach. 
We investigate the impact of different swarm densities and the shape of the environment on the emergent patterns. 
Furthermore, we study how evolution can be biased towards the emergence of desired patterns. We analyze the resilience of the resulting self-assembly behaviors by causing damages to the assembled pattern and observe the self-organized reassembly of the structure.
In summary, we evolved swarm behaviors for resilient self-assembly and successfully engineered self-organization in simulation. 
In future work, we plan to transfer our approach to a swarm of real robots. 
\end{abstract}


\section{Introduction}

In swarm robotics self-organizing robots collaborate to complete tasks~\cite{hamann2018,brambilla13}. 
As each robot acts autonomously and relies on local information only, the robot swarm forms a decentralized system. 
The manual design of decentralized systems is known to be difficult~\cite{hamann2018, trianni08}.
Methods of machine learning, such as multi-agent learning~\cite{busoniu2008comprehensive,panait05}, seem a reasonable option.
Despite of recent improvements in machine learning~~\cite{silver18,gupta2017cooperative}, the automatic generation of robot controllers for complex tasks and especially for collective robot systems remains challenging~\cite{doncieux2015}.
An additional challenge is how the decentralized system may be able to improve its performance at runtime, stay adaptive to changes in the environment, and increase its own capabilities and complexity.
A~possible solution can be the implementation of open-ended evolution~\cite{taylor2016open,ruiz2008enabling} of robot behaviors that may generate more and more complex, task-independent, and interesting behaviors by itself~\cite{bongard13, lehman08}. 
For example, we can evolve swarm behaviors by using evolutionary swarm robotics~\cite{trianni08} that is the application of evolutionary robotics to swarm robotics.
However, especially the design of fitness functions remains challenging in evolutionary robotics~\cite{nelson_2009_fitness}.
While a priori domain-specific knowledge influences the outcome and performance positively, it foils the idea of evolution as a black-box optimizer at the same time~\cite{doncieux14,divband15}. 
Furthermore, the fitness function has to be designed carefully such that the evolutionary algorithm does not converge prematurely~\cite{eiben_2003_introduction}. 

The creation of an intrinsic driver for pattern formation and self-organization may be a potential strategy to overcome these challenges. 
A built-in implicit motivation for forming patterns instead of task-specific rewards may have more potential to create complex robot behaviors. 
We are loosely inspired by Friston's work on a free-energy principle for natural brains in search for such a driver~\cite{friston10}. 
Friston assumes that vertebrates have an innate driver to minimize surprise as brains continually try to predict their environment. 
Organisms may have a tendency to stay in safe and boring environments meaning that the organism's brain can easily predict them and thus, an evolutionary advantage arises.  
We roughly follow this concept by making the innate driver explicit as selective pressure on making good predictions in our work~\cite{borkowski17,hamann14d,zahadat15a}. 
We evolve pairs of artificial neural networks (ANN) whereby one serves as a typical robot controller.
The other ANN serves as a world model which predicts the next sensor values based on the currently observed values. 
It represents a model of the robot's environment. 
Our previous studies were based on 1D and 2D continuous simulation environments and we observed only four basic swarm behaviors.
Self-assembly is an example for more complex collective behaviors that are found in nature. 
The system components
organize themselves autonomously and only through local interactions into patterns or structures~\cite{Whitesides2418}.  
This serves as inspiration for swarm robotics research. 
Self-assembly behaviors were already implemented on robot swarms of up to one thousand Kilobots~\cite{Rubenstein795}.
Here, we study self-assembly in order to move to a next level of complexity.
With minimal surprise, we propose an exploratory approach to robot self-assembly. 
We reward the robots' prediction accuracy and do not directly influence which structures are formed. 
Self-assembly behaviors emerge during the evolutionary process as a by-product of the evolutionary dynamics.
To ensure the feasibility of our study, we limit ourselves to a 2D grid world which leads to a simplification of sensing and equidistant positioning of robots.
Our objective is to exploit this paper's results and transfer them to robots, although this will be challenging.

This paper extends our previous work~\cite{kaiser18} by a comparison of different sensor models for use in our self-assembly scenario as well as by a study of the resilience of the emergent self-assembly behaviors. 
We introduce our minimal surprise approach in Sec.~\ref{sec:Approach} and present an extended discussion of related work in Sec.~\ref{sec:relatedWork}. 
Sec.~\ref{sec:SelfAssembly} presents the setup for evolving self-assembly behaviors with minimal surprise. 
We introduce metrics for structure classification in Sec.~\ref{sec:classification} that replace the qualitative analysis used in our previous work~\cite{kaiser18} by a quantitative analysis of the resulting structures. 
We describe our results in Sec.~\ref{sec:results}. 
In Sec.~\ref{sec:sensormodel}, we present the previously missing justification for the choice of our sensor model by comparing three candidate sensor models. 
As in our previous study~\cite{kaiser18}, we show evolved self-assembly behaviors in different swarm densities in Sec.~\ref{sec:Adaptation}. 
In Sec.~\ref{sec:SOSA}, we engineer self-organized self-assembly to guide evolution towards desired structures while keeping the implicit motivation. 
We use the new metrics for structure classification presented in Sec.~\ref{sec:classification}. 
In addition, we present two novel studies focused on the system's robustness.
In Sec.~\ref{sec:Repair}, we analyze the resilience of the evolved behaviors by damaging assembled parts.
In Sec.~\ref{sec:noise}, we study sensor noise which is also a first step towards real-world scenarios.  
We discuss our results and outline future work in Sec.~\ref{sec:Discussion}.

\subsection{The Minimal Surprise Approach}
\label{sec:Approach}

Using minimal surprise, we aim for the emergence of swarm behaviors ranging from simple behaviors like aggregation to more complex behaviors like self-assembly. 
Our robot swarms of size~$N$ lived on simulated 1D and 2D torus environments so far~\cite{borkowski17, hamann14d}, but we aim for the extension to real world scenarios in future work. 

\tikzset{%
  every neuron/.style={
    circle,
    draw,
    minimum size=0.7cm
  },
  neuron missing/.style={
    draw=none, 
    scale=2,
    text height=0.333cm,
    execute at begin node=\color{black}$\vdots$
  },
}

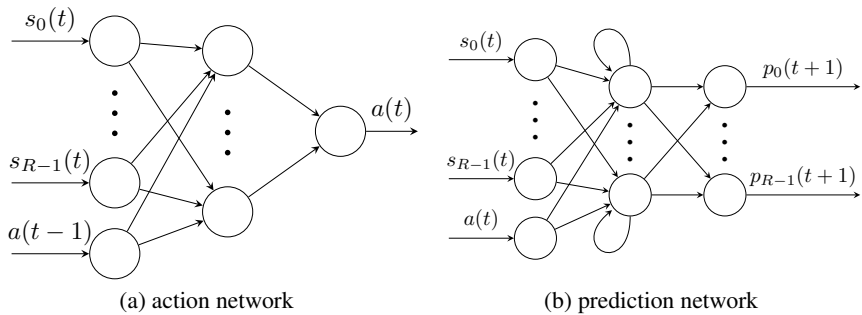
\begin{figure}[tph]
    \centering
    \subfloat[action network]{
\resizebox {0.35\textwidth} {!} {
\begin{tikzpicture}[x=0.9cm, y=0.9cm, >=stealth,]
\foreach \m/\l [count=\y] in {1,missing,2,4}
  \node [every neuron/.try, neuron \m/.try] (input-\m) at (0,2 -\y*1.1) {};
\foreach \m [count=\y] in {1,missing,2}
  \node [every neuron/.try, neuron \m/.try ] (hidden-\m) at (1.75,2-\y*1.25) {};
\foreach \m [count=\y] in {1}
  \node [every neuron/.try, neuron \m/.try ] (output-\m) at (3.5,0.5-\y) {};
\draw [<-] (input-1) -- ++(-1.6,0)
    node [above, midway] {$s_{0}(t)$};
\draw [<-] (input-2) -- ++(-1.6,0)
    node [above, midway] {$s_{R-1}(t)$};
\draw [<-] (input-4) -- ++(-1.6,0)
    node [above, midway] {$a(t-1)$};
\foreach \l [count=\i] in {1}
  \draw [->] (output-\i) -- ++(1.2,0)
    node [above, midway] {$a(t)$};
\foreach \i in {1,2,4}
  \foreach \j in {1,...,2}
    \draw [->] (input-\i) -- (hidden-\j);
\foreach \i in {1,...,2}
  \foreach \j in {1}
    \draw [->] (hidden-\i) -- (output-\j);
\end{tikzpicture}\label{fig:FIG1a} }}
    \subfloat[prediction network]{
    \resizebox {0.35\textwidth} {!} {
  \begin{tikzpicture}[x=0.9cm, y=0.9cm, >=stealth,]
\foreach \m/\l [count=\y] in {1,missing,2,4}
  \node [every neuron/.try, neuron \m/.try] (input-\m) at (0,2 -\y*1.1) {};
\foreach \m [count=\y] in {1,missing,2}
  \node [every neuron/.try, neuron \m/.try ] (hidden-\m) at (1.75,1.4-\y) {};
\foreach \m [count=\y] in {1,missing,2}
  \node [every neuron/.try, neuron \m/.try ] (output-\m) at (3.5,1.4-\y) {};
\draw [<-] (input-1) -- ++(-1.6,0)
    node [above, midway] {$s_0(t)$};
\draw [<-] (input-2) -- ++(-1.6,0)
    node [above, midway] {$s_{R-1}(t)$};
\draw [<-] (input-4) -- ++(-1.6,0)
    node [above, midway] {$a(t)$};
\foreach \l [count=\i] in {0, R-1}
  \draw [->] (output-\i) -- ++(2.5,0)
    node [above, midway] {$p_{\l}(t+1)$};
\foreach \i in {1,2,4}
  \foreach \j in {1,...,2}
    \draw [->] (input-\i) -- (hidden-\j);
\foreach \i in {1,...,2}
  \foreach \j in {1,...,2}
    \draw [->] (hidden-\i) -- (output-\j);
 \draw[->,shorten >=1pt] (hidden-1) to [out=90,in=140,loop,looseness=8.8] (hidden-1);
 \draw[->,shorten >=1pt] (hidden-2) to [out=270,in=220,loop,looseness=8.8] (hidden-2);
\end{tikzpicture}  }\label{fig:FIG1b}
    }
\caption{ANN pair of each robot. 
$a(t-1)$~is the robot's last action value and $a(t)$ is its next action. 
$s_0(t),\dots,s_{R-1}(t)$ are the R sensor values of the robot at time step~t, ${p_0(t+1),\dots,p_{R-1}(t+1)}$ are its sensor value predictions for time step $t+1$. Adapted by permission from Springer Nature Customer Service Centre GmbH: \cite{kaiser18}, \copyright Springer Nature Switzerland AG 2019.}
\label{fig:FIG1}
\end{figure}

Each swarm member (robot), is equipped with a pair of three-layered ANNs that are evolved jointly. 
A~regular controller is implemented by a feedforward network, that we call action network (see Fig.~\ref{fig:FIG1a}). 
One output of the action network represents the robot's next action, which is either straight motion or rotation. 
There may be additional outputs depending on the experimental setting.
In addition, the prediction network (see Fig.~\ref{fig:FIG1b}) is implemented as a recurrent network and enables robots to predict their sensor values of the next time step.  
The prediction network can be seen as a world model: It predicts each robot's environment by predicting the outputs of the robot's exteroceptive sensors.

The current sensor values of a robot are given as inputs to both networks.
The robot's last action value is given as an additional input to the action network while the next action is given to the prediction network. 
This updated information could be leveraged by the prediction network and may improve its performance.

We apply a simple genetic algorithm~\cite{holland75}.
The genomes consist of two sets of weights: one for the action network and one for the prediction network.
They are randomly generated for the initial population.
We have exclusively homogeneous swarms as all robots of an evaluation have instances of the same ANN pair, that is, they share the same genome. 
Notice that we have two concepts of populations: a population of genomes encoding pairs of ANN for the evolutionary algorithm and a population of robots forming a homogeneous swarm used in the evaluations of a genome. 
We use discrete sensors in all experiments. 
The sensors output a `1' if another robot is detected and a `0' otherwise. 
We reward good predictions, that is, we put selective pressure on the prediction network while the action network is not directly rewarded. 
Thus, the action network receives no direct selective pressure, but is subject to genetic drift. 
As action networks and prediction networks are evaluated as pairs, high fitness values can only be reached if the selected actions of the action network lead to behaviors that fit to the sensor value predictions of the prediction network. 
ANN pairs which receive high fitness values have a higher likelihood to survive in the evolutionary process and consequently, the action network receives indirect selective pressure. 

The fitness function rewards correct predictions with a fitness value normalized to a maximum of~$1$.
Fitness measures prediction accuracy.
We define the fitness over an evaluation period of $T$~time steps as 
\begin{equation}
F = \frac{1}{NTR} \sum_{t=0}^{T-1} \sum_{n=0}^{N-1} \sum_{r=0}^{R-1} 1 -  | p_{n}^{r}(t) - s_{n}^{r}(t) |,
\label{equ:fitness}
\end{equation}
where $N$ is the swarm size, $R$~is the number of sensors per robot, $p_{n}^{r}(t)$ is the prediction for sensor~$r$ of robot~$n$ at time step~$t$, and~$s_{n}^{r}(t)$~is the value of sensor~$r$ of robot~$n$ at time step $t$.

\subsection{Related Work} \label{sec:relatedWork}

We review two areas of research that are related to our experiments on self-assembly with minimal surprise. 
First, we discuss conceptually similar approaches that use pairs of neural networks to generate agent behaviors. 
As we focus on grid-based self-assembly scenarios, we also discuss related applications of cellular automata in the second part of this section.

\subsubsection{Pairs of Neural Networks}

There are several similar approaches to minimal surprise that make use of pairs of neural networks ranging from machine learning to (swarm) robotics applications.

Ha and Schmidhuber \cite{schmidhuber_2018, ha_2018} train controllers using features extracted from a world model.
This world model is implemented by a large recurrent neural network that is combined with a Mixture Density Network~\cite{bishop1994}. 
It is trained by unsupervised learning independently from the controller.
It predicts the future by determining a probability distribution of the next state based on information about its current state and current action.
The controller is implemented as a perceptron and optimized by maximizing a task-specific reward using evolution strategies. 
For example, the agent is rewarded for visiting a maximum of tiles of the track in a minimum of time in the presented car racing scenario. 
The controller can even be trained entirely within hypothetical scenarios or `dreams' generated by the world model.
In that case, a world state is sampled from the probability distribution generated by the world model and is used as input for the next observation of the controller.
In contrast to minimize surprise, Ha and Schmidhuber applied their approach to single agents in OpenAI Gym scenarios.
As Ha and Schmidhuber train world models and controllers separately, different controllers can be trained using the same world model and independently from the actual environment. 
This is only possible as long as single agent scenarios are considered. 
In collective systems, the world model depends not only on the (dynamic) environment but also on the existence and behavior of surrounding agents.
The prediction network necessarily depends on the action network and  they need to be evolved in pairs using minimize surprise. 
Furthermore, the world model used in Ha and Schmidhuber's approach is represented by a large neural network and its training is computational expensive. 

Nolfi et al.~\cite{nolfi_learning_1994,nolfi_learning_2002} combine learning during the lifetime of an individual with evolution across generations.  
The changes of the genome by learning at runtime are not inherited during evolution (cf. Lamarckian inheritance). 
They showed that learning has a beneficial effect on evolution by using different tasks for learning and evolution. 
In a sample setting, a single simulated agent has the evolutionary task to find food in a 2D grid world while its learning task is to predict the sensor values of the next time step, that is the next position of food. 
The reward given during learning in this setting is similar to the reward given in minimal surprise during evolution. 
However, our learning process is driven fully intrinsically while Nolfi et al. rely on task-specific rewards. 

Generative Adversarial Nets (GANs) were first proposed by Goodfellow et al.~\cite{goodfellow_generative_2014}. 
A~generative model captures a data distribution while a discriminator estimates whether a sample is genuine or generated by the model. 
The generator is rewarded for tricking the discriminator to misclassify samples.
This generates an adversarial process to estimate generative models. 
Goodfellow et al. represent both generator and discriminator as multilayer perceptrons and train them with backpropagation.
Radford et al.~\cite{radford2015adversarial}, for example, use deep convolutional generative adversarial networks to learn general image representations in an unsupervised way.  

Turing learning is conceptually similar to GANs~\cite{gross2017generalizing,Li2016}.
This system identification method allows machines to infer behaviors of natural or artificial systems without requiring predefined metrics. 
Two populations are optimized concurrently using coevolutionary algorithms: one population consists of models of the behavior of the system under investigation while the other consists of classifiers. 
The classifiers are rewarded for discriminating data between model and real system while models aim for tricking the classifier to categorize their data as genuine. 
Turing learning was applied to swarm robotics~\cite{Li2016}. 

Neither GANs, Turing learning, nor minimize surprise make use of explicit reward functions. 
In contrary to minimize surprise, GANs and Turing learning aim to mimic an existing system based on genuine data samples. 

Der, Martius, and Herrmann~\cite{der2012playful, martius10} present the concepts of homeostasis and homeokinesis.
Homeostasis is a principle for self-regulation to reach maximum stability. 
Agents predict the consequences of their actions in the near future using forward models. 
The agent's controller can be adapted so that the behavior is easily predictable by the forward model. 
Forward model and controller are trained simultaneously with supervised learning using the prediction error as a learning signal. 
Both are adapted to each other and a stabilized system is reached using only the agent's internal perspective.
Homeokinesis is the dynamical counterpart of homeostasis. 
It aims for self-exploratory robots that are active and innovative.
Der et al. introduce the time-loop error or reconstruction error: 
the difference between the reconstructed sensor values for time step~$t+1$ and the observed sensor values at time step~$t+1$.  
The time-loop error destabilizes the system when activity stagnates and thus, enables self-exploration.
Both concepts were studied in simulations and on real robots. 
Most of these robots were equipped with and reacted to proprioceptive sensors. 
In contrast to minimize surprise, both controller and forward model are trained using the prediction error in homeostasis and the time-loop error in homeokinesis.

Our minimal surprise approach aims for the evolution of collective robotics behaviors.
We do not rely on task-specific rewards~\cite{ha_2018, nolfi_learning_1994} or data from a system to be imitated~\cite{goodfellow_generative_2014, gross2017generalizing}. 
In contrast to all of the above approaches, we do not train both networks explicitly. 
In minimize surprise, only the prediction network is rewarded directly while the action network is subject to genetic drift. 
In the standard version of our approach, we do not directly influence emerging behaviors and allow the evolutionary process to come up with own, creative solutions almost only in the form of a by-product of the evolutionary dynamics. 

\subsubsection{Cellular Automata}

Cellular automata (CA) are mathematical models for dynamical systems that are, similar to our minimal surprise approach, discrete in space (grid-based) and discrete in time~\cite{wolfram1984}. 
The state of each grid cell is updated based on the state of its neighboring cells at every time step using a deterministic update rule. 
In our scenario, grid cells are not agents and are not associated to an update rule. However, our swarm system can be modeled such that cells switch their states based on whether they and their neighbors are occupied by robots.
Then the action networks of robots implement a cell's update rule. 
This is similar to cellular automata agents~\cite{hoffmann14}. 
The update rules of cellular automata can be manually designed or automatically generated, for example, using genetic algorithms. 
They are used for various applications, such as self-assembly and pattern formation. 
We give a brief overview over related approaches. 

The most popular cellular automaton is Conway's Game of Life~\cite{gardner1970, adamatzky2010}. 
In a 2D square grid, each cell changes its state between death and alive based on its Moore neighborhood. 
Living cells with two or three living neighbors survive and otherwise die. 
A~dead cell with three living neighbors becomes alive. 
These rules are applied simultaneously for each time step. 
Depending on the starting configuration, some patterns die out over time while most patterns become either stable or oscillating. 
Additionally, translating oscillators form that move across the grid (e.g., so-called `gliders'). 
Similar to Game of Life, we do not aim for specific patterns. 
However, we want to design a system that forms clusters and patterns independent of initial agent positions. 

Elmenreich and Feh\'{e}r\'{a}ri~\cite{Elmenreich2011} use cellular automata to reproduce a given image. 
They represent the CA's update rules with a time-discrete, recurrent artificial neural network that determines the cell's color.
The grid size matches the size of the reference image in pixels. 
Elmenreich and Feh\'{e}r\'{a}ri evolve the weights of the ANN and give fitness for a high pixel-wise resemblance of the formed pattern to the reference image.
They successfully evolved ANNs leading to the formation of regular patterns, such as flags (cf. results from morphogenetic engineering~\cite{doursat12}). 
More complex and larger reference images, like the Mona Lisa, did not lead to satisfying results. 
The researchers propose to apply their approach to swarm robotics for self-organized pattern formation.

Hoffmann and D\'{e}s\'{e}rable evolved a variety of patterns on a square grid using cellular automata agents (CAA).
Each grid cell contains a colored particle. 
The color can be changed by the agents to create a targeted pattern: path patterns~\cite{hoffmann14}, line patterns~\cite{hoffmann16}, domino patterns~\cite{Hoffmann2018b}, and checkerboard patterns~\cite{Hoffmann2018}. 
Additionally, a finite state machine (FSM) is located in every grid cell and is activated when an agent enters the cell. 
The FSM determines the agent's next action: a combination of moving, turning, and changing the cell's color. 
A~move forward is prevented if the grid cell in front is occupied by another agent. 
We use the same restriction in our self-assembly scenario. 
The FSM is evolved using a genetic algorithm.
Fitness is given based on the time required to form a pattern which has a predefined degree of resemblance to predefined local templates. 
The concept of local templates is similar to our predefined sensor predictions studied in Sec.~\ref{sec:SOSA}.  

\"Ozt\"urkeri and Johnson~\cite{ozturkeri2011} extended cellular automata to developmental cellular models (DCM) aiming for the self-assembly of target patterns. 
In contrast to classical CA, the update rule of DCMs can influence the neighboring cells, too. 
They use a (1+4)~evolutionary strategy and run evolution for $100,000$ generations.  
Fitness is given for stable patterns that have a high resemblance to the predefined target pattern. 
\"Ozt\"urkeri and Johnson successfully evolved regular patterns, that is, patterns with a repetitive or modular structure, such as a square, a diamond, or a french flag. 
The evolution of random patterns was less successful, which indicates that non-regular patterns are harder to construct. 
\"Ozt\"urkeri and Johnson proposed to use a fitness function that is independent of the target pattern in future.  
This may be similar to our intrinsically motivated minimal surprise approach.

Except for Conway's Game of Life~\cite{gardner1970}, all of the presented approaches aim for the creation of predefined patterns either specifying the target fully~\cite{Elmenreich2011, ozturkeri2011} or partially using local templates~\cite{hoffmann14}. 
Fitness is given for high resemblance to these structures. 
This is contrary to our minimal surprise approach. 
We make use of a fitness function that is completely independent of the formed structure. 
With our approach, we follow the idea to give swarm robotic systems freedom to evolve a variety of self-assembly behaviors with emergent behavioral diversity.

\section{Methods}
\label{sec:Methods} 

\subsection{Experimental Setup for Self-Assembly} 
\label{sec:SelfAssembly}

In previous works~\cite{borkowski17, hamann14d}, we found four basic swarm behaviors of rather low complexity using minimal surprise. 
We aim for the evolution of self-assembly behaviors to investigate more complex collective behaviors.
To govern the difficulty of our initial study, we restrict us to a 2D torus grid world which simplifies sensing and the equidistant positioning of robots.

We do experiments with varying swarm densities (i.e., robots per area: $\frac{N}{L \times L}$) on our 2D grid. 
As shown in previous work~\cite{hamann14d}, emerging swarm behaviors depend on the swarm density. 
For medium densities, the number of neighbors and the local pattern they form in a robot's sensor input change often making the prediction task harder than for low densities (no neighbors, no sensor input) or high densities (many neighbors, homogeneous sensor input).
Therefore, we keep the swarm size $N=100$ fixed while changing the side lengths $L\in\{15,20\}$ of the square representing our torus grid environment.
This leads to swarm densities of $44.4\%$ and $25\%$, respectively. 
Considerably lower or higher swarm densities would result in fewer and less interesting behaviors. 
Robots could achieve high fitness by rotating and staying on their grid cell while predicting either that no or all neighboring grid cells will be occupied.
Therefore, we focus on intermediate swarm densities ($L\in\{15,20\}$). 

In all evolutionary runs, we use a population size of~$50$, proportionate selection, elitism of one, and a mutation rate of~$0.1$ for both networks. 
We run evolution for~$100$ generations and evaluate each genome ten~times for~$500$ time steps with random initial robot positions. 
The fitness of a genome is the minimum fitness value~(Eq.~\ref{equ:fitness}) of those ten evaluations. 
Table~\ref{tab:TAB3} summarizes the parameters.

\begin{table}[tph]
       \centering
    \caption{Parameter settings~\cite{kaiser18}.
        \label{tab:TAB3}}
       \pgfplotstabletypeset[normal,
        column type=l,
       ]{ %
       parameter & value \\
 	   grid side length $L$ & \{15, 20\} \\ 
 		\# of sensors $R$& \{6, 8, 14\} \\ 
 		swarm size $N$ & 100 \\ \hline 
		population size & 50 \\ 
		number of generations & 100 \\ 
 		evaluation length in time steps $T$ & 500 \\ 
 		\# of sim. runs per fitness evaluation & 10 \\  
 		elitism & 1 \\ 
 		mutation rate & 0.1 \\   
       }
\end{table}

We simulate a swarm of robots with discrete headings: North, East, South, and West. 
Using its action network, a robot selects either to move one grid cell forward or to rotate by $\pm 90^\circ$ in each time step.
A robot can only move forward if the grid cell in front is empty as each grid cell can be occupied by only one robot at a time. 
Otherwise, an intended move forward is ignored and the robot stays on its current grid cell. 
This is similar to the intervention of a hardware protection layer in real robots that would be implemented to prevent robots from driving into obstacles, such as walls or other robots.
We require the robots to move (translation or rotation) to prohibit uninteresting behaviors where robots always stay stopped.
Nevertheless, robots can stay on their current grid cell by either constantly turning or by exploiting that forward movement is prevented when the grid cell in front is occupied.

To evaluate the resulting behaviors, we measure the fitness (prediction accuracy, Eq.~\ref{equ:fitness}) of the genomes and newly introduce temperature~$T$ of the system at runtime. 
We define temperature as the robots' motion, that is, the covered distance of the robots per time step averaged over all robots. 
So the forward motion of robots is considered, while their rotations are ignored.
This measurement roughly follows the concepts of thermodynamics where temperature is proportional to the average kinetic energy of the molecule's center-of-mass motion~\cite{feynman10}. 
The temperature cools down when there are more robots that stay on their grid cells, that is, they turn or are blocked in their forward movement.
A~hot system is in a disordered state with many moving robots and a cool system relates to a more ordered system which has assembled into a structure (cf. laws of thermodynamics~\cite{feynman10}). 
We define~$T$ as 

\begin{equation}
	T(t) = M_{x}(t) + M_{y}(t)\, ,
	\label{equ:temperature}
\end{equation} 

where $M_{x}(t)$ and $M_{y}(t)$ are the movements in $x$- and $y$-direction of the grid in time step $t$.
We define $M_{x}(t)$ as

\begin{equation}
M_{x}(t) = \frac{1}{N} \sum_{n=0}^{N-1} | x_{n}(t) - x_{n}(t+1) |\, ,
\label{equ:dx}
\end{equation} 

and $M_{y}$ accordingly. 
This is the sum over the robots' movements, normalized by the number of robots~$N$. 

Furthermore, we measure the robot movement~$M$, that is the average covered distances of the robots as an integral of displacement at each time step over a time period of $\tau = \frac{L \times L}{2}$ time steps~(cf.~\cite{hamann14a}). 
This equals the average temperature over $\tau$.
Thus, the robot movement~$M$ is the normalized sum of the system's temperature over the time period $\tau$. 
We define the robot movement~$M$ as 

\begin{equation}
	M = \frac{1}{\tau} \sum_{t = T - \tau}^{T-1} T(t)\, . 
	\label{equ:movement}
\end{equation} 

In addition, we determine the intended robot movement~$I$. 
It is the average distance that all robots would have covered over time period $\tau$ according to their selected action values if there were no obstacles. 
We define the intended robot movement~$I$ as 

\begin{equation}
I = \frac{1}{N\tau} \sum_{n=0}^{N-1} \sum_{t = T - \tau}^{T-1} a_{n}(t)\, ,
\label{equ:intendedmovement}
\end{equation} 

where $a_{n}(t)$ is the action value of robot~$n$ at time step $t$, which is~1 for moving one grid cell forward and~0 for rotation. 
The robots are prevented to move straight when a targeted grid cell is occupied and thus, $M$ and $I$ can deviate.

\subsection{Classification of Emergent Structures} \label{sec:classification}

\begin{figure}[tph]
    \centering
    \subfloat[line]{\includegraphics[width=0.2\textwidth]{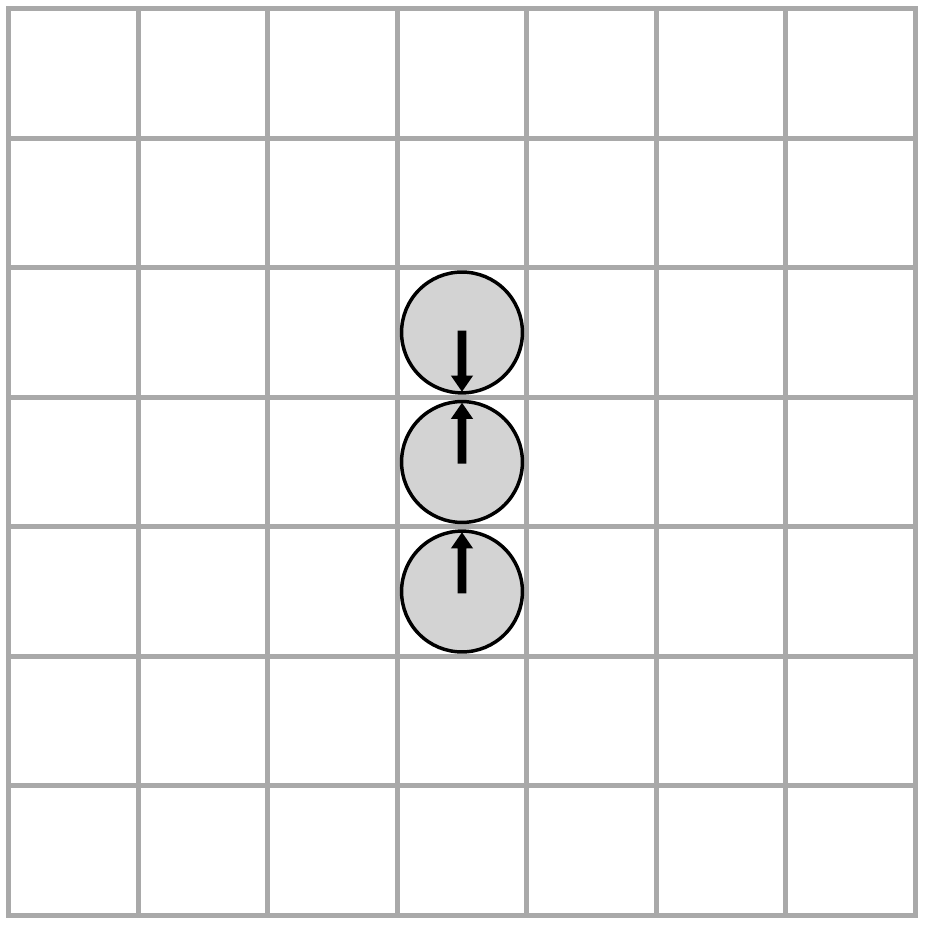}\label{fig:line}}
    \hspace{4mm}
    \subfloat[pairs]{\includegraphics[width=0.2\textwidth]{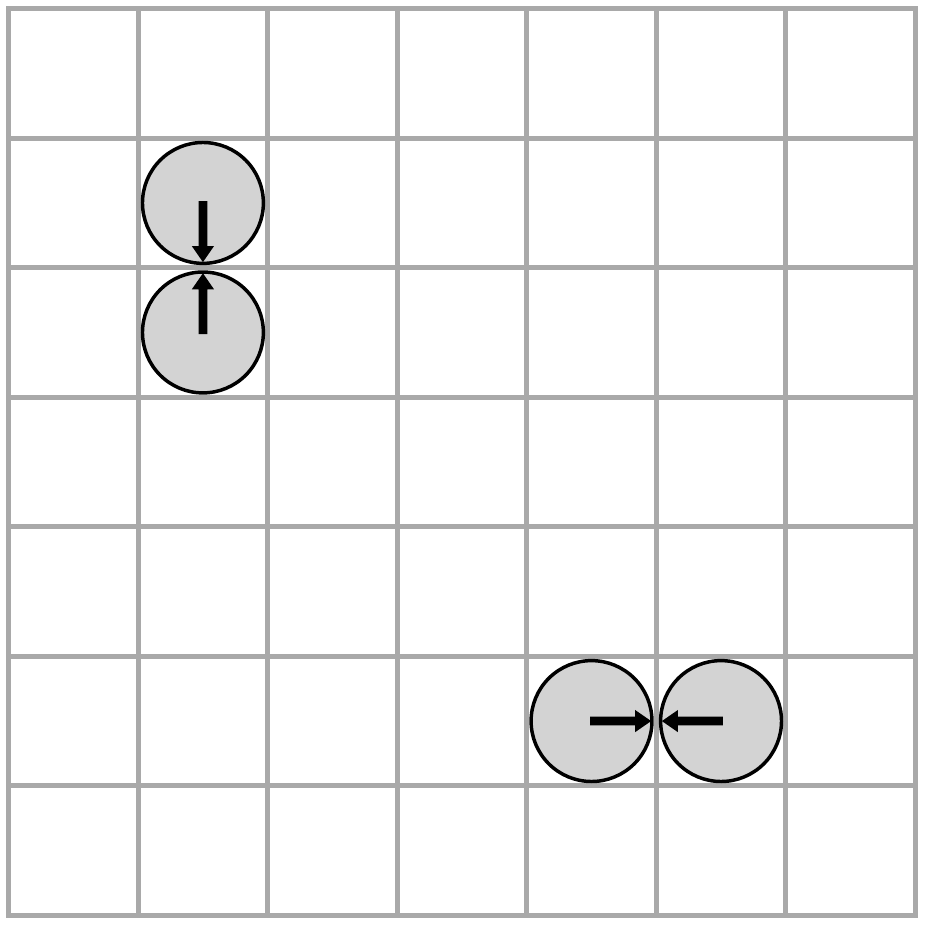}\label{fig:pairs}}
    \\
    \subfloat[aggregation]{\includegraphics[width=0.2\textwidth]{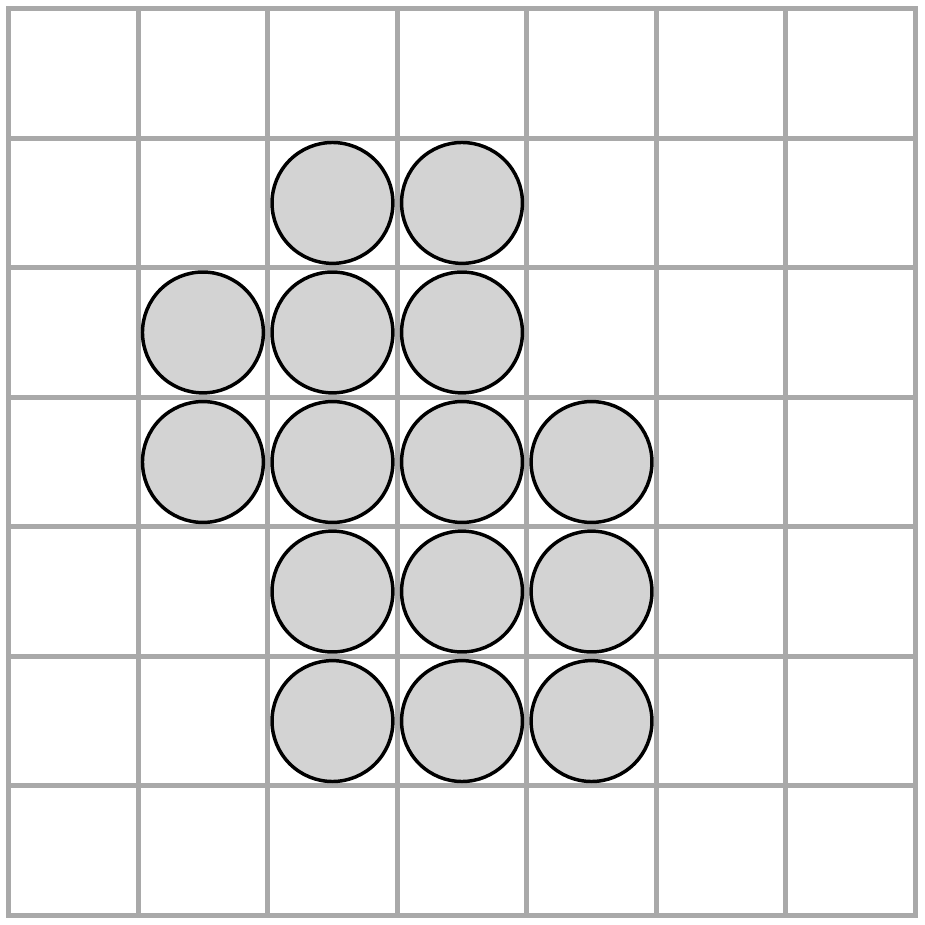}\label{fig:aggregation}}
    \hspace{4mm}
    \subfloat[clustering]{\includegraphics[width=0.2\textwidth]{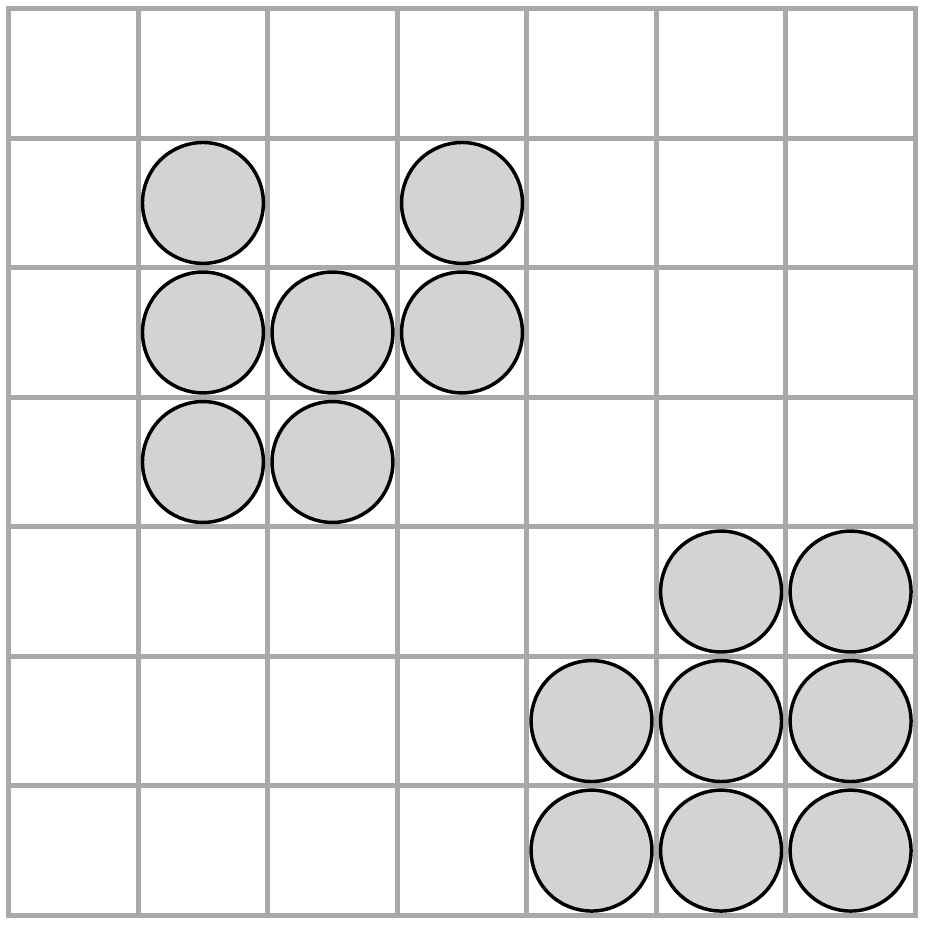}\label{fig:clustering}}
    \hspace{4mm}
    \subfloat[loose grouping]{\includegraphics[width=0.2\textwidth]{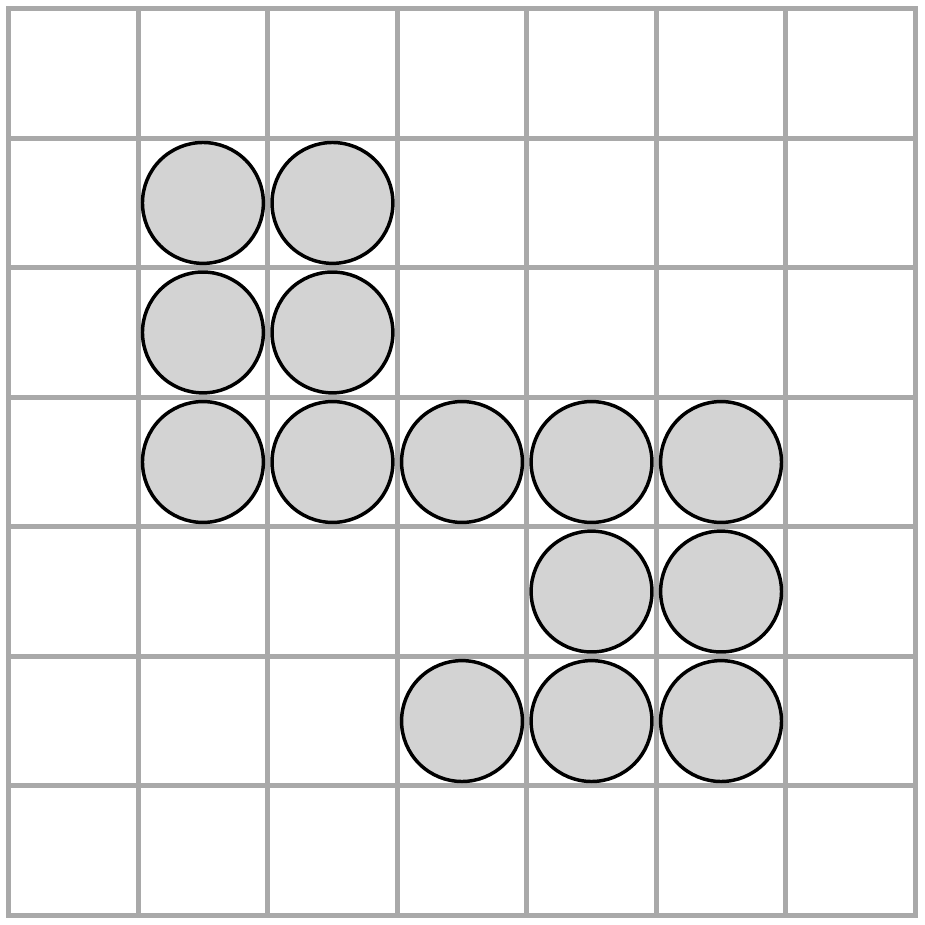}\label{fig:grouping}}
    \\
    \subfloat[square pattern]{\includegraphics[width=0.2\textwidth]{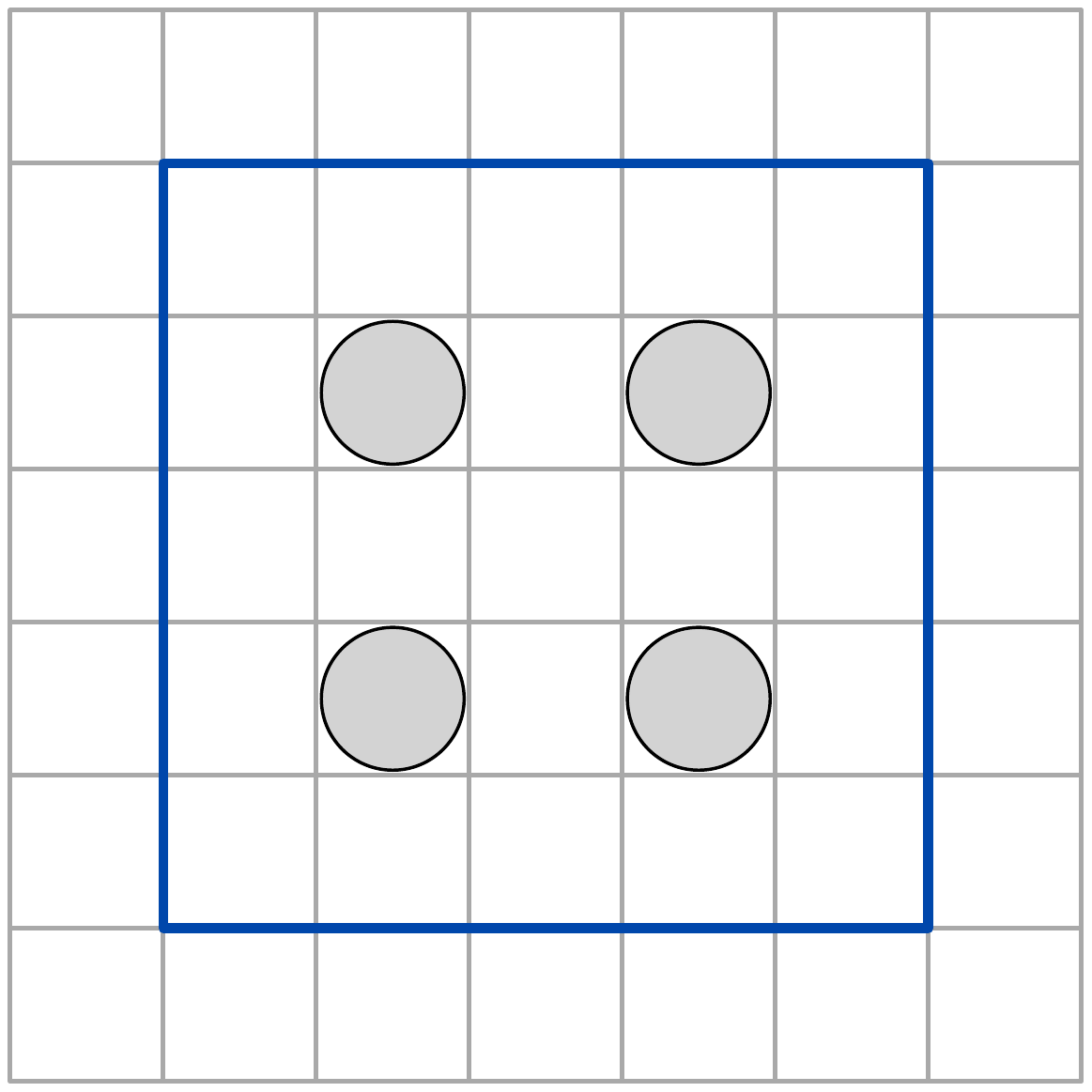}\label{fig:FIG2a}} 
    \hspace{4mm}
    \subfloat[triangular lattice]{\includegraphics[width=0.2\textwidth]{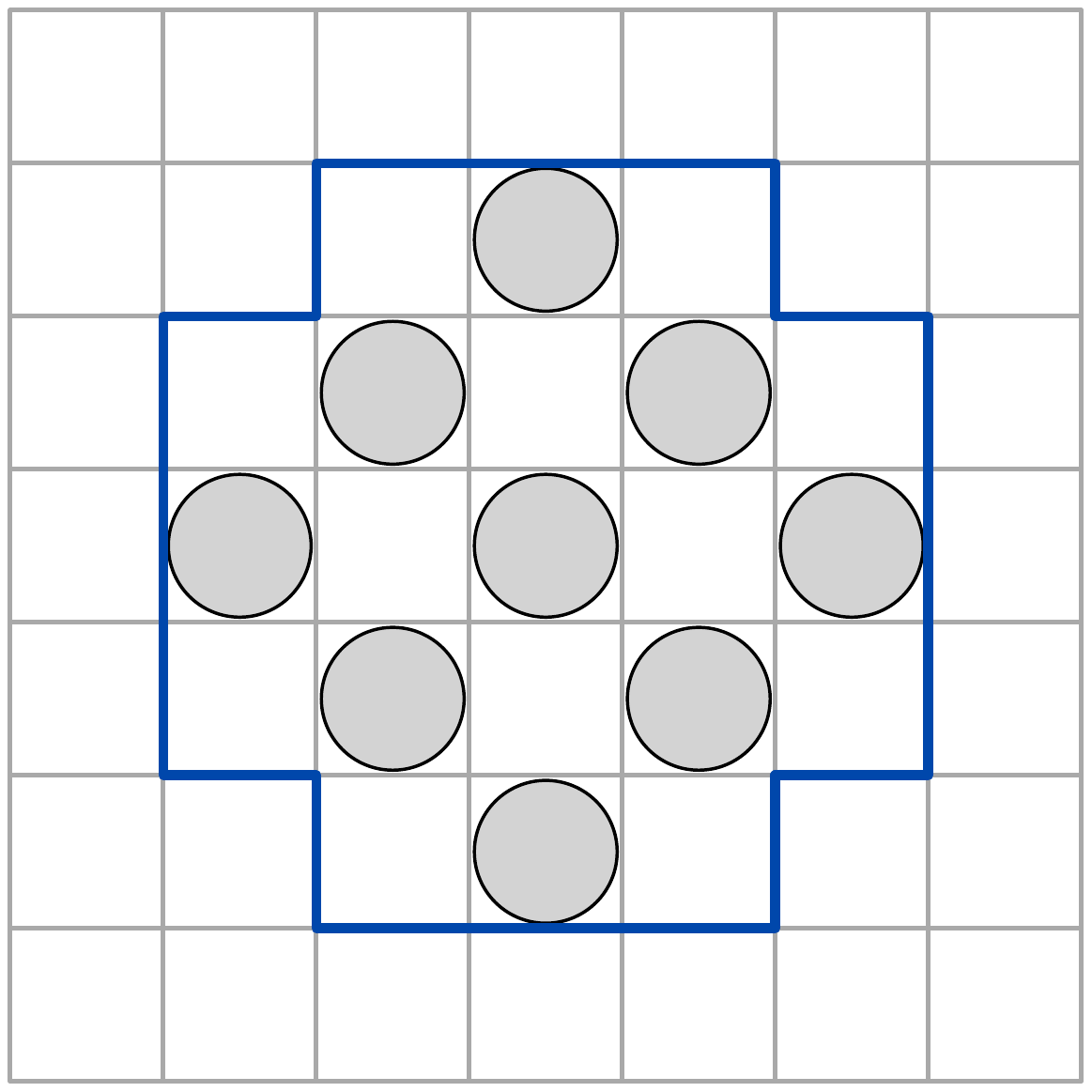}\label{fig:FIG2b}} 
    \hspace{4mm}
    \subfloat[random dispersion]{\includegraphics[width=0.2\textwidth]{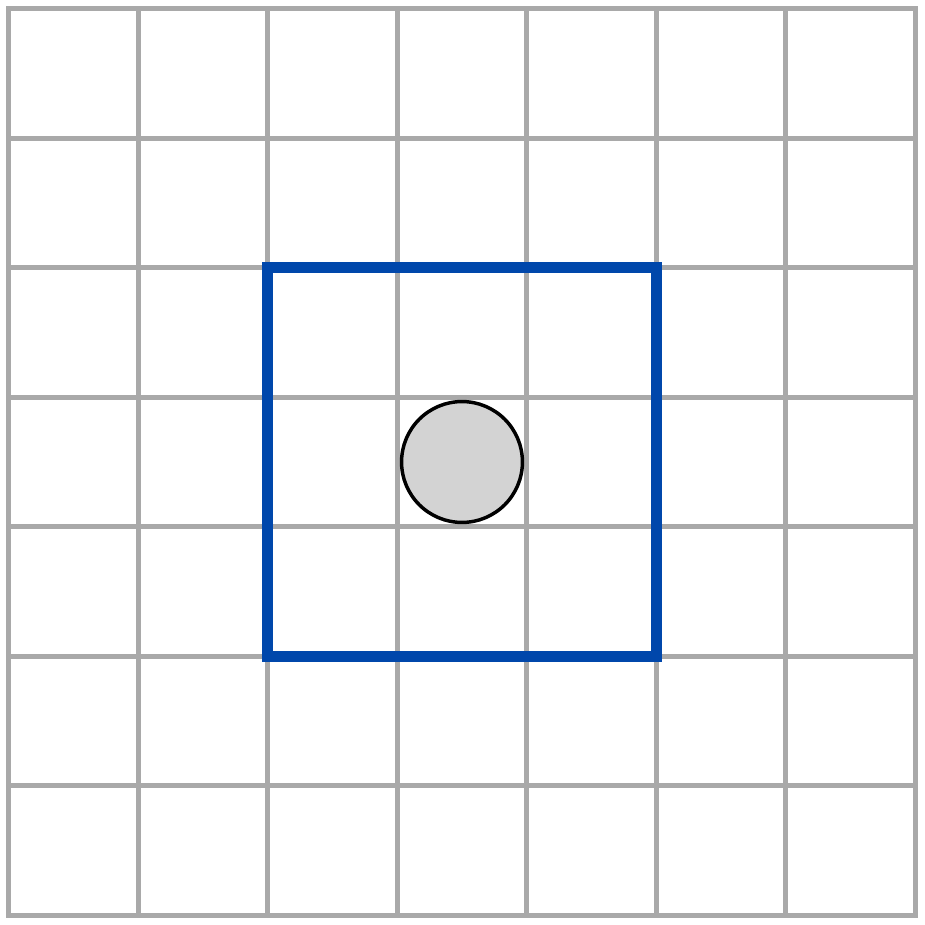}\label{fig:dispersion}}
    \caption{Examples of patterns, robots are represented by \textit{gray circles}; where relevant, headings are represented by \textit{black arrows}. \textit{White cells} inside areas marked by \textit{blue rectangles} have to be empty, with the only exception in~(h): one additional robot is allowed within the marked area.}
    \label{fig:FIG2}
\end{figure}

We define metrics for eight different patterns to replace our previously qualitative analysis with a quantitative classification of the resulting structures.
These patterns are lines, pairs, aggregation, clustering, loose grouping, random dispersion, squares, and triangular lattices, see Fig.~\ref{fig:FIG2}. 
All of these patterns allow for a high prediction accuracy.
The patterns are either rotation symmetric (e.g., triangular lattices or squares) or form a repetitive pattern by exploiting that movement is blocked when the targeted grid cell is occupied (e.g., lines or pairs). 
This allows robots to create boring and easy to predict environments that only require a constant prediction of the same sensor values to reach a high fitness value when staying/rotating on a grid cell. 
In all following experiments, we categorize resulting structures based on the highest resemblance to one of these patterns. 

Lines (Fig.~\ref{fig:line}) and pairs (Fig.~\ref{fig:pairs}) are structurally similar but differentiate in length. 
Lines have a minimum length of three robots while pairs consist of exactly two robots. 
The following rules apply to both structures: 

\begin{enumerate}
    \item The headings of all robots have to be parallel. 
    \item The structure has to be terminated by robots pointing inwards the structure at both ends. 
    \item The maximum number of allowed neighbors on each side next to the structure is half of the structure's length. Neighbors are not allowed to be positioned on two adjacent grid cells parallel to the structure.   
\end{enumerate}

Rule number two does not apply if a line spans the whole grid length, that is, it is a ring around the whole torus. 
The first two rules ensure that the formed structure is stable as robots stay in the structure when attempting to move straight.  

We define three grouping behaviors: aggregation (Fig.~\ref{fig:aggregation}), clustering (Fig.~\ref{fig:clustering}) and loose grouping (Fig.~\ref{fig:grouping}). 
A cluster is formed by robots which are connected by common neighbors in their Moore neighborhoods. 
A robot requires at least six neighbors in its Moore neighborhood to be a member of a cluster. 
Only one neighbor in the von Neumann neighborhood is allowed to be missing as only such can block the movement of another robot and trigger the formation of a stable structure. 
All neighbors of a robot in a cluster are considered as parts of that cluster and hence some robots at the border of a cluster might have less than six neighbors.  
This can lead to the fact that these robots are part of two clusters and connect them. 
Consequently, separate clusters are at least one grid cell apart. 
We define clustering as the formation of several separate clusters and aggregation as the formation of one single cluster.
Completely interconnected clusters are considered as loosely grouped meaning that all clusters are either directly connected or connected via other clusters. 

In addition, we define three dispersion behaviors: random dispersion, triangular lattices, and squares. 
The headings of the robots are not relevant for dispersion patterns as the robots have to constantly turn to stay on their grid cell. 
Each robot having maximally one neighbor in its Moore neighborhood is classified as randomly dispersed. 

The square pattern is illustrated in Fig.~\ref{fig:FIG2a}. 
The corners of the inner $3\times 3$~grid are occupied by robots while all other grid cells in the surrounding $5\times 5$~grid have to be empty. 
The triangular lattice pattern ensures that the robot in the center sees a rotation-symmetric pattern formed by its neighbors as shown in Fig.~\ref{fig:FIG2b}. 
All robots within these patterns are classified to be part of the respective structure. 

All defined metrics are based on the results of the experiments presented in the next sections. 
We automatically determine the resemblance to the individual patterns using Python scripts\footnote{\url{https://github.com/msminirobot/minimal-surprise-self-assembly}}.

\section{Results} \label{sec:results}

\subsection{Sensor Model}
\label{sec:sensormodel} 

\begin{figure}[tph]
    \centering
    \subfloat[sensor model A]{\includegraphics[width=0.2\textwidth]{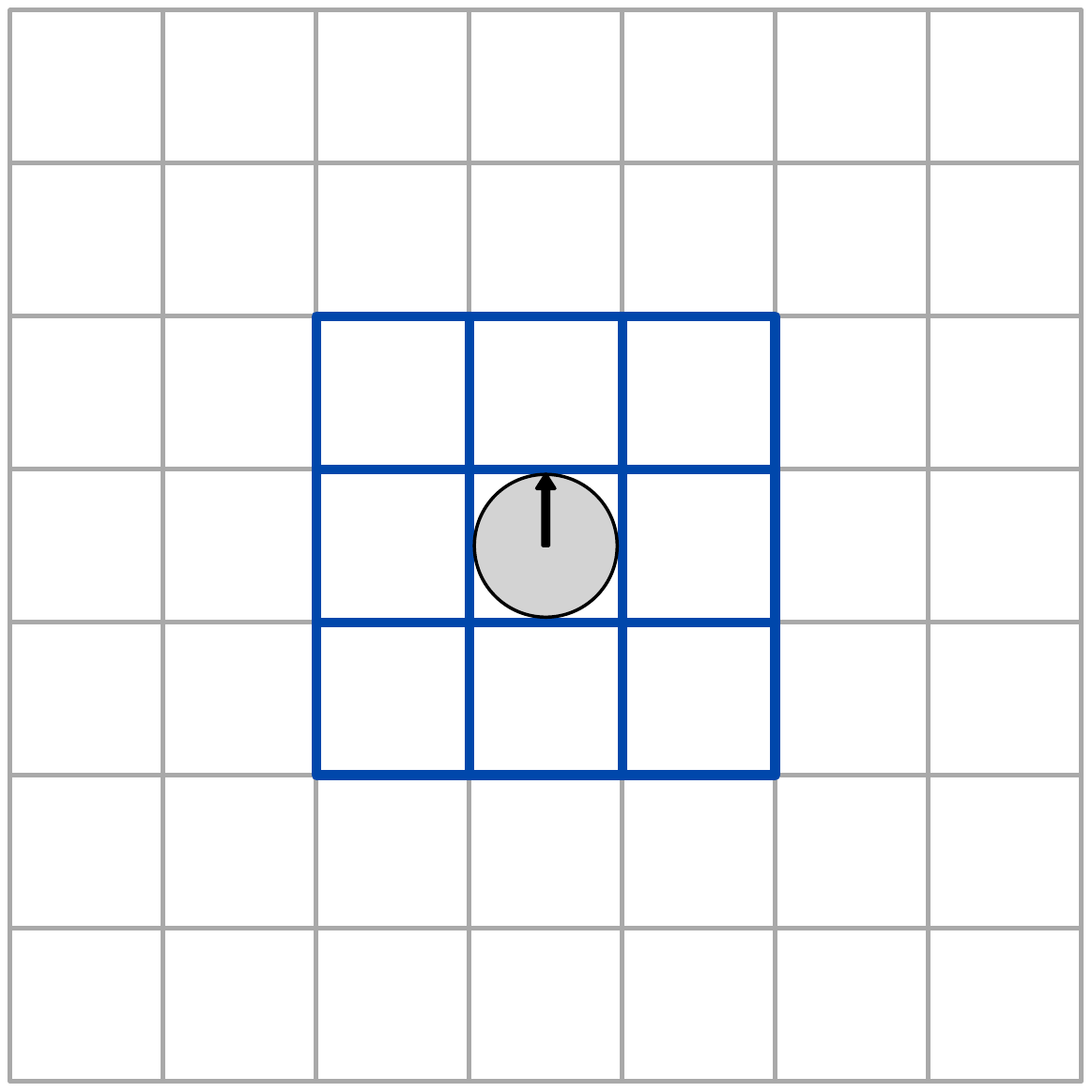}\label{fig:FIG3a}}
    \subfloat[sensor model B]{\includegraphics[width=0.2\textwidth]{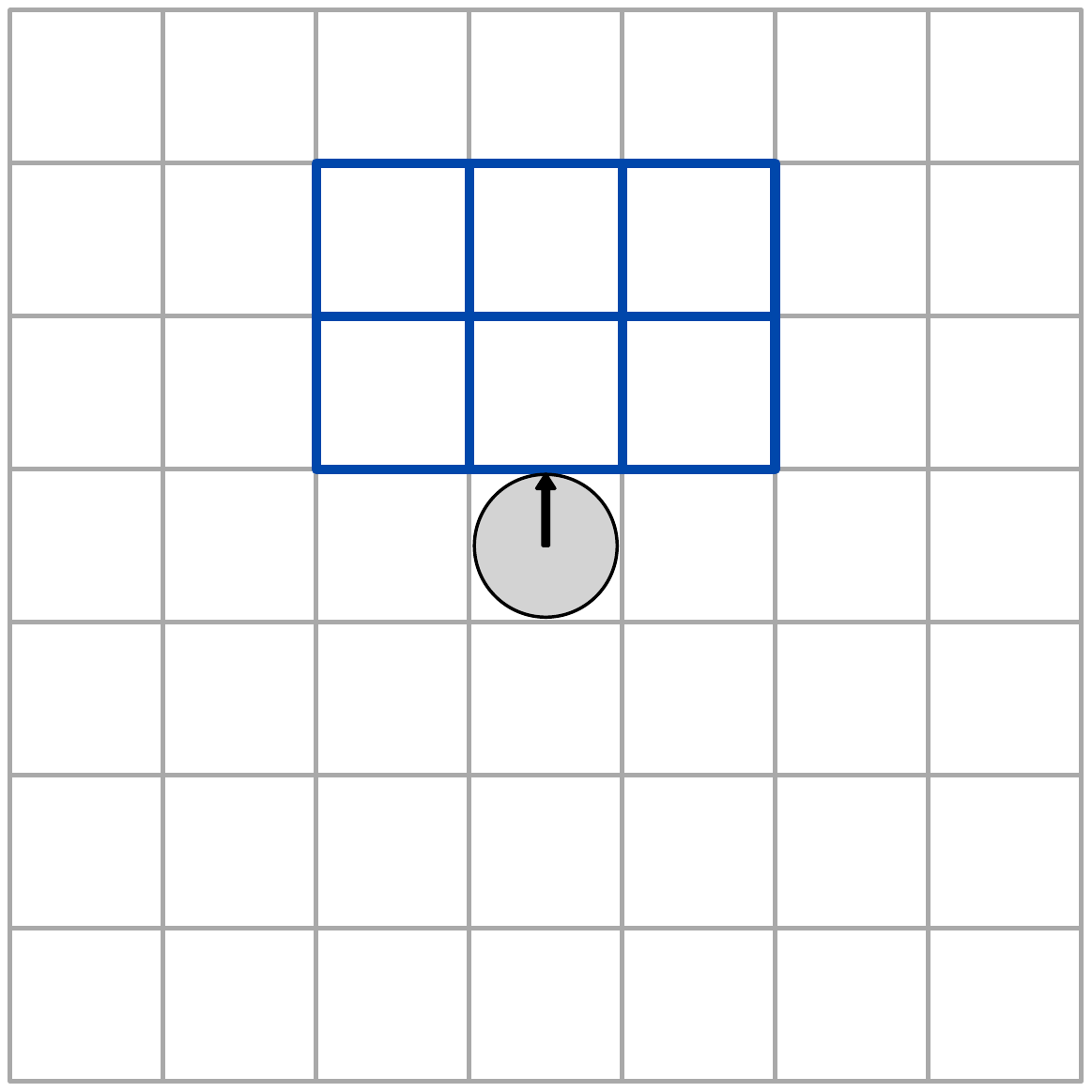}\label{fig:FIG3b}} 
    \subfloat[sensor model C]{\includegraphics[width=0.2\textwidth]{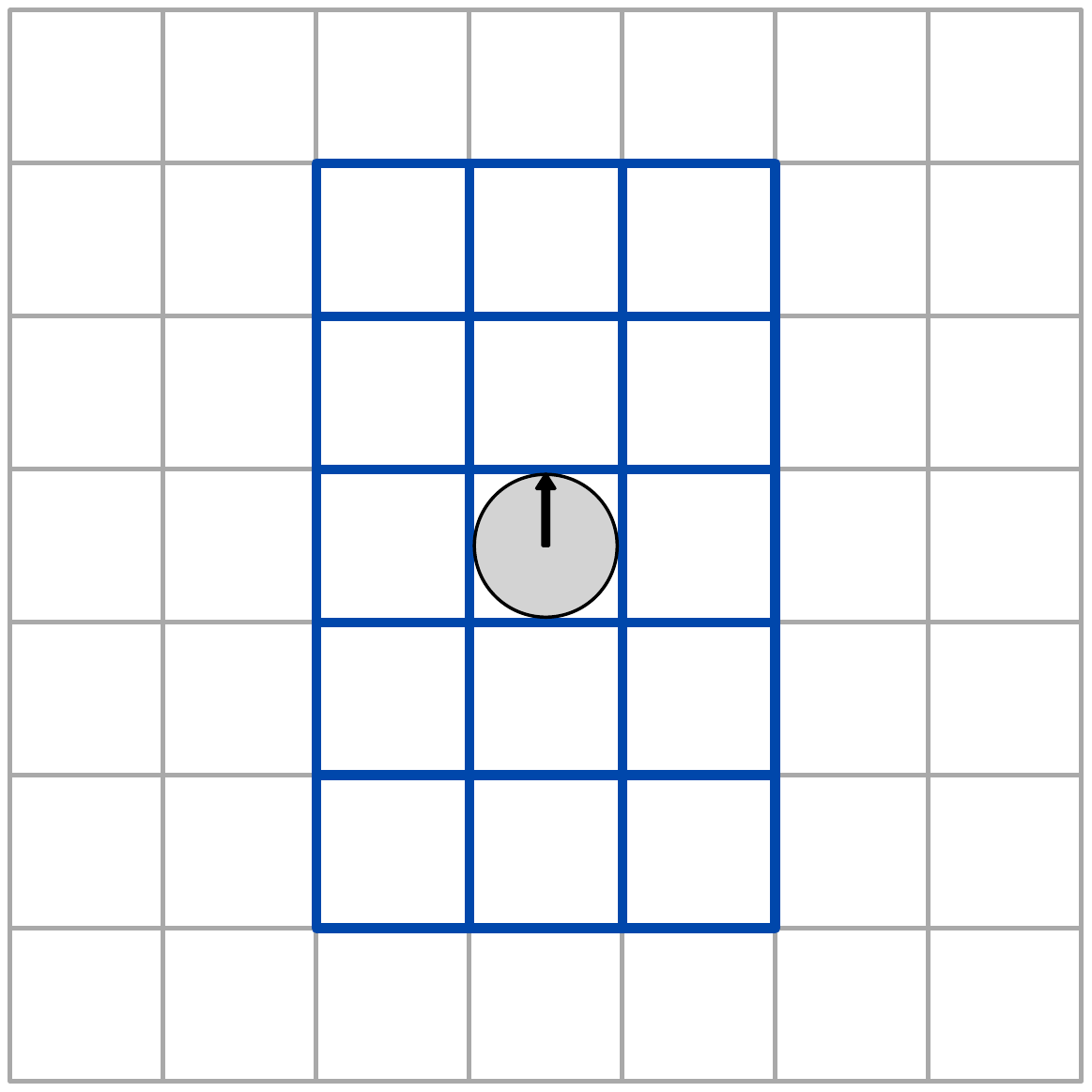}\label{fig:FIG3c}} 
    \caption{Sensor models evaluated in preliminary investigations. \textit{Gray circles} represent robots, {black arrows} indicate their headings.}
    \label{fig:FIG3}
\end{figure}

We extend our previous work~\cite{kaiser18} by a comparison of three different sensor models with discrete sensors to justify our selected sensor model in all following experiments.
Sensor model~A covers the robot's Moore neighborhood (see Fig.~\ref{fig:FIG3a}), sensor model~B the six grid cells in front of the robot (see Fig.~\ref{fig:FIG3b}), and sensor model~C 14~surrounding grid cells (see Fig.~\ref{fig:FIG3c}). 
We evaluate each sensor model in 20~independent evolutionary runs and classify the resulting structures based on the metrics presented in Sec.~\ref{sec:classification}. 

\begin{table}[tph]
       \centering
        \caption{Percentage of resulting structures on a $15 \times 15$ grid per sensor model. \label{tab:TAB1}}
       \pgfplotstabletypeset[normal,
        columns/\space/.style={column name = , 
        column type = l},
        columns/aggre- gation/.style = { column type = C }, 
        columns/loose grouping/.style = { column type = C }, 
        columns/triang. lattice/.style = {column type = C },
        columns/loose disp./.style = {column type = C },
       ]{ %
         \space & lines & aggre- gation & clustering & loose grouping & triang. lattice & pairs\\ 
         sensor model B & 30 & 15 & 35 & 15 & 0 & 5 \\ 
         sensor model C & 15 & 10 & 55 & 15 & 5 & 0 \\  
       }

        \bigskip
        
        \caption{Percentage of resulting structures on a $20\times 20$ grid per sensor model. \label{tab:TAB2}}
       \pgfplotstabletypeset[normal,
        columns/\space/.style={column name = , 
        column type = l},
        columns/aggre- gation/.style = { column type = C }, 
        columns/loosely grouped/.style = { column type = C }, 
        columns/triang. lattice/.style = {column type = C },
        columns/random dispersion/.style = {column type = E },
       ]{ %
         \space & pairs & lines & clustering & random dispersion & squares \\ 
         sensor model A  & 5 & 0 & 5 & 75 & 15 \\ 
         sensor model B & 35 & 65 & 0 & 0 & 0 \\ 
         sensor model C & 27.5 & 22.5 & 10 & 40 & 0 \\ 
       }
\end{table} 

\begin{figure}[tph]
\centering
\includegraphics[width=0.3\textwidth]{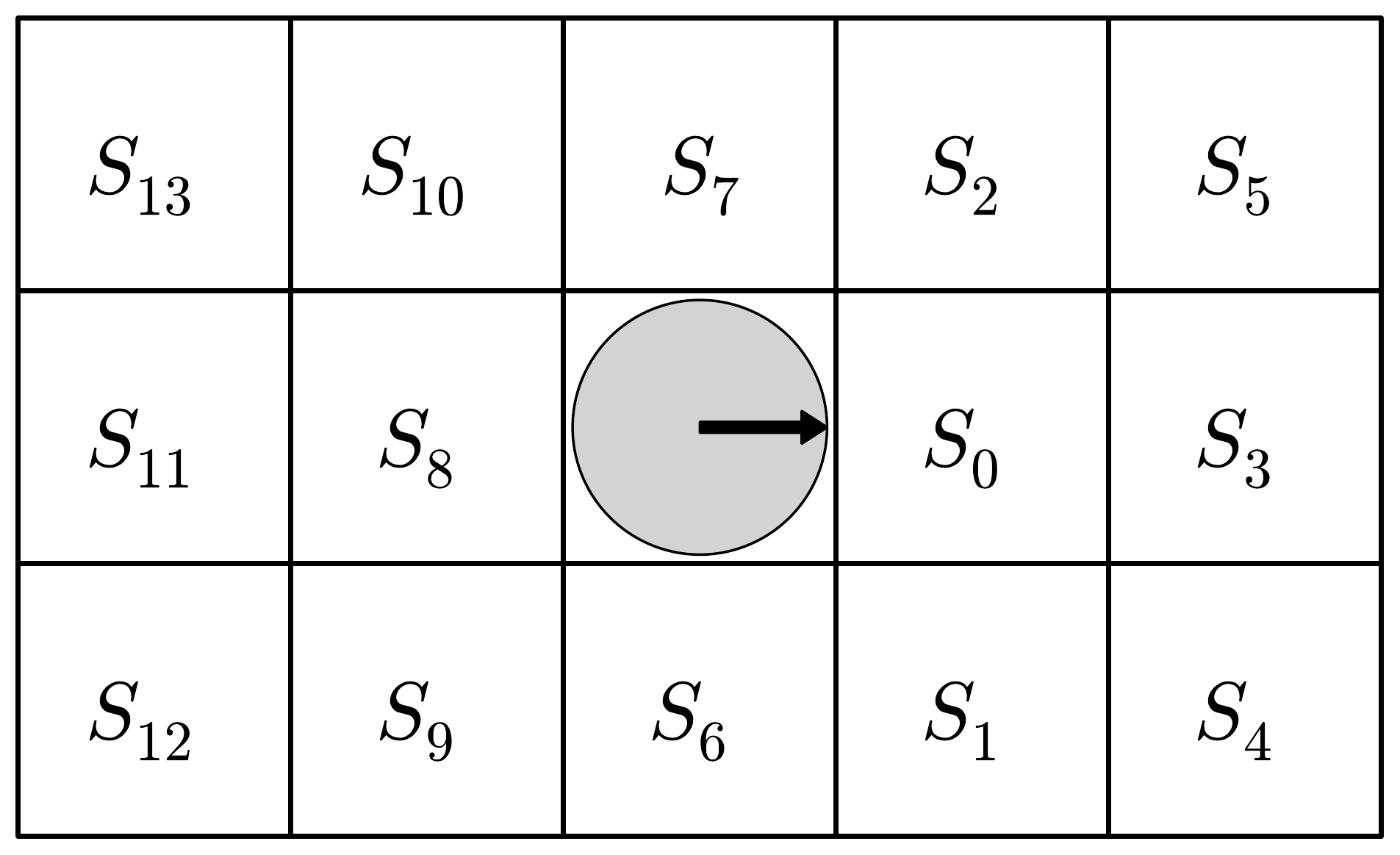}
\caption{Labels for the individual sensors of sensor model~C. The \textit{gray circle} represents the robot and the \textit{black arrow} its heading. Reprinted by permission from Springer Nature Customer Service Centre GmbH: \cite{kaiser18}, \copyright Springer Nature Switzerland AG 2019.}
\label{fig:FIG4}
\end{figure}

On the $15\times 15$ grid, all three sensor models lead to the emergence of a variety of behaviors, cf. Table~\ref{tab:TAB1}. 
In comparison to sensor model~A, sensor model~B did not lead to the emergence of triangular lattices while no pairs emerged using sensor model~C. 
On the $20\times 20$ grid the difference between the sensor models becomes more significant, cf. Table~\ref{tab:TAB2}. 
Sensor model~B resulted solely in the emergence of lines and pairs. 
Both sensor models~A and~C led to the emergence of four different structures. 
No line structures, but a majority of dispersion behaviors emerged using sensor model~A.
In contrast, using sensor model~C led to the formation of lines and pairs in many cases and in less cases to dispersion behaviors. 
We use sensor model~C in all following experiments as it enables us to evolve varieties of structures independent of swarm density.
We reference the individual sensors of sensor model~C as indicated in Fig.~\ref{fig:FIG4}.

Due to the chosen sensor model, both networks have 15~input neurons consisting of 14~sensor values and one action value each. 
Furthermore, the action network has eight hidden neurons and two output neurons which determine action value and turning direction.
The prediction network has 14~hidden neurons and 14~output neurons which give the discrete predictions for the 14~sensors. 

We evaluate all following experiments in 50~independent evolutionary runs. 

\subsection{Impact of Varying Swarm Densities}
\label{sec:Adaptation} 

We investigate the effects of different swarm densities on the emergence of structures in two scenarios.
The classification of the patterns was done qualitatively based on visual appearance in our previous work~\cite{kaiser18}, but we updated the numbers of the next two sections using the metrics defined in Sec.~\ref{sec:classification} to classify the structures.
In the first setting, we use a $15\times 15$ grid resulting in a swarm density of approx.~$0.44$ while we have a swarm density of $0.25$ in the second scenario using a $20\times 20$ grid. 
A video illustrating the resulting self-assembly behaviors is available online\footnote{\textit{Video1.mp4} -  \url{https://doi.org/10.5281/zenodo.3362285}}. 

\begin{figure}[t]
    \centering
        \subfloat[grid size: $15\times15$]{\includegraphics[width=0.4\textwidth]{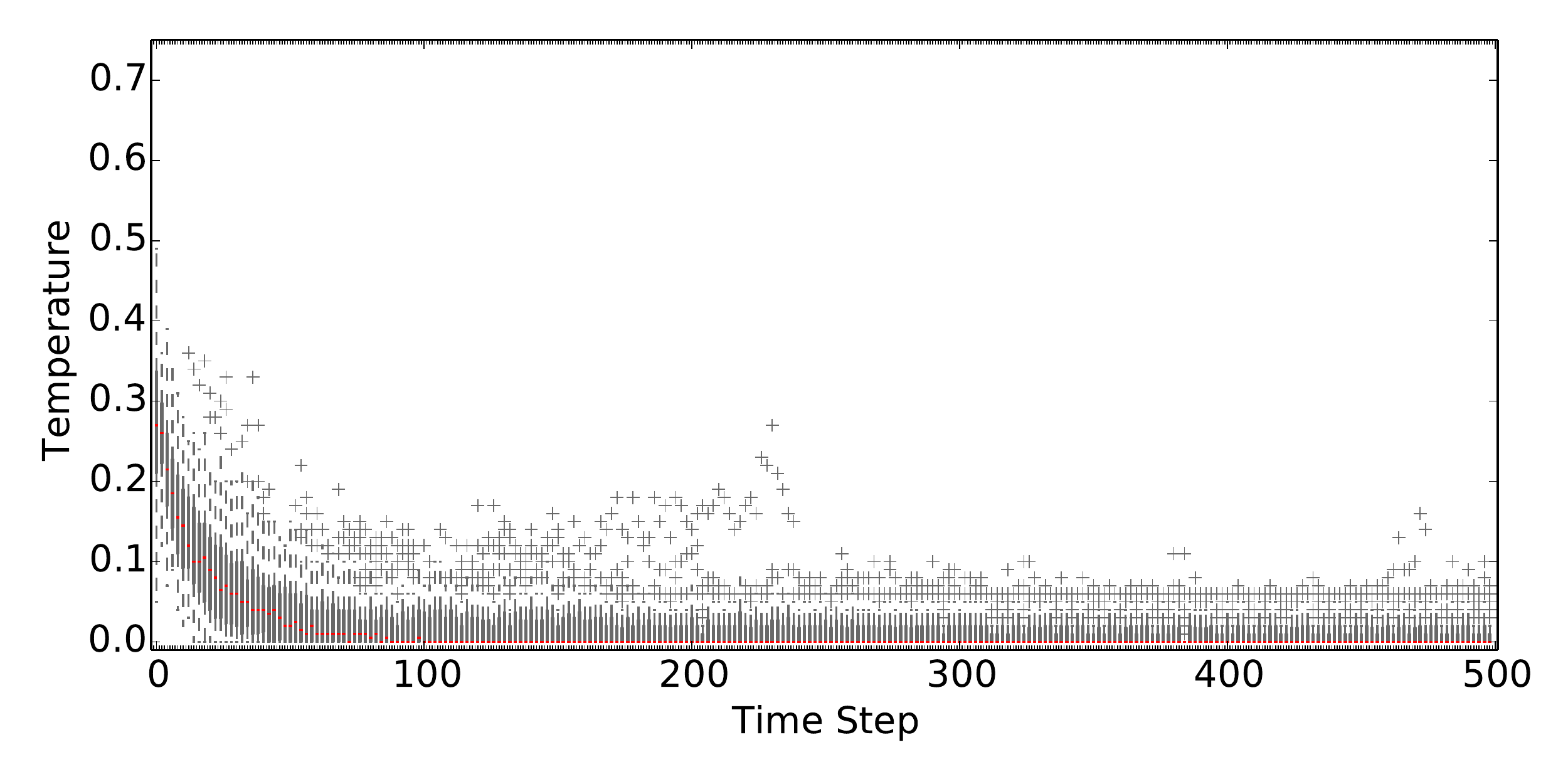}}
        \subfloat[grid size: $20\times20$]{\includegraphics[width=0.4\textwidth]{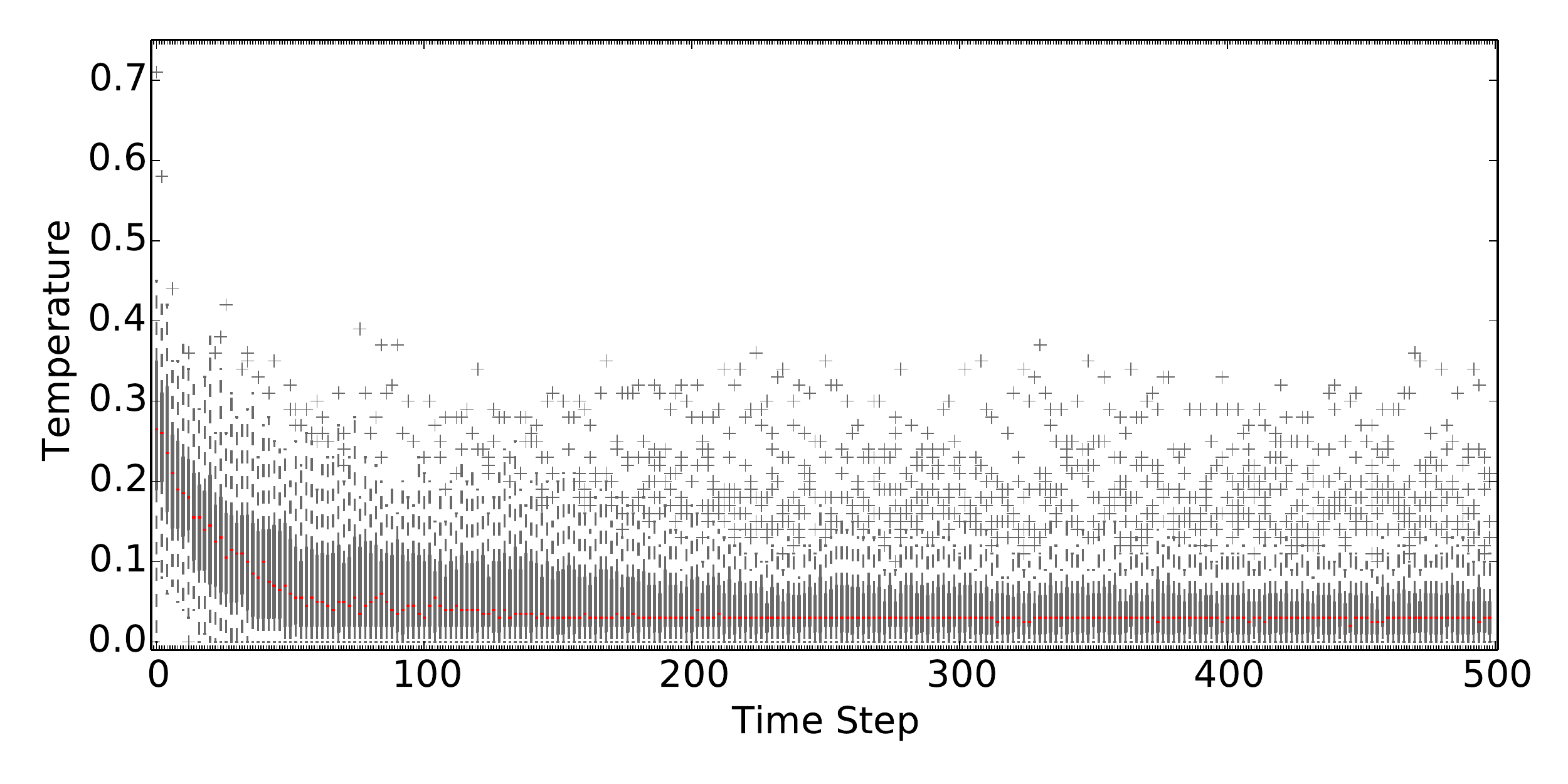}\label{fig:FIG5b}}\\
        \subfloat[grid size: $20\times 20$, random dispersion]{\includegraphics[width=0.4\textwidth]{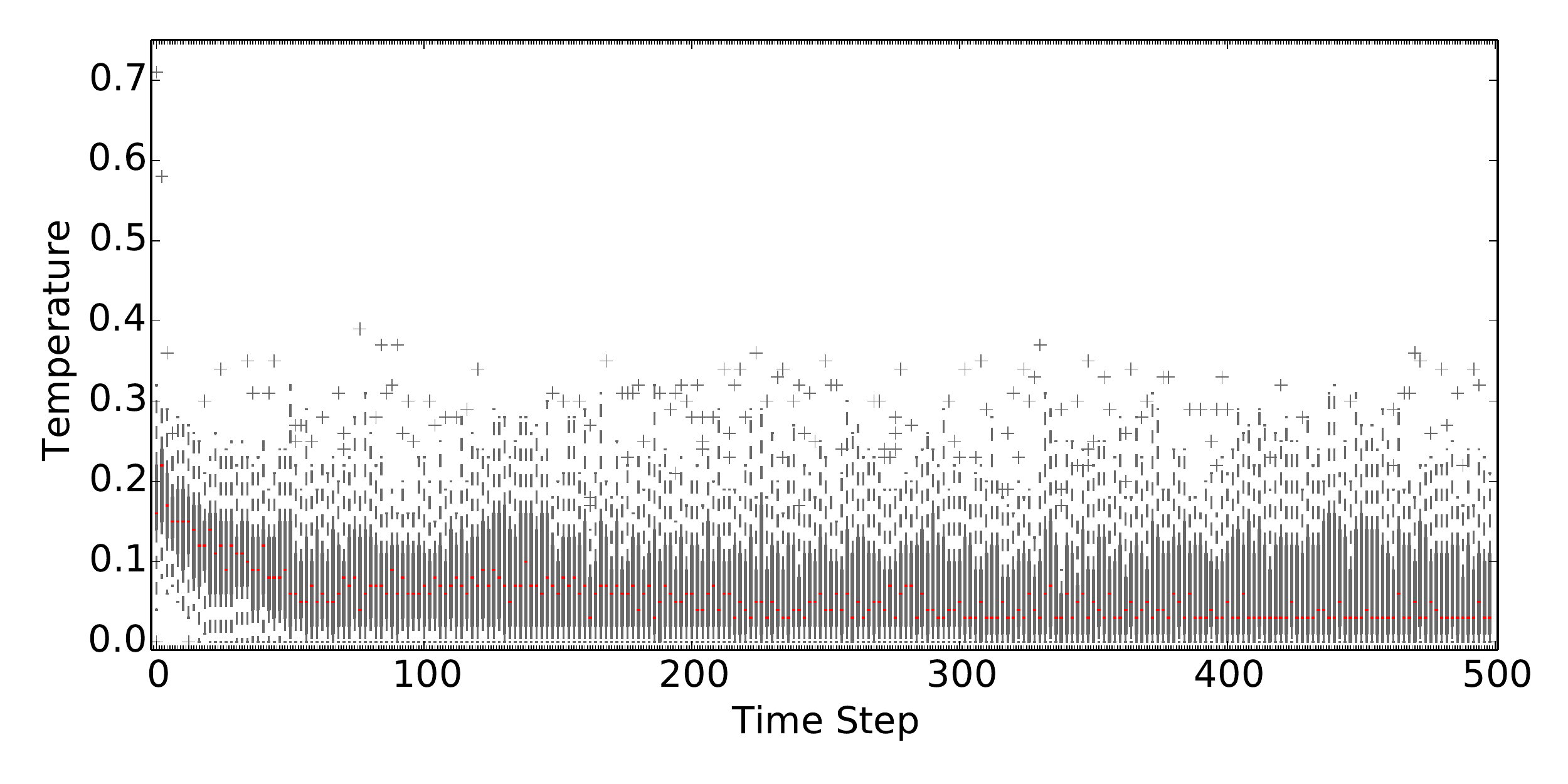}\label{fig:FIG5c}}
        \subfloat[grid size: $20\times 20$, other behaviors]{\includegraphics[width=0.4\textwidth]{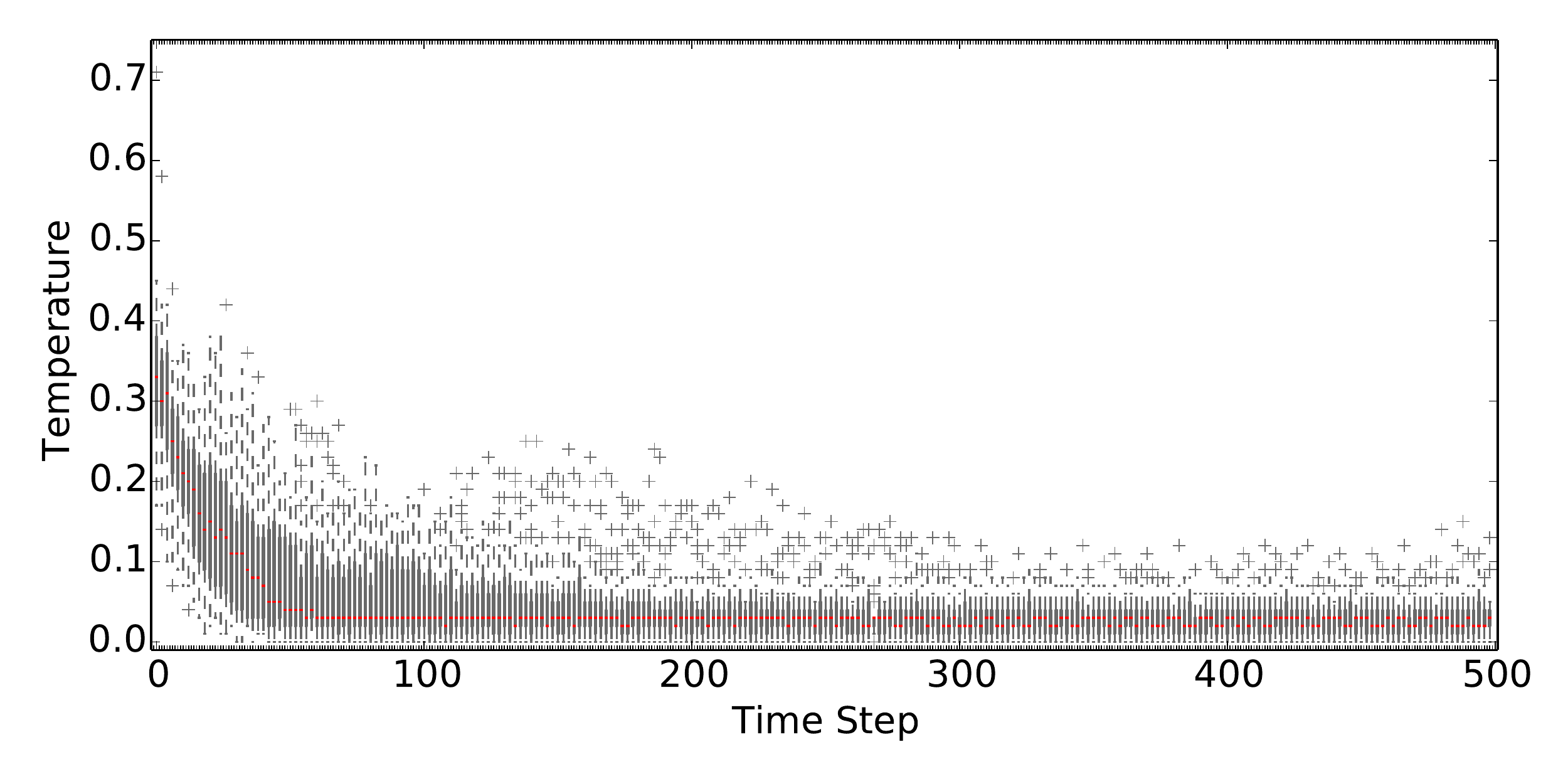} \label{fig:FIG5d}}
    \caption{Temperature of 50~independent runs for different grid sizes in~(a) and~(b). (c)~and~(d) show the measured temperature of the data given in~(b) separately for random dispersion behaviors and all other behaviors. Medians are indicated by \textit{red bars}. Only data of even time steps is plotted.}
    \label{fig:FIG5}
\end{figure}

The measured temperature (Eq.~\ref{equ:temperature}) shown in Fig.~\ref{fig:FIG5} illustrates how the systems cool down over time. 
Already after around 100~time steps almost all robots are staying stopped and thus, the robots quickly assemble into a structure. 
We measure a median temperature value of $0.03$ on the $20\times 20$ grid and of zero on the $15 \times 15$ grid in the last time step. 
This means that $3\%$ and $0\%$, respectively, of the robots moved one grid cell forward.
The outliers in Fig.~\ref{fig:FIG5b} are mostly caused by random dispersion behaviors. 
This is illustrated in Figs.~\ref{fig:FIG5c} and~\ref{fig:FIG5d} which show the measured temperature separately for random dispersion behaviors and all other behaviors. 
A reason might be that robots attempt to spread as widely apart as possible, but the swarm density does not allow to have no neighbors within sensor view.  
On the $15\times 15$ grid, no random dispersion behaviors emerge (cf.~Table~\ref{tab:TAB4}) and thus, there are less outliers in the temperature curve. 

\begin{figure}[tph]
    \centering
        \subfloat[grid size: $15\times15$]{\includegraphics[width=0.4\textwidth]{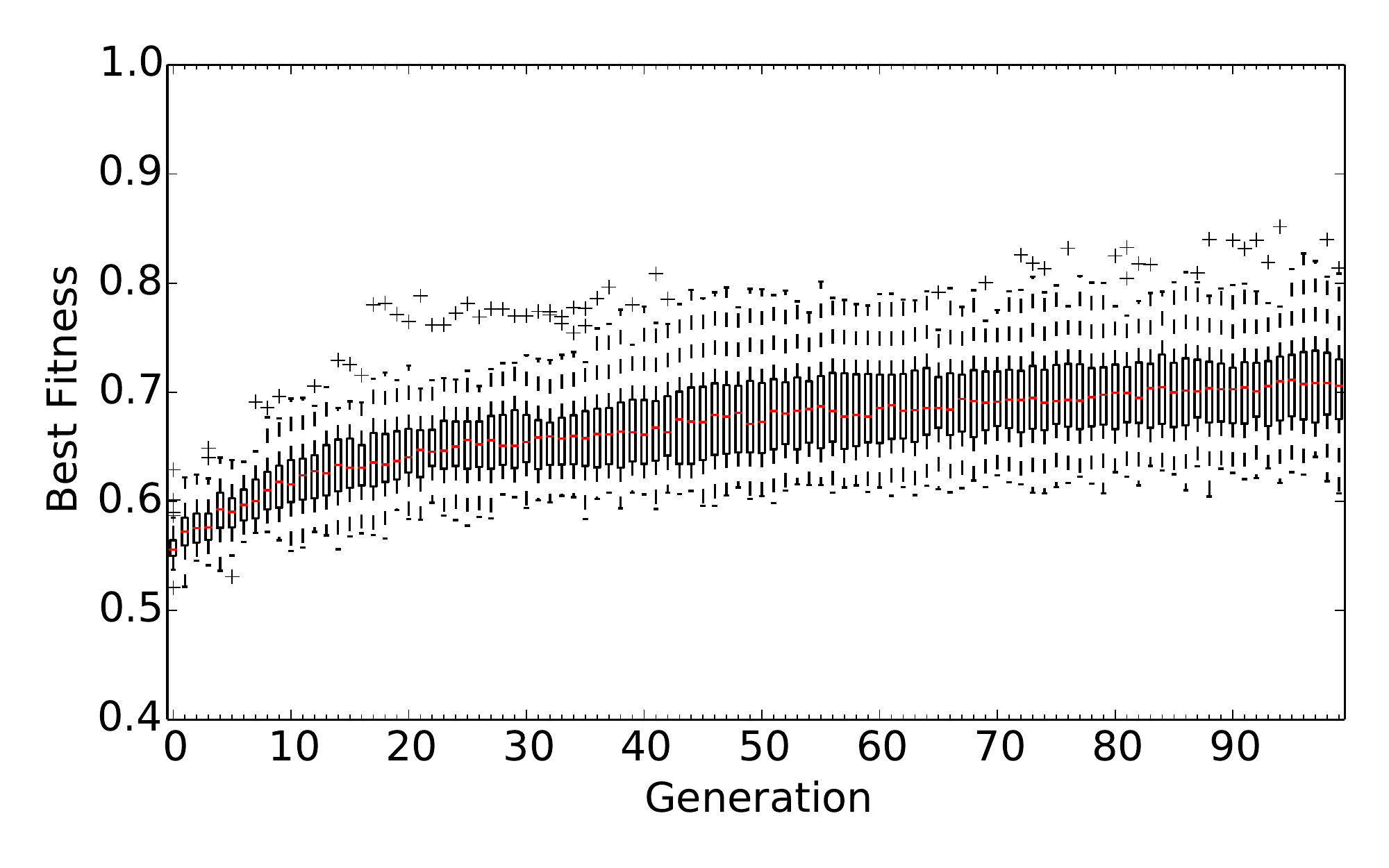}\label{fig:FIG6a}}
        \subfloat[grid size: $20\times20$]{\includegraphics[width=0.4\textwidth]{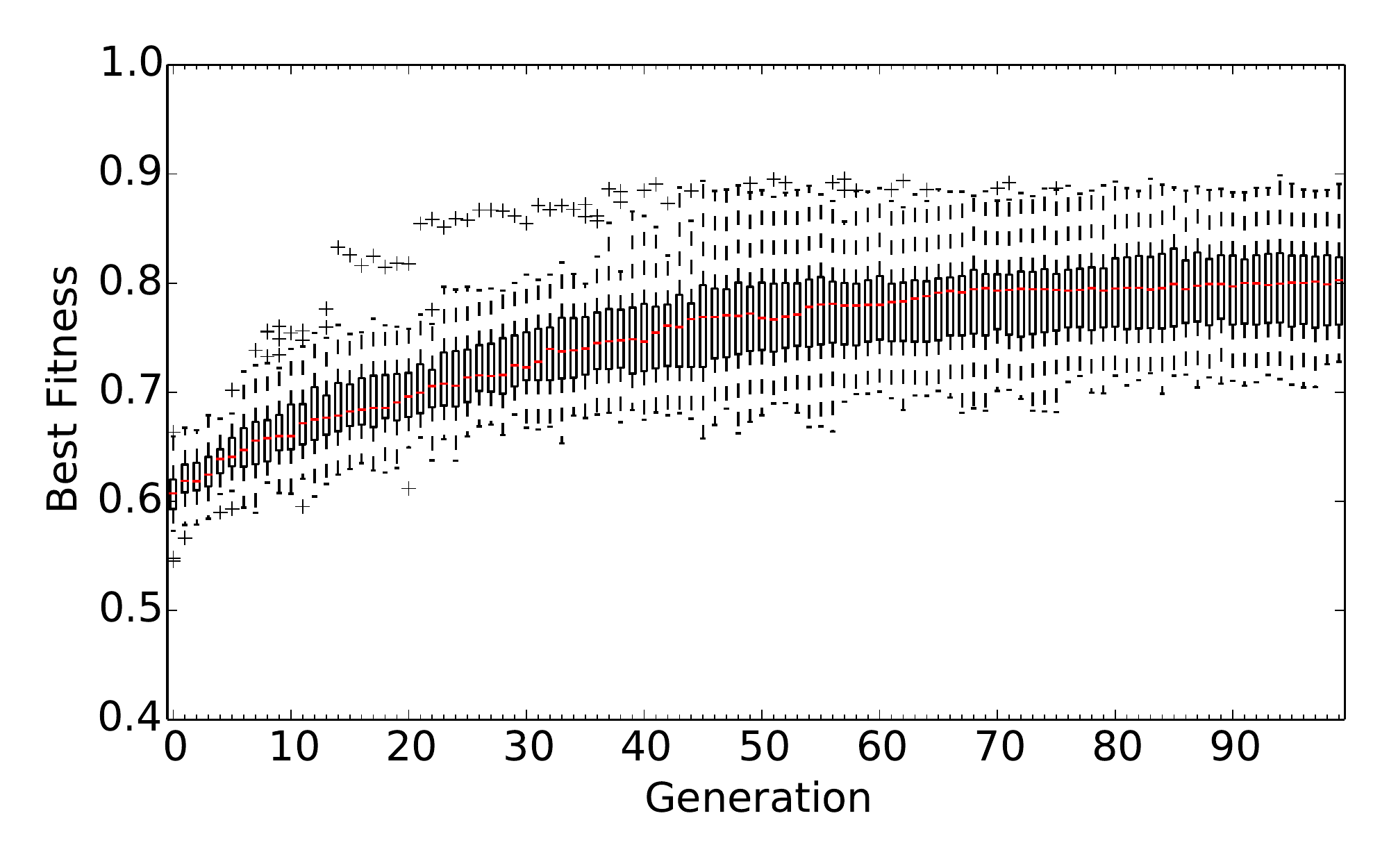}\label{fig:FIG6b}}
    \caption{Best fitness of 50~independent runs. Medians are indicated by the \textit{red bars}. Reprinted by permission from Springer Nature Customer Service Centre GmbH:~\cite{kaiser18}, \copyright Springer Nature Switzerland AG 2019.}
    \label{fig:FIG6}
\end{figure}

Figure~\ref{fig:FIG6} shows the increase of the best fitness (Eq.~\ref{equ:fitness}) over generations for both swarm densities and is representative for the fitness curves observed in all experiments. 
For the experiments on the $15\times 15$ grid, the median best fitness (i.e., the prediction accuracy) of the last generation is $0.71$ and $0.80$ for the $20\times 20$ grid.
This means that the prediction network assesses about $71\%$ and $80\%$, respectively, of the sensor values correctly. 
Thus, the prediction task gets easier for sparse swarm densities. 

\begin{figure}[t]
    \centering
    \subfloat[clustering]{\includegraphics[width=0.25\textwidth]{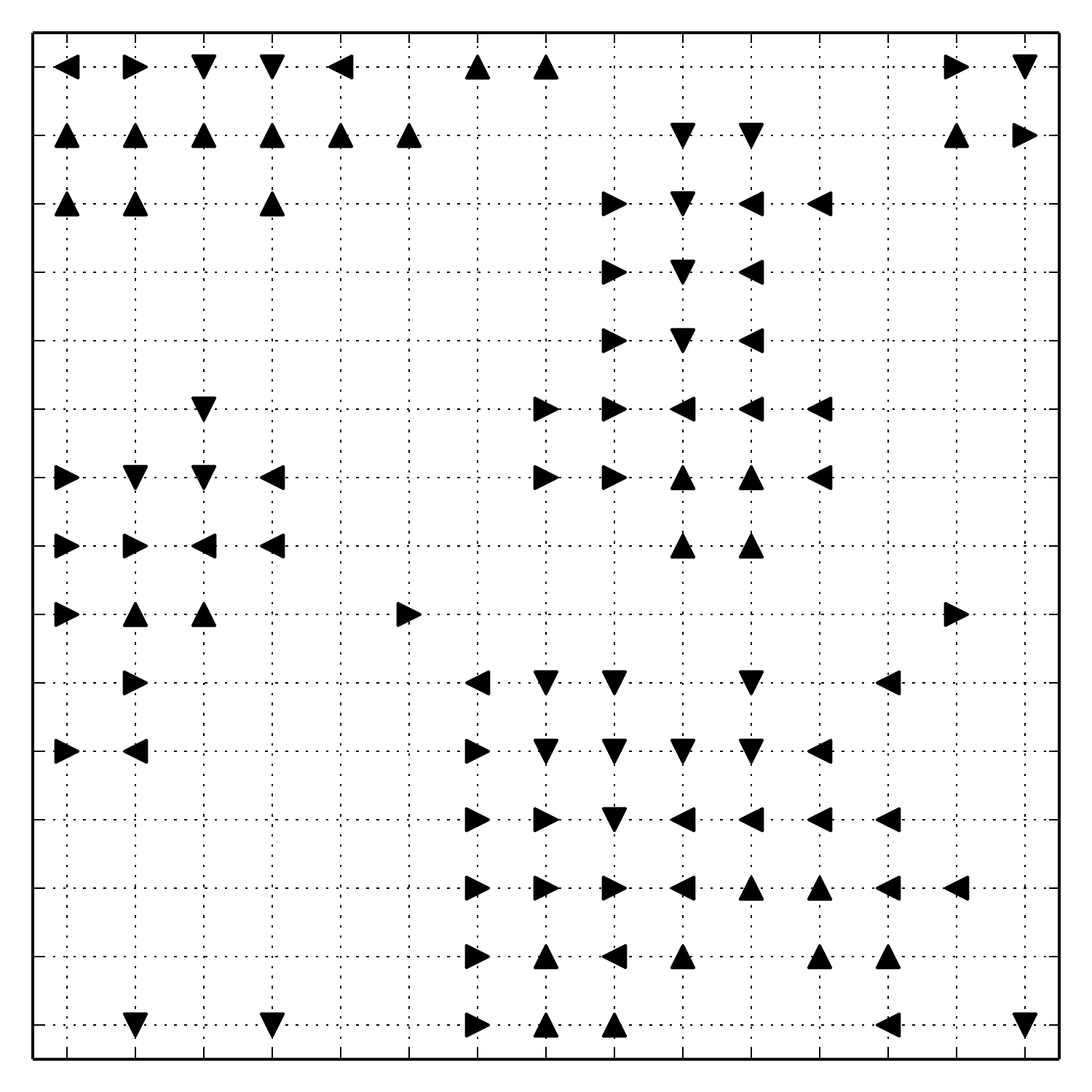}\label{fig:FIG7a}}
    \hspace{3mm}
    \subfloat[aggregation]{\includegraphics[width=0.25\textwidth]{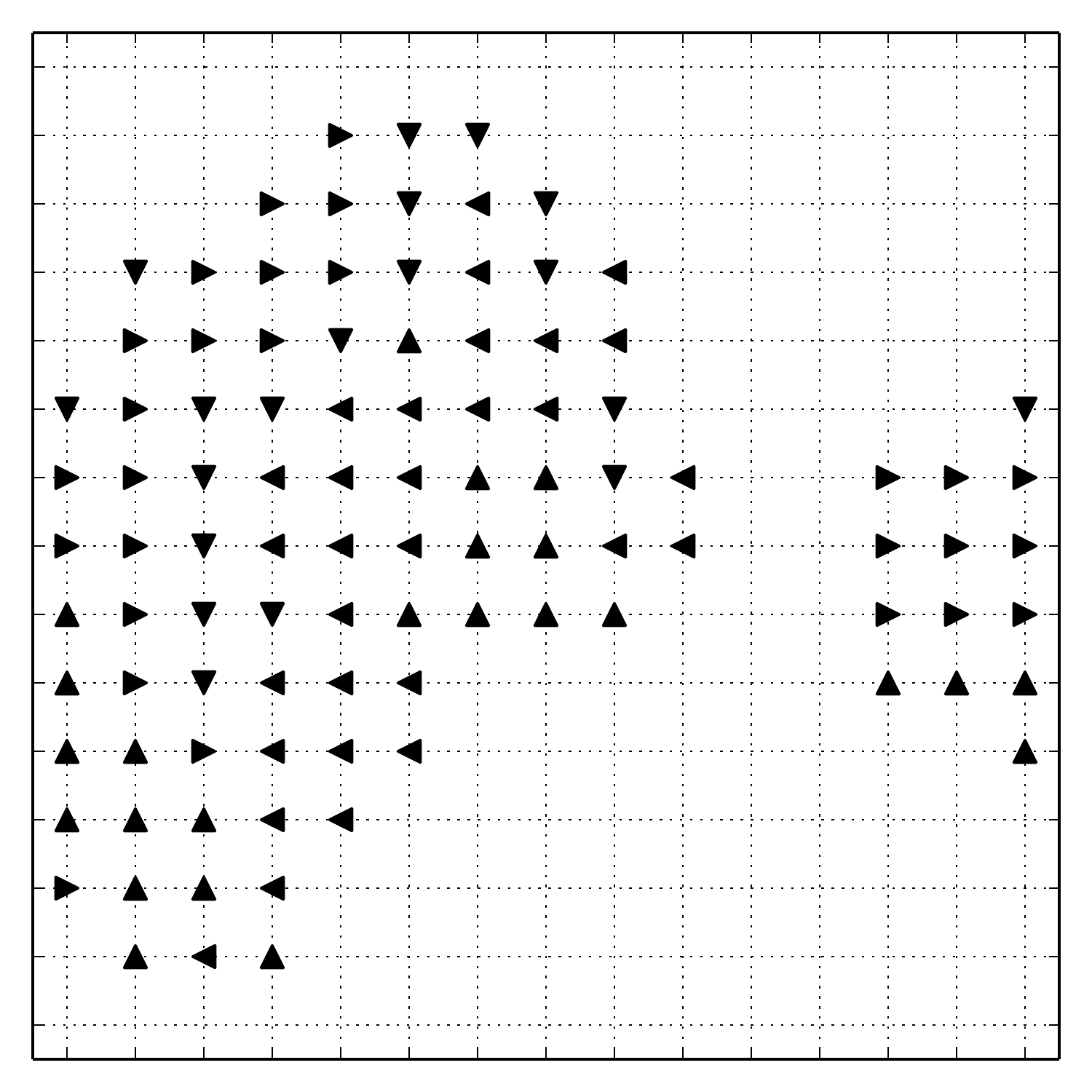}\label{fig:FIG7b}} 
    \hspace{3mm}
    \subfloat[loose grouping]{\includegraphics[width=0.25\textwidth]{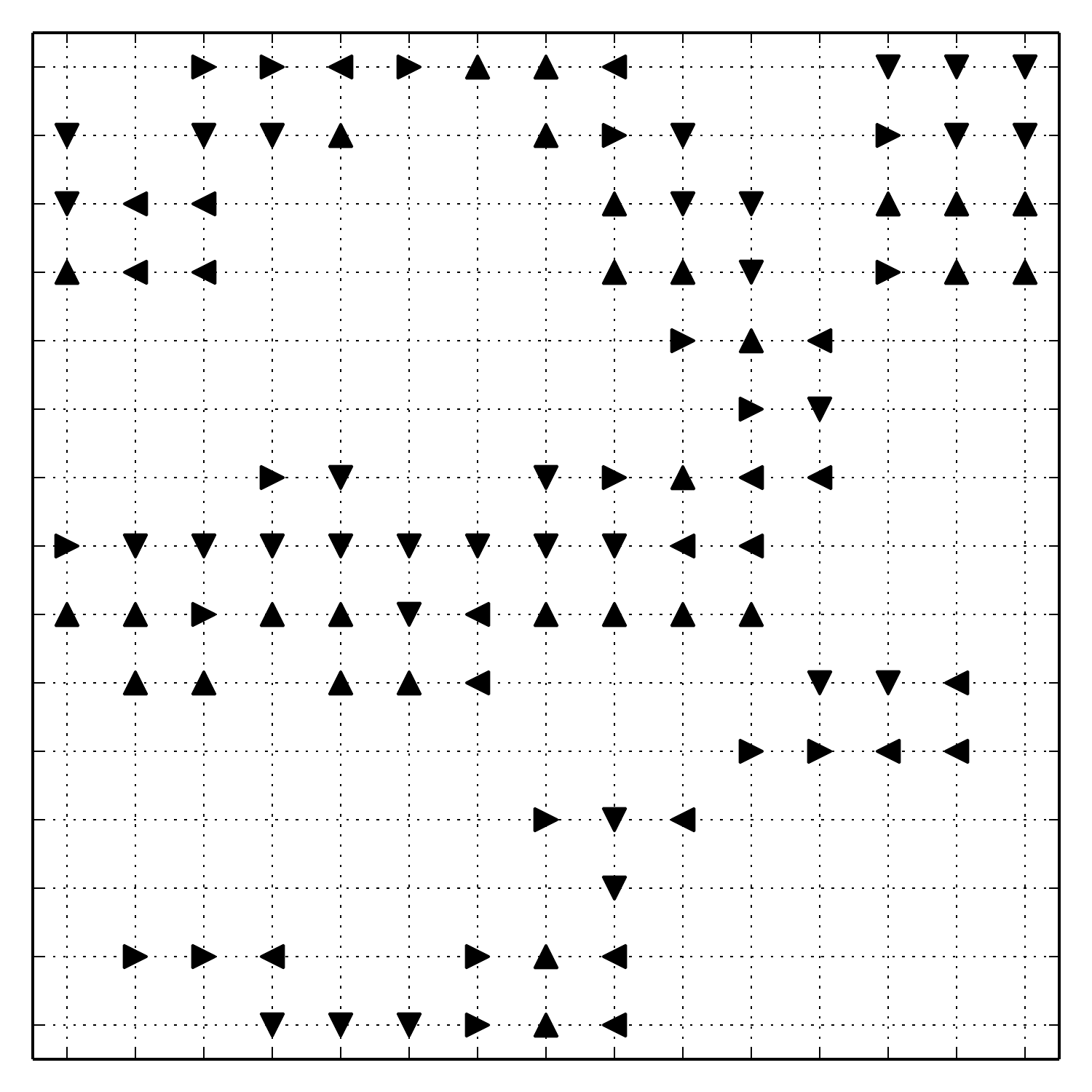}\label{fig:FIG7c}} \\
    \subfloat[lines]{\includegraphics[width=0.25\textwidth]{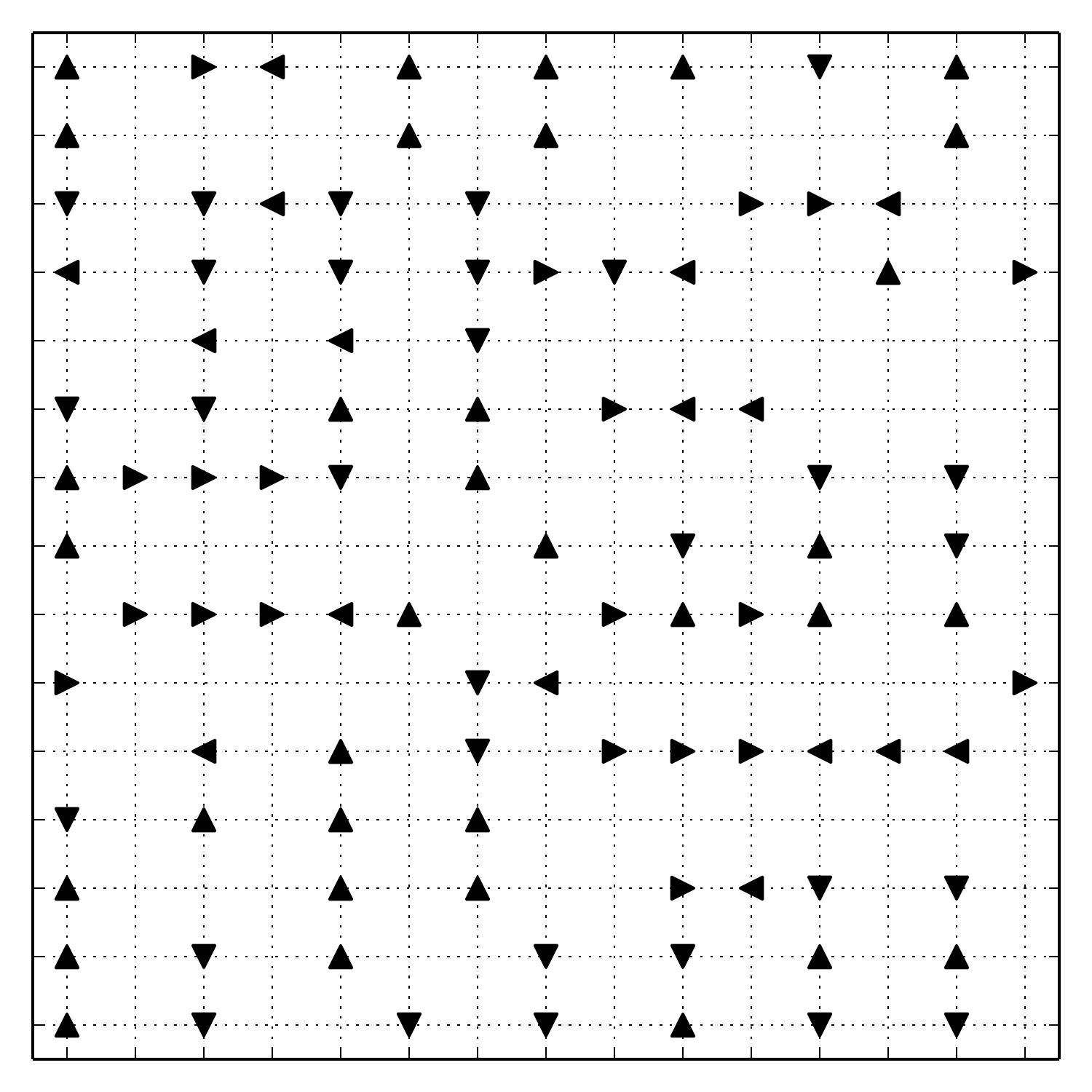}\label{fig:FIG7d}} 
    \hspace{3mm}
    \subfloat[lines]{\includegraphics[width=0.25\textwidth]{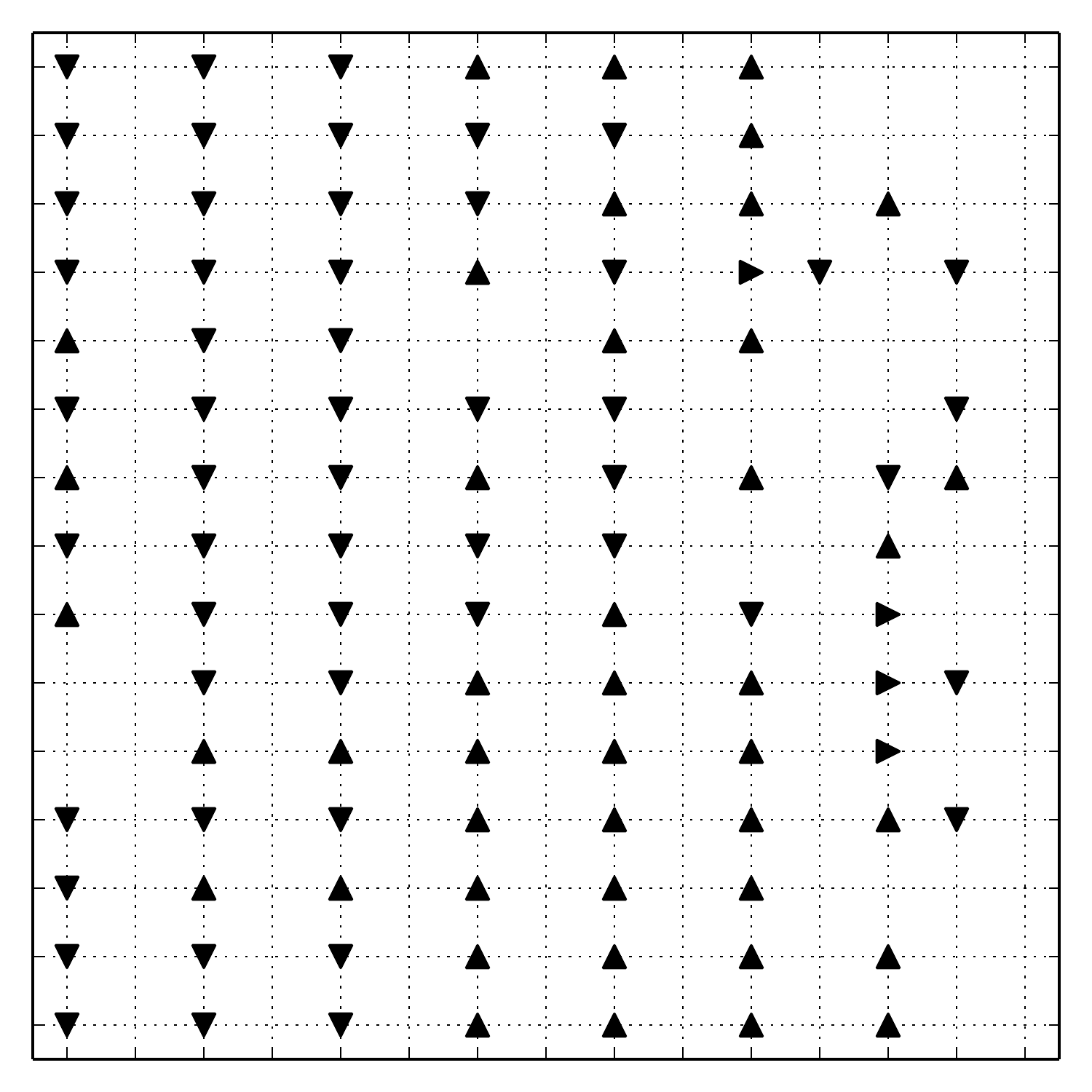}\label{fig:FIG7e}}
    \hspace{3mm}
    \subfloat[triangular lattice]{\includegraphics[width=0.25\textwidth]{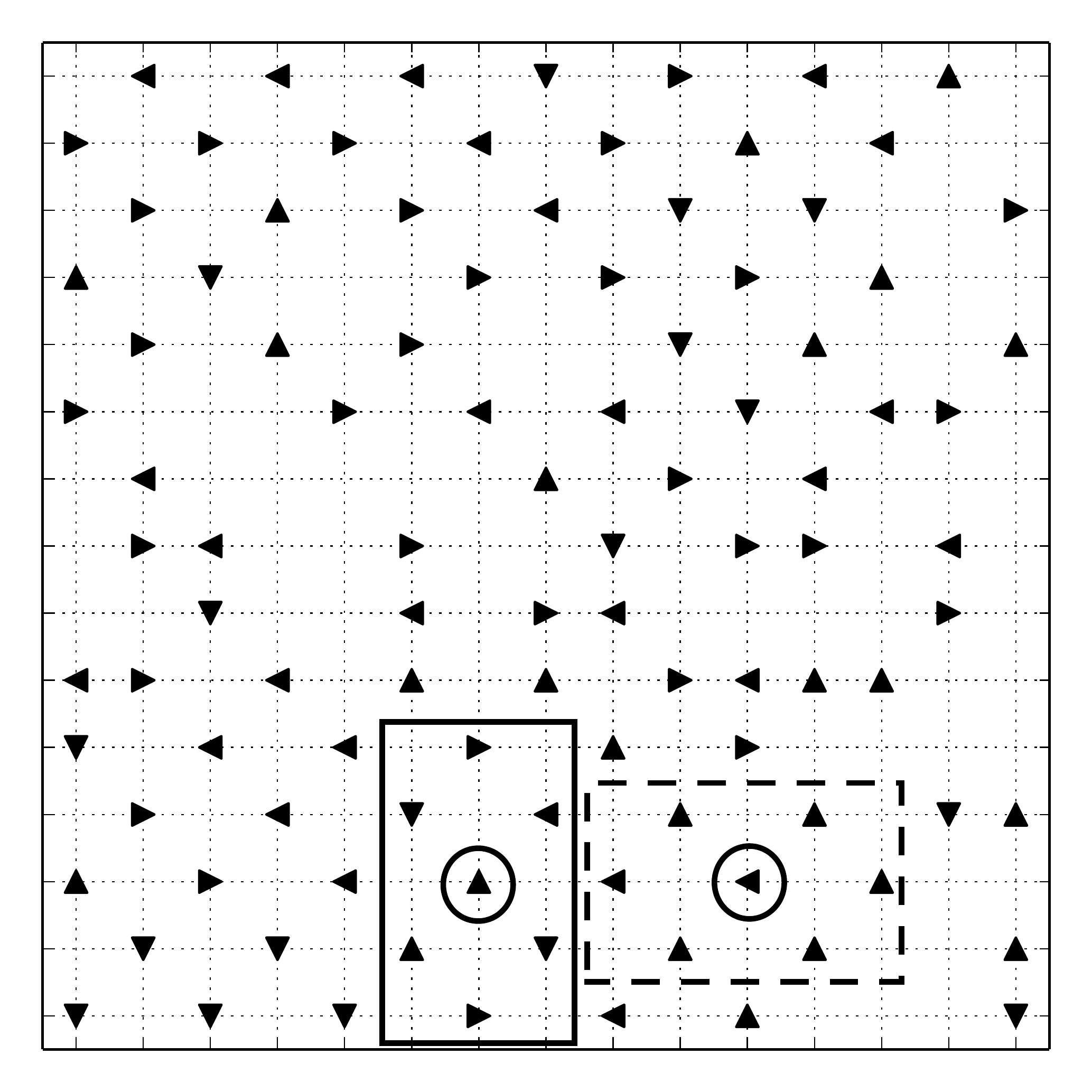}\label{fig:FIG7f}}
    \caption{Resulting structures on a $15\times 15$ grid, the \textit{rectangles} in (f) give the sensor view of the encircled robot as example. The \textit{triangles} give the robots' headings. Reprinted by permission from Springer Nature Customer Service Centre GmbH: \cite{kaiser18}, \copyright Springer Nature Switzerland AG 2019.}
    \label{fig:FIG7}
\end{figure}

\begin{figure}[t]
    \centering
        \subfloat[random dispersion]{\includegraphics[width=0.25\textwidth]{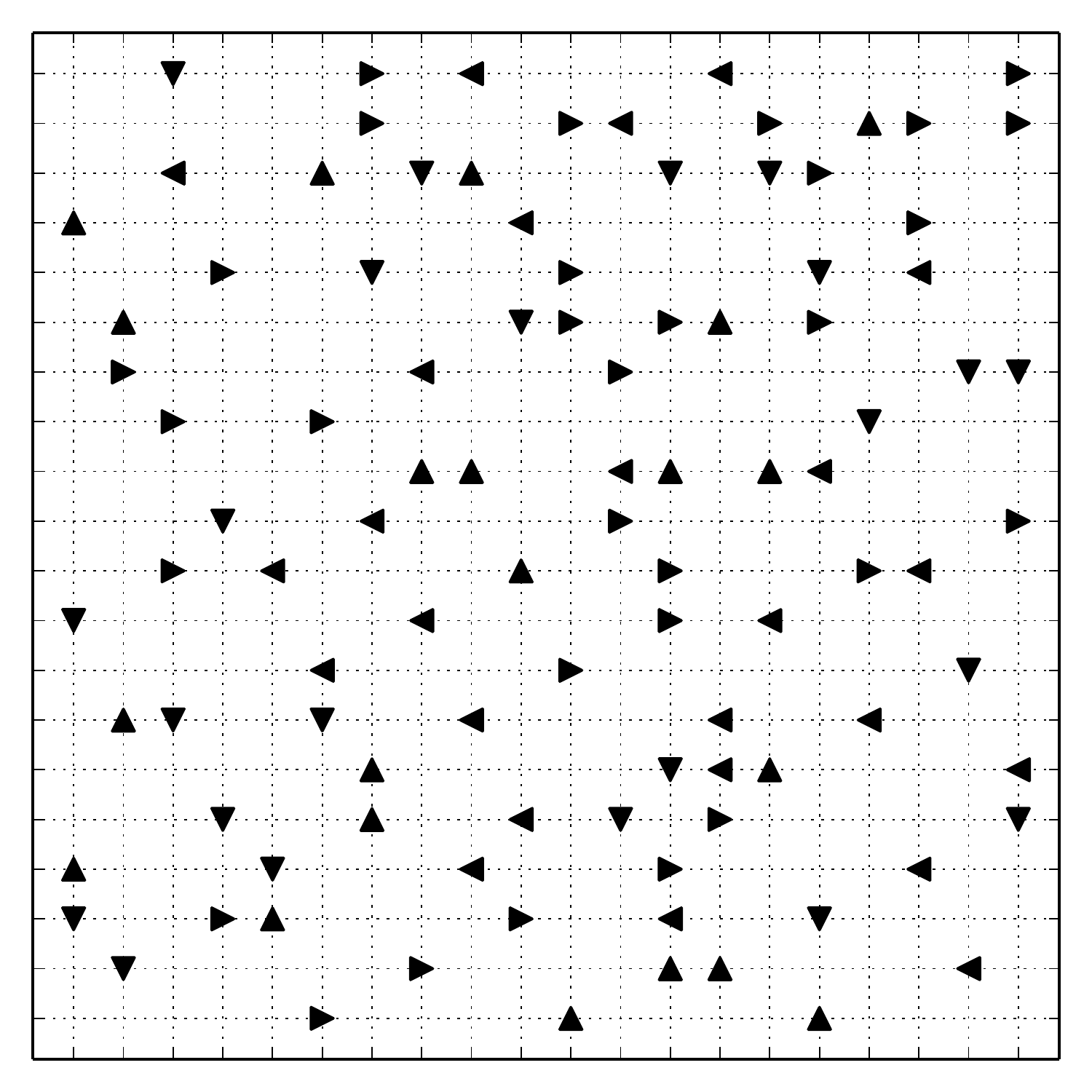}\label{fig:FIG8a}}
        \hspace{3mm}
     \subfloat[random dispersion (squares)]{\includegraphics[width=0.25\textwidth]{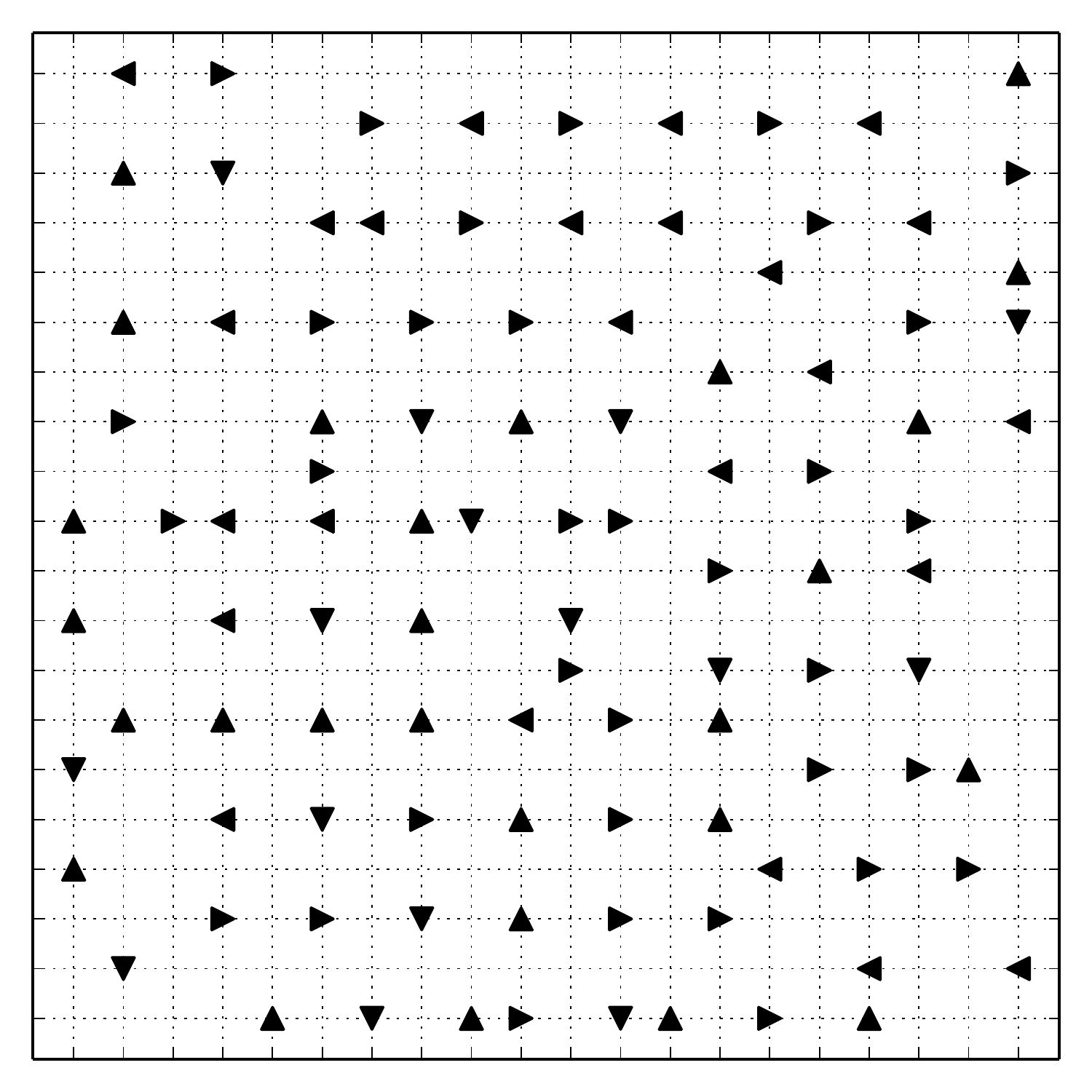}\label{fig:FIG8b}}
      \hspace{3mm}
         \subfloat[random dispersion (triangular lattice)]{\includegraphics[width=0.25\textwidth]{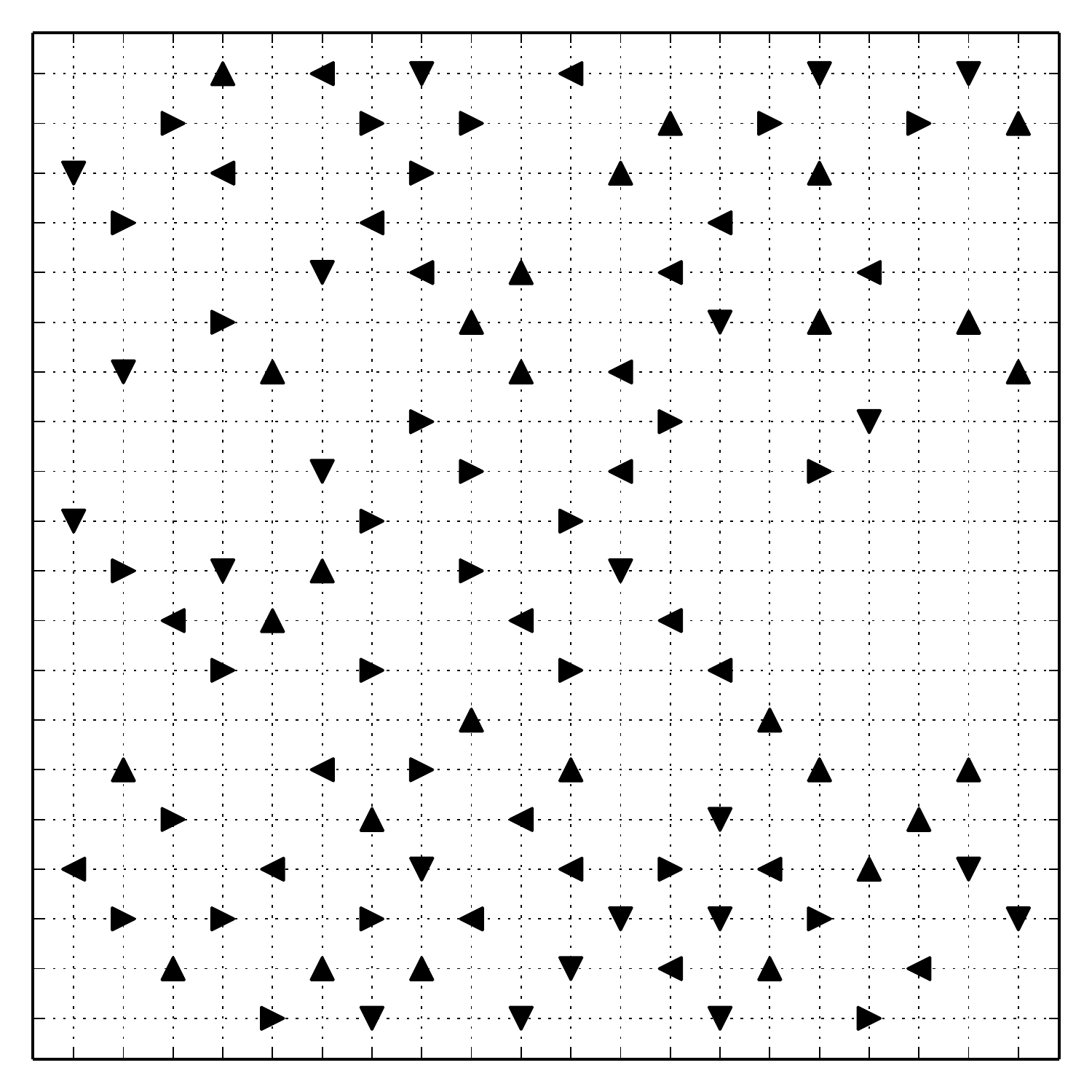}\label{fig:FIG8c}}\\
    \subfloat[lines]{\includegraphics[width=0.25\textwidth]{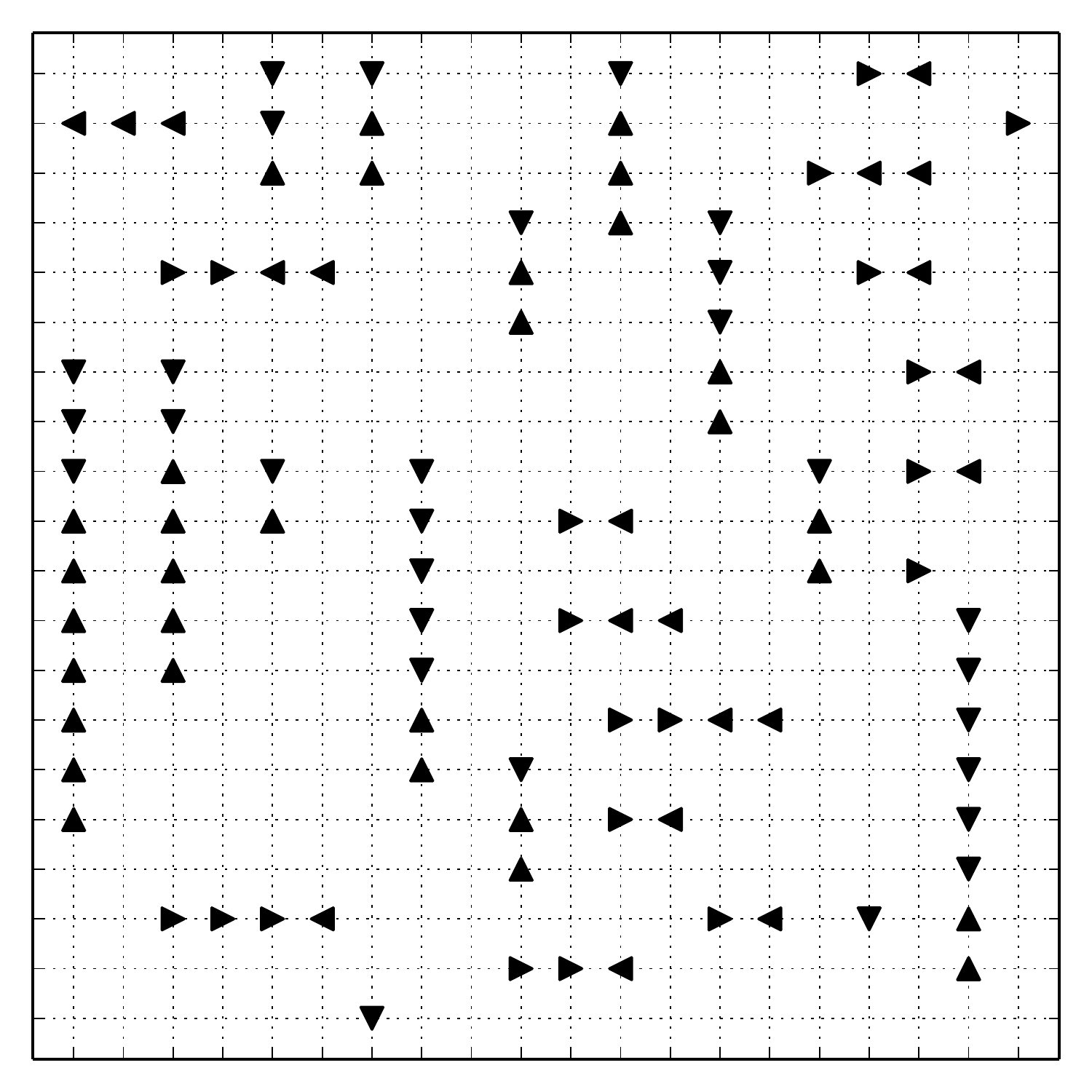}\label{fig:FIG8d}}
     \hspace{2mm}
    \subfloat[pairs]{\includegraphics[width=0.25\textwidth]{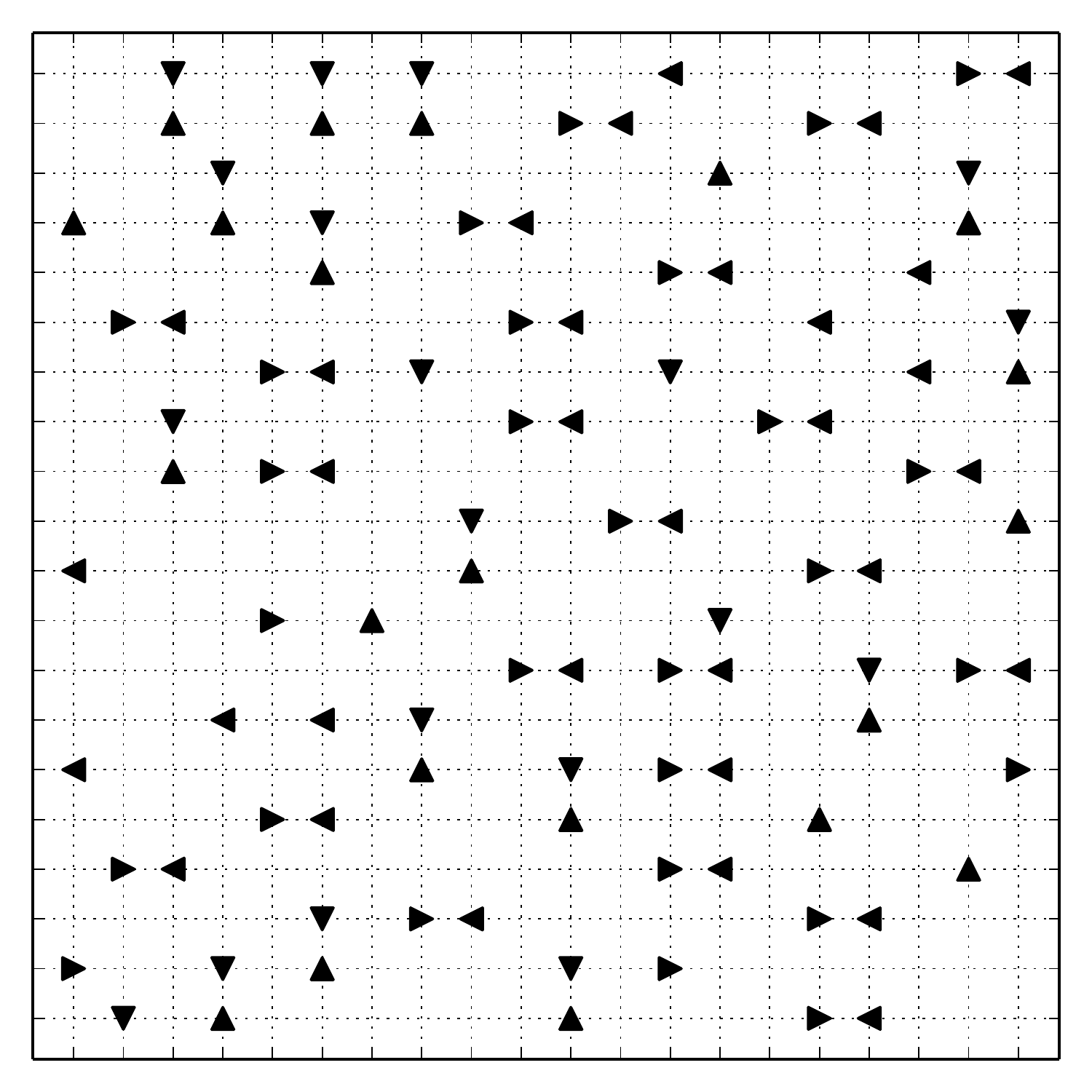}\label{fig:FIG8e}} 
     \hspace{2mm}
    \subfloat[clustering]{\includegraphics[width=0.25\textwidth]{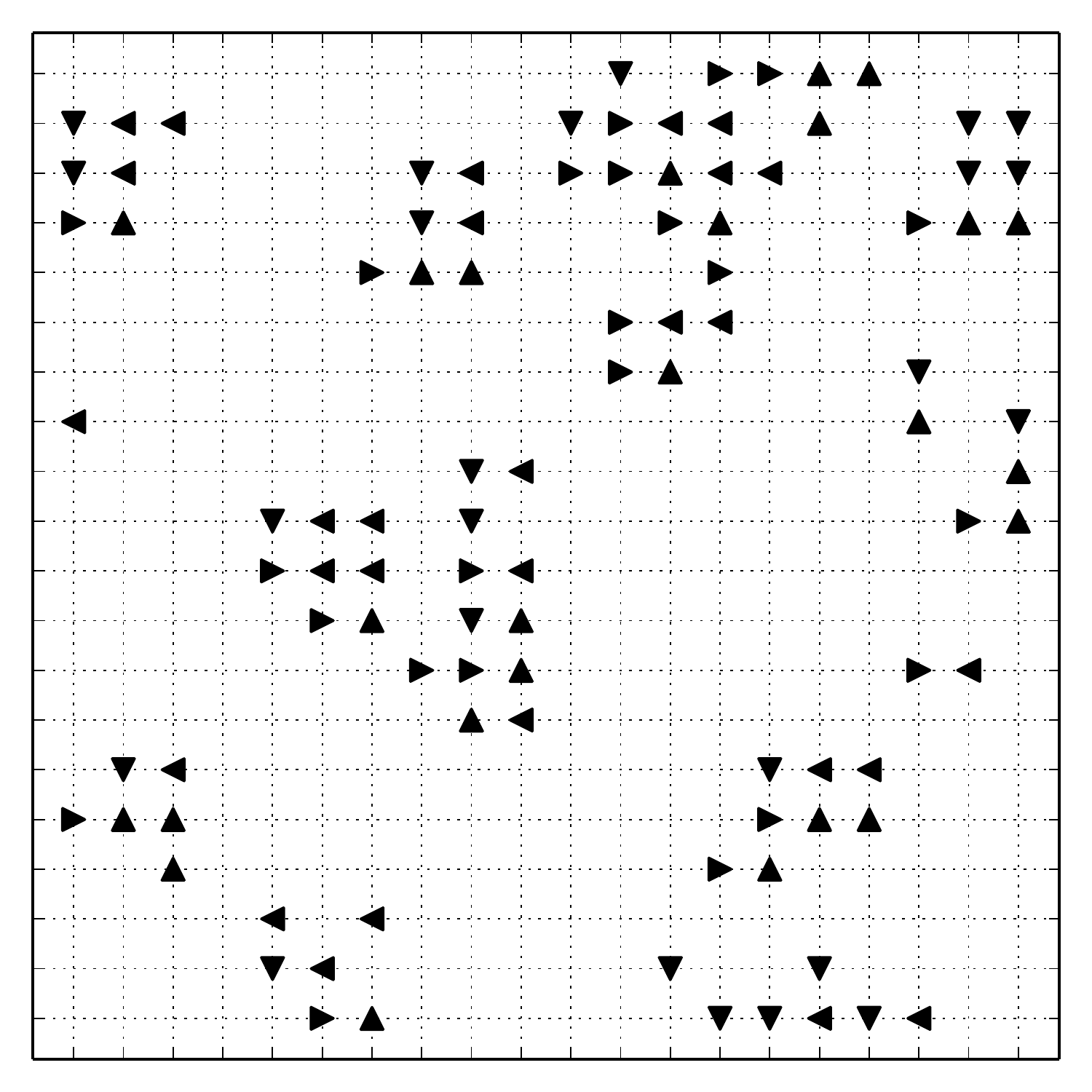}\label{fig:FIG8f}}
    \caption{Resulting structures on a $20\times 20$ grid. The \textit{triangles} give the robots' headings. Reprinted by permission from Springer Nature Customer Service Centre GmbH:~\cite{kaiser18}, \copyright Springer Nature Switzerland AG 2019.}
    \label{fig:FIG8}
\end{figure}

\begin{table}[tph]
       \centering
        \caption{Mean best fitness (Eq.~\ref{equ:fitness}) over 500~time steps of 50~independent runs, percentage of robots within the structure in the last time step, robot movement (Eq.~\ref{equ:movement}), and intended robot movement (Eq.~\ref{equ:intendedmovement}) in the last 113~time steps by emerging structures on a $15 \times 15$ grid. Median values in brackets.\label{tab:TAB4}}
       \pgfplotstabletypeset[normal,
        columns/\space/.style={column name = , 
        column type = l},
        columns/aggre- gation/.style = { column type = C }, 
        columns/clus- tering/.style = { column type = C }, 
        columns/loose grouping/.style = { column type = C }, 
        columns/triang. lattice/.style = {column type = C },
       ]{ %
         \space & cluster & aggre- gation & loose grouping & lines & pairs & triang. lattice  \\ 
         \% of structures & 44 & 12 & 16 & 20 & 2 & 6  \\ \hline
         \% of robots  & 92.8 & 96.5 & 95.5 & 49.7 & 30.0 & 64.7  \\ 
            in structure & (93.0) & (96.5) & (96.5) & (49.0) &  (30.0) & (64.0)  \\ \hline 
		 fitness $F$ & 0.701 & 0.695 & 0.7 & 0.708 & 0.707 & 0.751  \\ 
		  & (0.704) & (0.69) & (0.708) & (0.698)& (0.707) & (0.764)  \\ \hline
		 movement $M$ & 0.008 & 0.01 & 0.003 & 0.019& 0.02  & 0.03 \\ 
		  & (0.0) & (0.0) & (0.0) & (0.012) & (0.02) & (0.029)  \\ \hline
		 intended & 0.911 & 0.865 & 0.901 & 0.849 & 0.816 & 0.05  \\ 
		 movement $I$ & (0.927) & (0.947) & (0.926) & (0.838) & (0.816) & (0.063) \\  
       }
\end{table} 

\begin{table}[tph]
    \caption{Mean best fitness (Eq.~\ref{equ:fitness}) over 500~time steps of 50~independent runs, percentage of robots within the structure in the last time step, robot movement (Eq.~\ref{equ:movement}), and intended robot movement (Eq.~\ref{equ:intendedmovement}) in the last 200~time steps by emerging structures on a $20 \times 20$ grid. Median values in brackets. \label{tab:TAB5}}
       \centering
       \pgfplotstabletypeset[normal,
        columns/\space/.style={column name = , 
       column type = l},
       ]{ %
         \space & lines & pairs & random dispersion & clustering\\ 
         \% of structures & 25 & 27 & 42 & 6 \\ \hline
          \% of robots & 66.3 & 61.57 & 61.6 & 65.3  \\ 
          in structure & (65.0) & (62.0) & (65.0) & (67.0)  \\ \hline 
	 fitness $F$ & 0.833 & 0.821 & 0.779 & 0.734 \\ 
		  & (0.838) & (0.812) & (0.79) &  (0.734)  \\ \hline
		 movement $M$ & 0.017 & 0.043 & 0.074 & 0.021  \\ 
		    & (0.016) & (0.049) & (0.065) & (0.027)  \\ \hline
		 intended & 0.966 & 0.886 & 0.128 &  0.98  \\ 
		 movement $I$ & (0.978) & (0.905) & (0.125) & (0.979)  \\  
       }
\end{table}

We classify various emergent structures in our experiments, see Figs.~\ref{fig:FIG7} and~\ref{fig:FIG8}, based on the metrics defined in Sec.~\ref{sec:classification}. 
For higher swarm densities (smaller grid), mostly grouping behaviors evolve, see Table~\ref{tab:TAB4} and Fig.~\ref{fig:FIG7}, while for low densities (larger grid) many randomly dispersed structures emerge, see Table~\ref{tab:TAB5} and Fig.~\ref{fig:FIG8}. 
On the $15 \times 15$ grid, nearly all robots assemble into the grouping structures while the number is lower in lines, pairs and triangular lattices (see Table~\ref{tab:TAB4}). 
On the larger grid, the median percentage of robots assembling into line structures is higher than on the smaller grid. 
Considering the different resulting structures, more than $61\%$ of robots assemble on average into the structure on the $20\times 20$ grid (see Table~\ref{tab:TAB5}).

On the $15\times 15$ grid, emerging line structures span over the whole torus in $20\%$ of the runs in which lines form while this is not observed on the $20 \times 20$ grid. 
Triangular lattices solely emerge on the smaller grid, as the swarm density on the larger grid is too small to allow the formation of a repetitive triangular pattern over most of the grid. 
Nevertheless, a tendency towards triangular lattices is visually observable in $6\%$ of the runs, as illustrated in Fig.~\ref{fig:FIG8c}. 
Similarly, the square pattern can be visually observed in $6\%$ of the runs as shown in Fig.~\ref{fig:FIG8b} with a median of $19\%$ of robots assembling into the structure. 
All of those runs are classified as random dispersion. 
In summary, the swarm density has a strong influence on the emerging structures. 

A comparison of mean robot movement~(Eq.~\ref{equ:movement}), mean intended robot movement~(Eq.~\ref{equ:intendedmovement}), and mean fitness~(Eq.~\ref{equ:fitness}, prediction accuracy) of the best evolved individuals enables us to identify differences in behavioral characteristics, see Tables~\ref{tab:TAB4} and~\ref{tab:TAB5}.
We observe a similar mean fitness for all emerging structures, whereby it is slightly higher in the experiments on the larger grid.
In clustering, aggregation, loose grouping, line structures and pairs the intended movement~$I$ is high while the robot movement~$M$ is low. 
Robots keep their positions and headings if the grid cell in front is occupied and they intend to move straight, which happens frequently in those structures. 
Thus, the robots' sensor values are mostly constant and can be easily predicted. 
Robots favor rotations in triangular lattices and random dispersion, that is, $I$ and~$M$ are low.
In triangular lattices, each robot has the same sensory perception in every orientation as illustrated in Fig.~\ref{fig:FIG7f}. 
Thus, robots can keep their sensor values constant by turning. 
This means means that these behaviors are evolutionary stable, too.
Overall, the robots' movement is low for all structures in the last $\tau$ time steps. 
This is supported by the temperature curve, see Fig.~\ref{fig:FIG5}, which shows that almost all robots stay stopped after 100~time steps in both swarm densities. 

\begin{figure}[tph]
    \centering
    \subfloat[aggregation (beh. in Fig~\ref{fig:FIG7b})]{\includegraphics[width=0.2\textwidth]{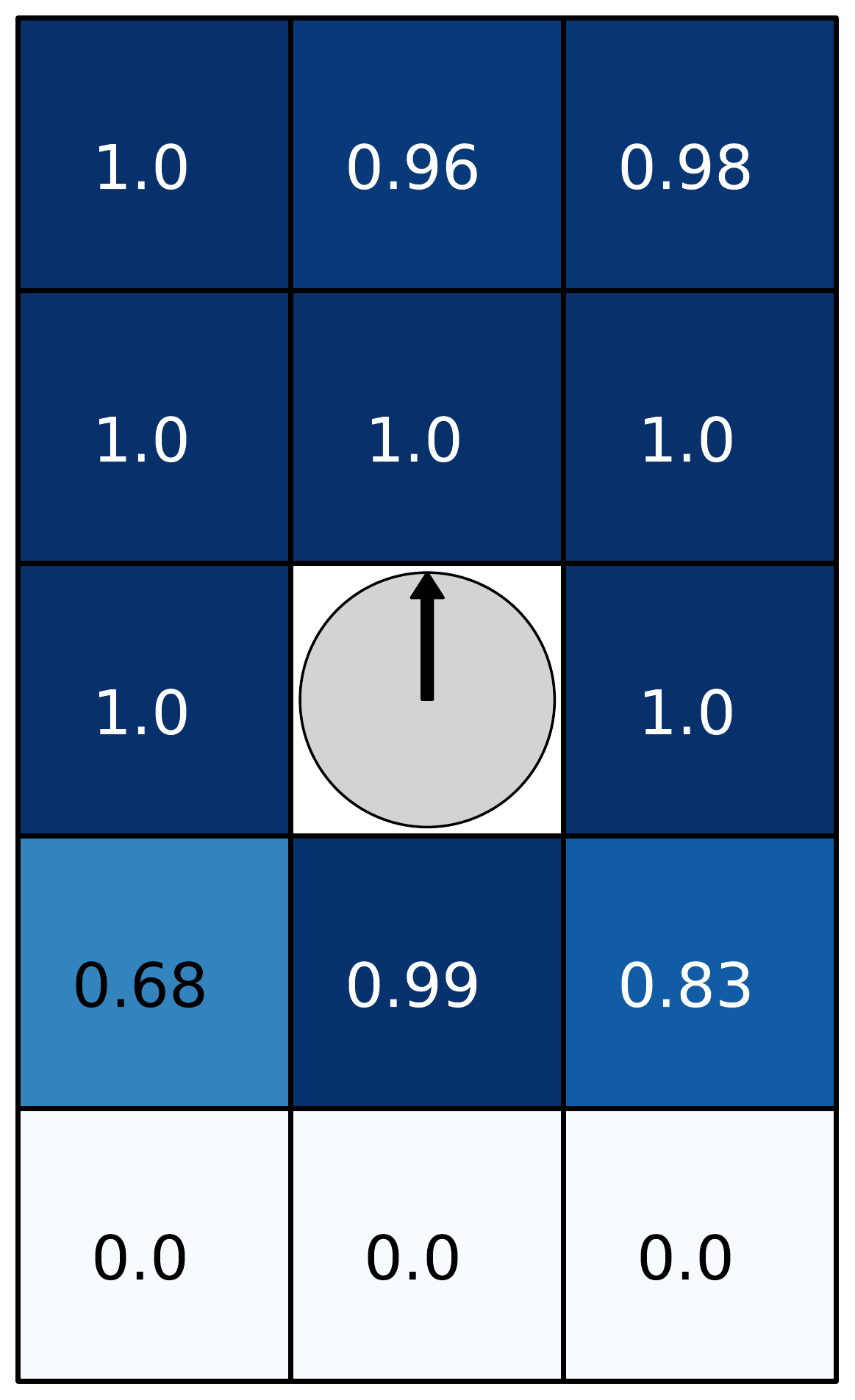}\label{fig:FIG9a}}
    \hspace{5mm}
     \subfloat[random dispersion (beh. in Fig.~\ref{fig:FIG8b})]{\includegraphics[width=0.2\textwidth]{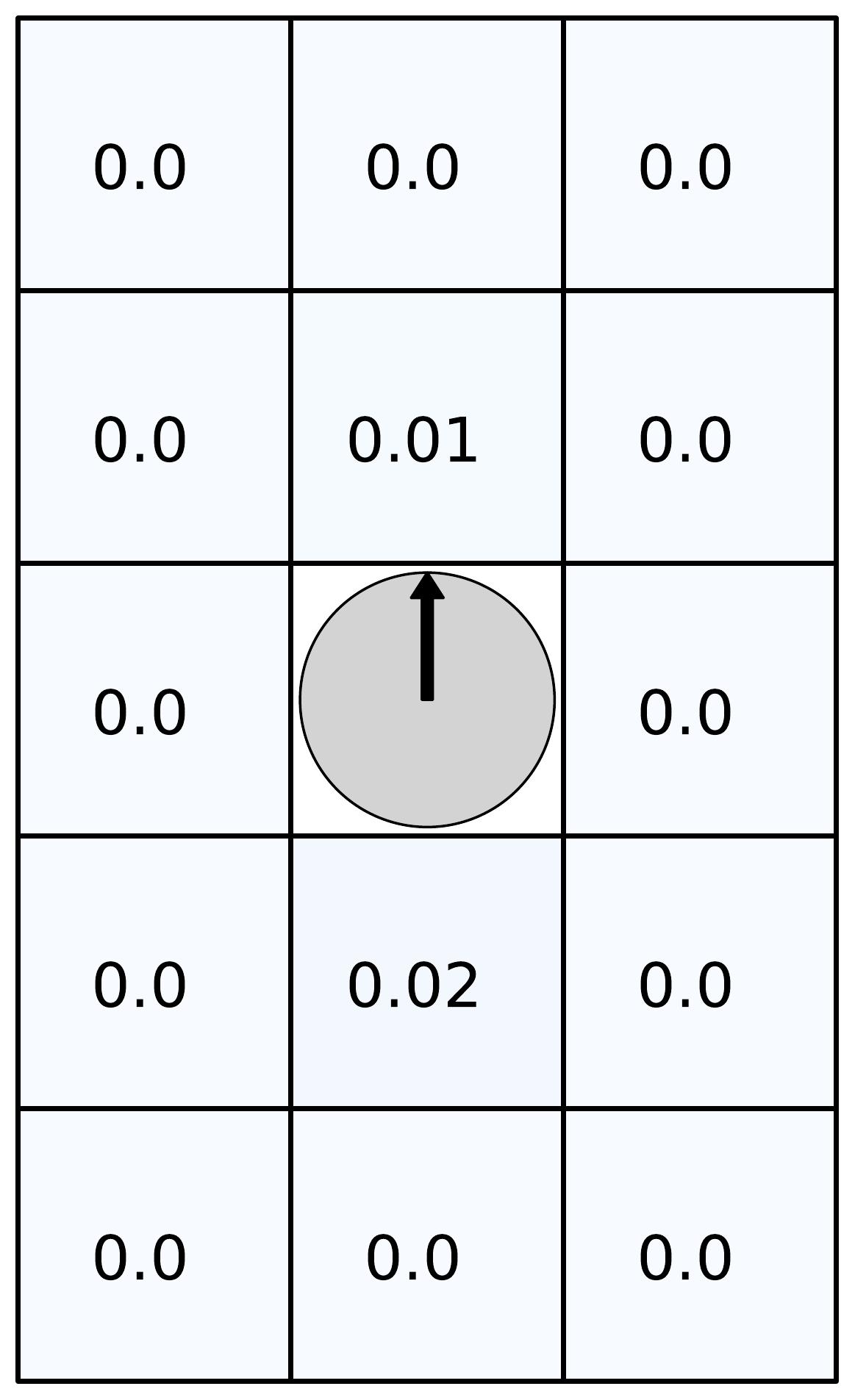}\label{fig:FIG9b}} 
        \hspace{5mm} 
    \subfloat[lines (behavior in Fig.~\ref{fig:FIG7e})]{\includegraphics[width=0.2\textwidth]{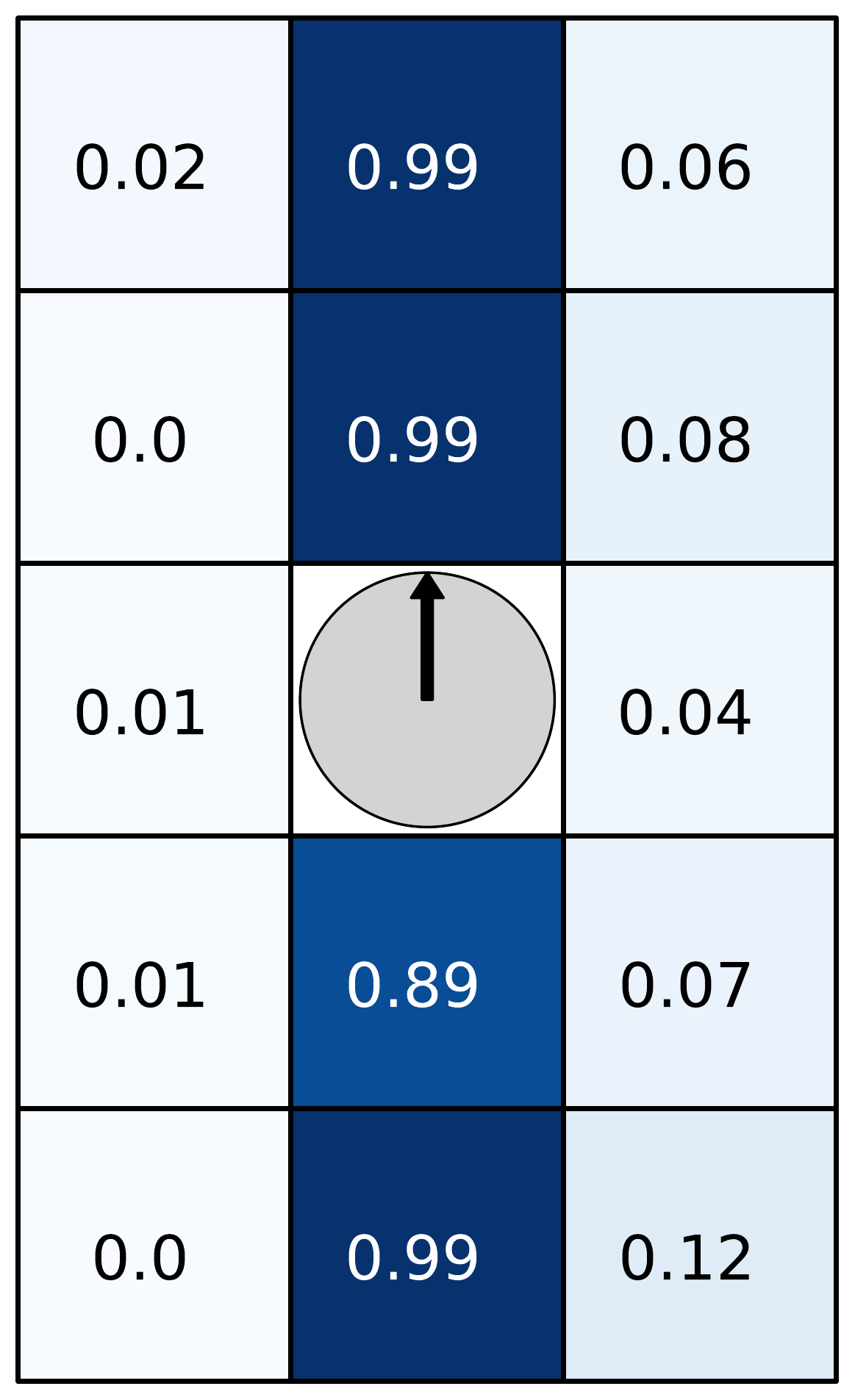}\label{fig:FIG9c}}
        \hspace{5mm}
     \subfloat[triangular lattice (beh. Fig.~\ref{fig:FIG7f})]{\includegraphics[width=0.2\textwidth]{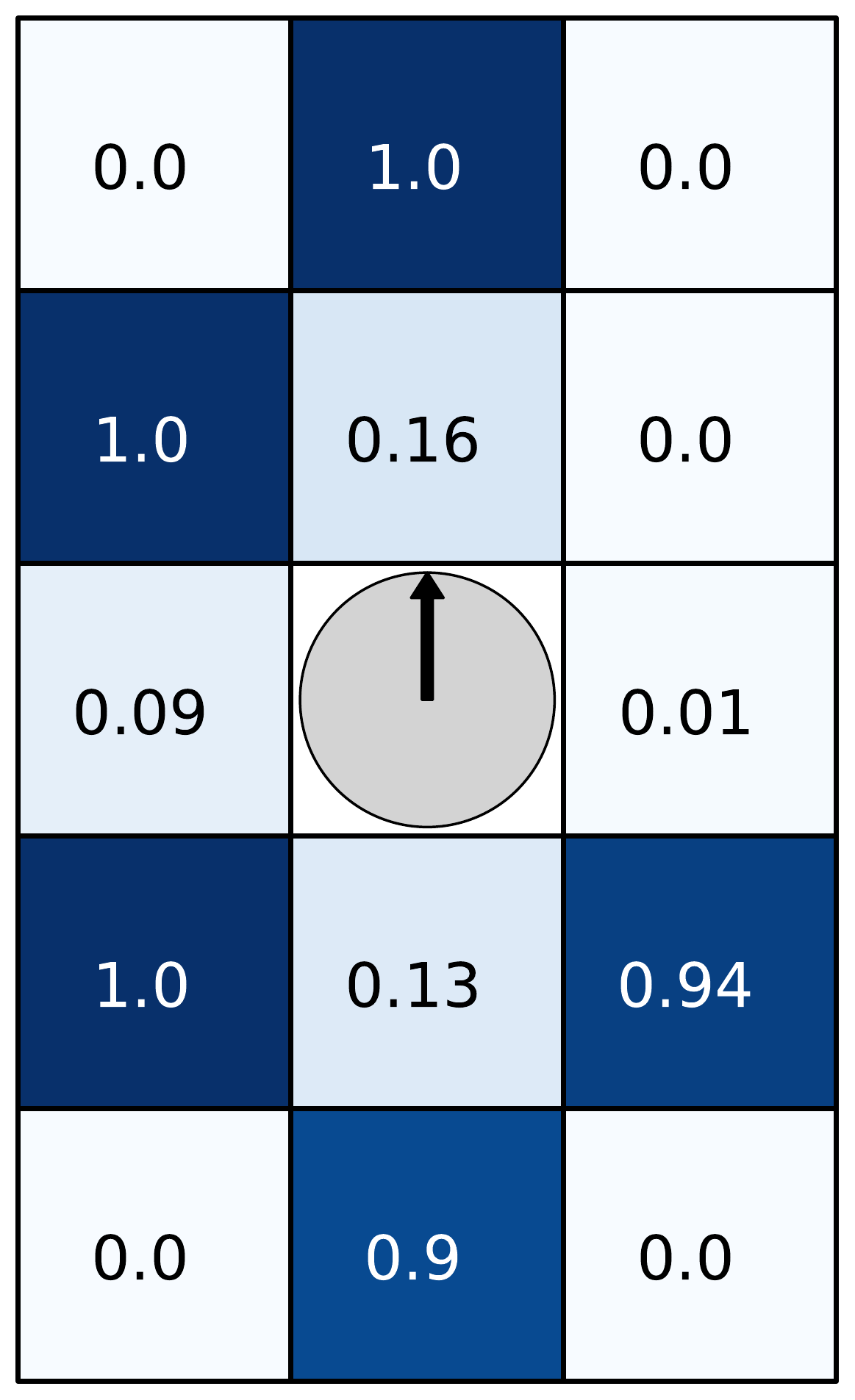}\label{fig:FIG9d}}
    \caption{Average sensor predictions of all 100~robots over 500~time steps for resulting behaviors shown in Figs.~\ref{fig:FIG7} and \ref{fig:FIG8}. Robots are represented by \textit{gray circles}, \textit{black arrows} indicate their headings. Reprinted by permission from Springer Nature Customer Service Centre GmbH:~\cite{kaiser18}, \copyright Springer Nature Switzerland AG 2019.}
    \label{fig:FIG9}
\end{figure}

We aim to find the relationship between the world model of the robot and the emergent structures. 
Therefore, we assess the average sensor predictions during the evolutionary run of the best individuals which correspond to the robot's anticipated environment.  
In grouping behaviors like aggregation (see Fig.~\ref{fig:FIG9a}), clustering and loose grouping, robots predict that the majority of their adjacent grid cells are occupied.
In contrast, none of the adjacent grid cells are predicted to be occupied when forming random dispersion, see Fig.~\ref{fig:FIG9b}. 
Robots predict that the grid cells in front of and behind them are occupied for line structures as shown in Fig.~\ref{fig:FIG9c}. 
Forming pairs, a neighbor is only expected to be sensed on the grid cell directly in front of a robot. 
The sensor predictions in triangular lattices follow this intuitive scheme, too, and match almost completely the visually observed robot structure as shown in Fig.~\ref{fig:FIG9d}. 
Thus, we find that the robots' world models (i.e. their average sensor predictions) and the observed structures coincide closely.

\subsection{Engineering Self-Organized Self-Assembly} \label{sec:SOSA}

\subsubsection{Predefining Sensor Predictions} \label{sec:PRED}
A strength of our minimal-surprise approach is the emergence of non-trivial swarm behaviors, despite selective pressure to minimize the prediction error. 
When running the approach with complete freedom as in Sec.~\ref{sec:Adaptation}, we have no influence on the resulting behavior or the forming patterns. 
Therefore, we show in a next step that the evolutionary process can be manipulated to push towards the emergence of desired behaviors and targeted patterns.
We predefine several or even all of the robots' sensor value predictions by fixing some or all outputs $p_0,\dots,p_{R-1}$ of the prediction network to desired values. 
High fitness is then achieved by good predictions of the unfixed outputs~$p_r$ (if any) and by appropriate behaviors.
We measure the success by investigating the impact on the resulting structures.
In our initial experiments, we keep the swarm size of $N = 100$ and the grid side lengths of $L = \{ 15, 20 \}$.
Without loss of generality, we aim for the emergence of lines. 
A~video is available online\footnote{\textit{Video2.mp4} -  \url{https://doi.org/10.5281/zenodo.3362285}} and shows the self-assembly behaviors. 
As in the previous section, we use the metrics defined in Sec.~\ref{sec:classification} to classify the resulting structures.

We partially predefine prediction values in the first step.
We set the predictions of the sensors in front and behind the robot to~1 (i.e., $S_0=S_3=S_8=S_{11}=1$, cf. Fig.~\ref{fig:FIG4}) while the other ten remaining sensors still need to be predicted by the prediction network.
Thus, the prediction network has 15~input neurons, 12~hidden neurons, and ten output neurons. 
To obtain a converging fitness curve, the mutation rate was adjusted to $0.3$ for the runs on the $20\times 20$ grid.
For the $15 \times 15$ grid, the median best fitness (Eq.~\ref{equ:fitness}) of the last generation is $0.72$ and for the $20 \times 20$ grid $0.78$. 

We notice a decrease in the variety of resulting structures by partially predefining sensor predictions.
The best evolved individuals lead to clustering, aggregation and loose grouping in $64\%$ and to the formation of lines in $36\%$ of the runs on the $15 \times 15$ grid. 
Thus, we observe an increase of $16$ percentage points~(pp) in the formation of line structures and a decrease of $8$~pp in grouping behaviors compared to running the approach without predefining any sensor predictions. 
Pairs and triangular lattices do not emerge anymore. 
On the $20 \times 20$ grid, robots cluster in $8\%$, form pairs in $2\%$ and form lines in $90\%$ of the runs. 
Consequently, we notice an increase of $2$~pp in clustering and of $65$~pp in the formation of lines.  
Pairs form only in one run and no random dispersion as well as no triangular lattices emerge as a consequence of predefining the sensors in front and behind the robot to~1. 
The average sensor value predictions of these structures deviate at least in two of the four predefined values, cf. Fig~\ref{fig:FIG9}.
Grouping behaviors can easily emerge in this scenario as a robot requires more than four neighbors for these structures and thus, just predicts more cells to be occupied. 
We observe that in the emergent grouping behaviors the average sensor predictions of all robots are above $0.5$ for at least three of the ten sensors which are still predicted by the prediction network on the $15\times 15$ grid and for at least four sensors on the $20\times 20$ grid, respectively.
An average sensor prediction above $0.5$ states that a grid cell is predicted to be occupied in at least half of the time steps by all robots. 
Similarly, for line structures on the $15\times 15$ grid maximally two of the sensors and, respectively, three sensors on the $20\times 20$ are on average predicted above $0.5$ as the predefined predictions already match the structure.  

Overall, the emergence of lines increases and thus, we successfully engineered an influence on the emerging structures. 
However, we cannot avoid a dependence on the environment (robot density). 
We observe lines spanning over the whole torus in $67\%$ of the runs resulting into line structures on the smaller grid while none are observed on the larger grid as before. 

Next, we predefine all sensor predictions of the robots while still aiming for the emergence of line structures. 
In consequence, the prediction network is not evolved anymore. 
We keep the fitness as defined in Eq.~\ref{equ:fitness} and thus, measure prediction accuracy. 
We rely exclusively on the action network being subject to genetic drift as it does not generate fitness by itself.
The fitness value is still determined by comparing the predefined predictions with the actual sensor values.  
As before, we predefine the sensor predictions in front and behind the robot to~1 while we set all other sensor value predictions to~0 now. 
The median best fitness of the last generation is $0.85$ on the $15 \times 15$~grid and $0.88$ on the $20 \times 20$~grid.

\begin{figure}[t]
    \centering
    \subfloat[grid size: $15 \times 15$]{\includegraphics[width=0.25\textwidth]{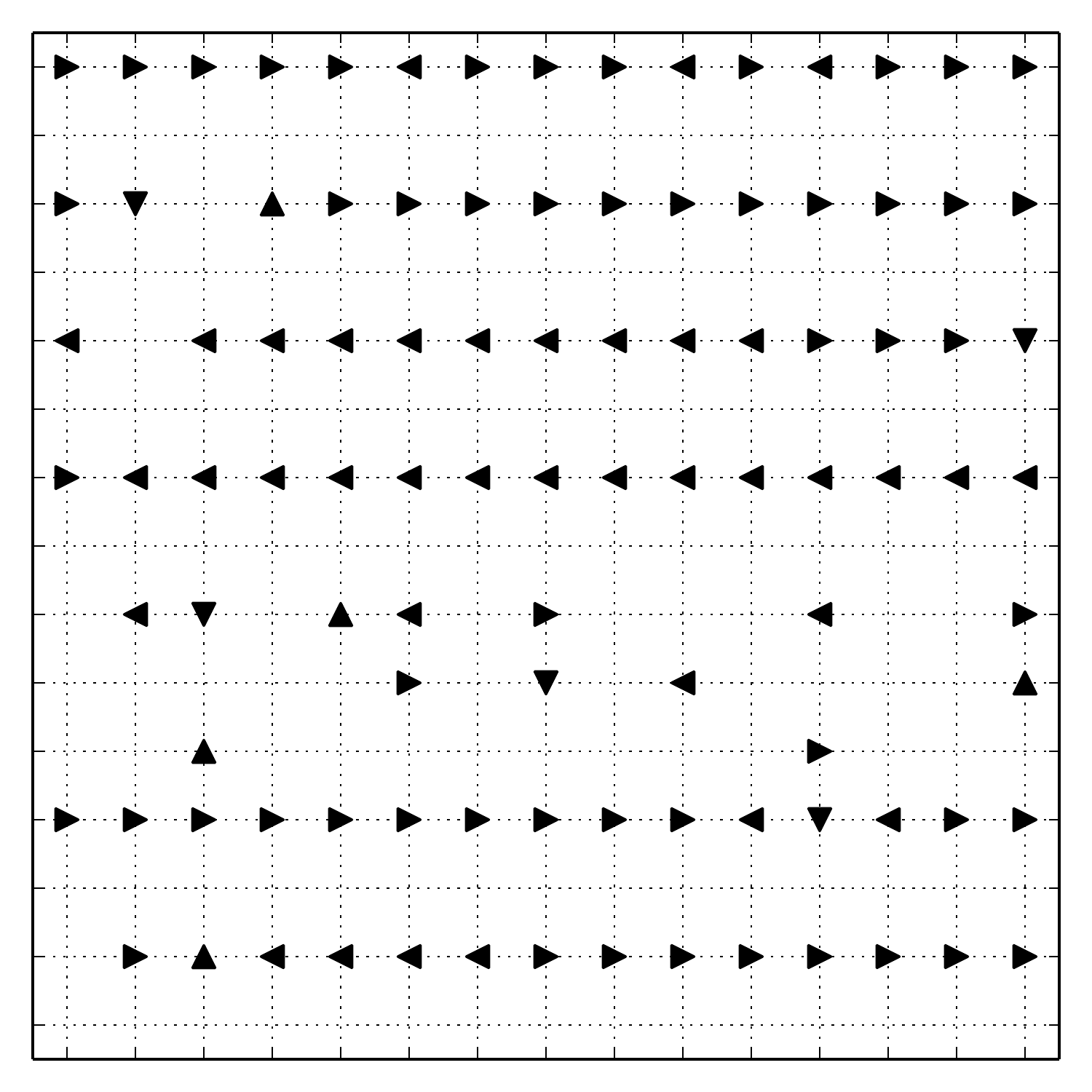}\includegraphics[width=0.25\textwidth]{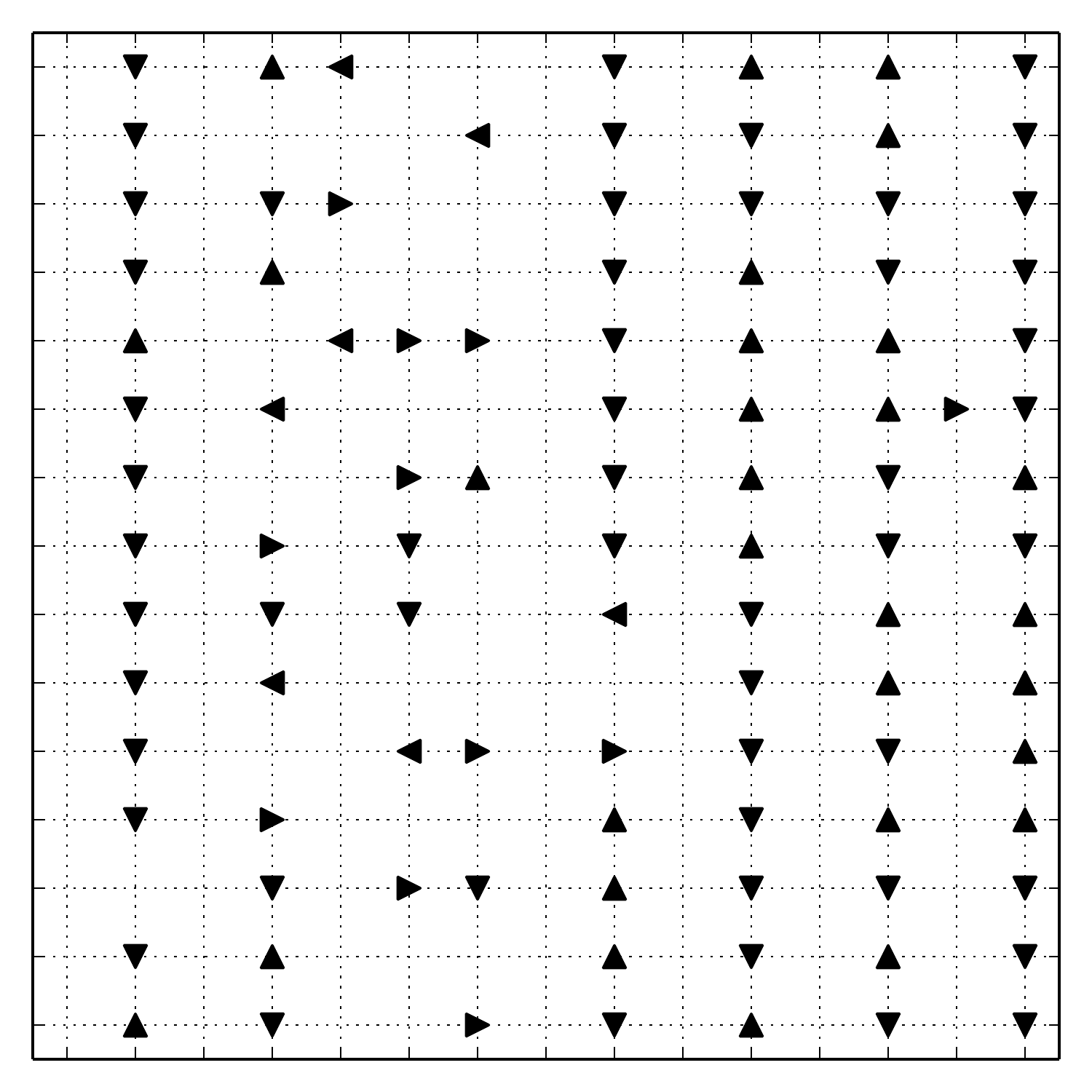}\includegraphics[width=0.25\textwidth]{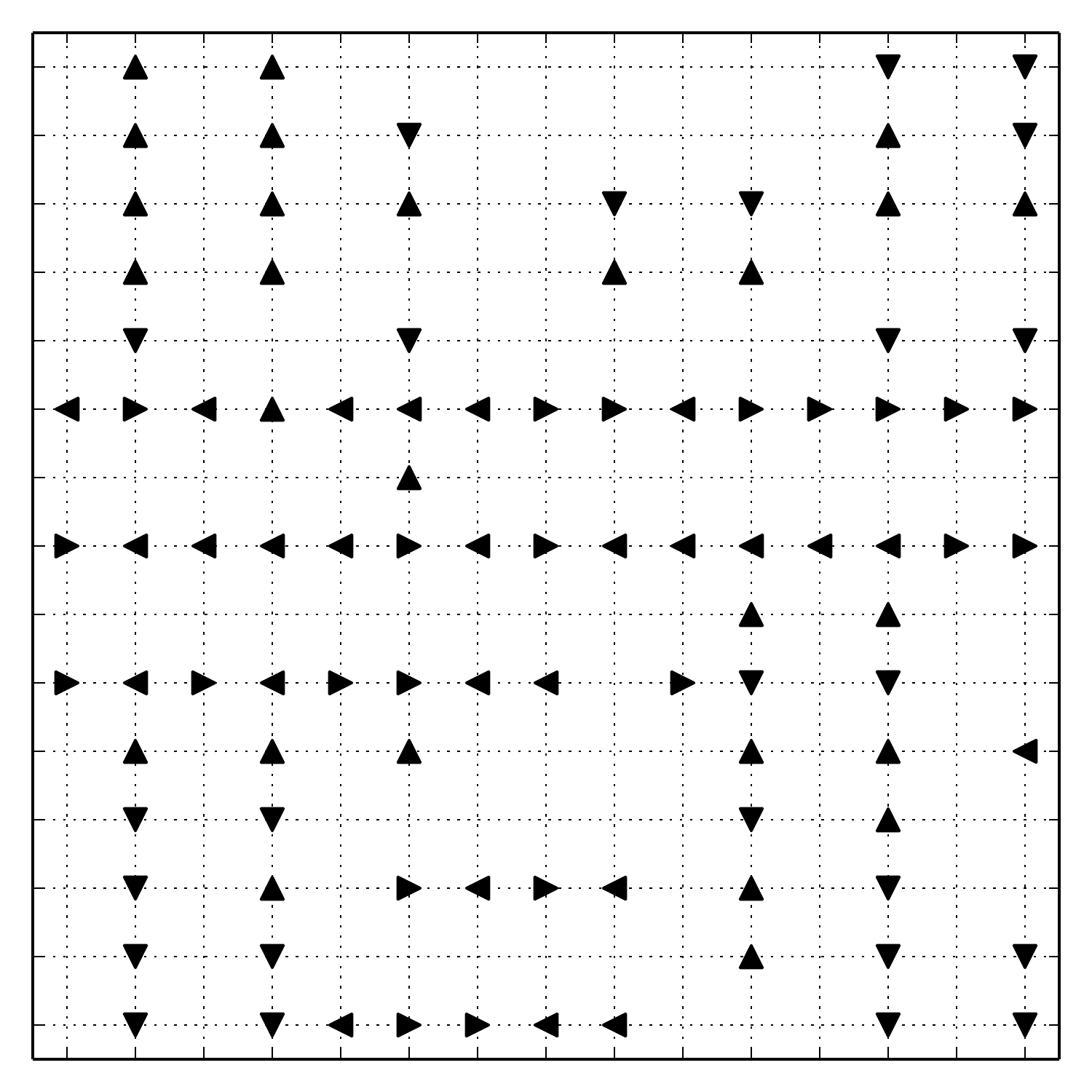}\label{fig:FIG10a}} \\
     \subfloat[grid size: $20 \times 20$] {\includegraphics[width=0.25\textwidth]{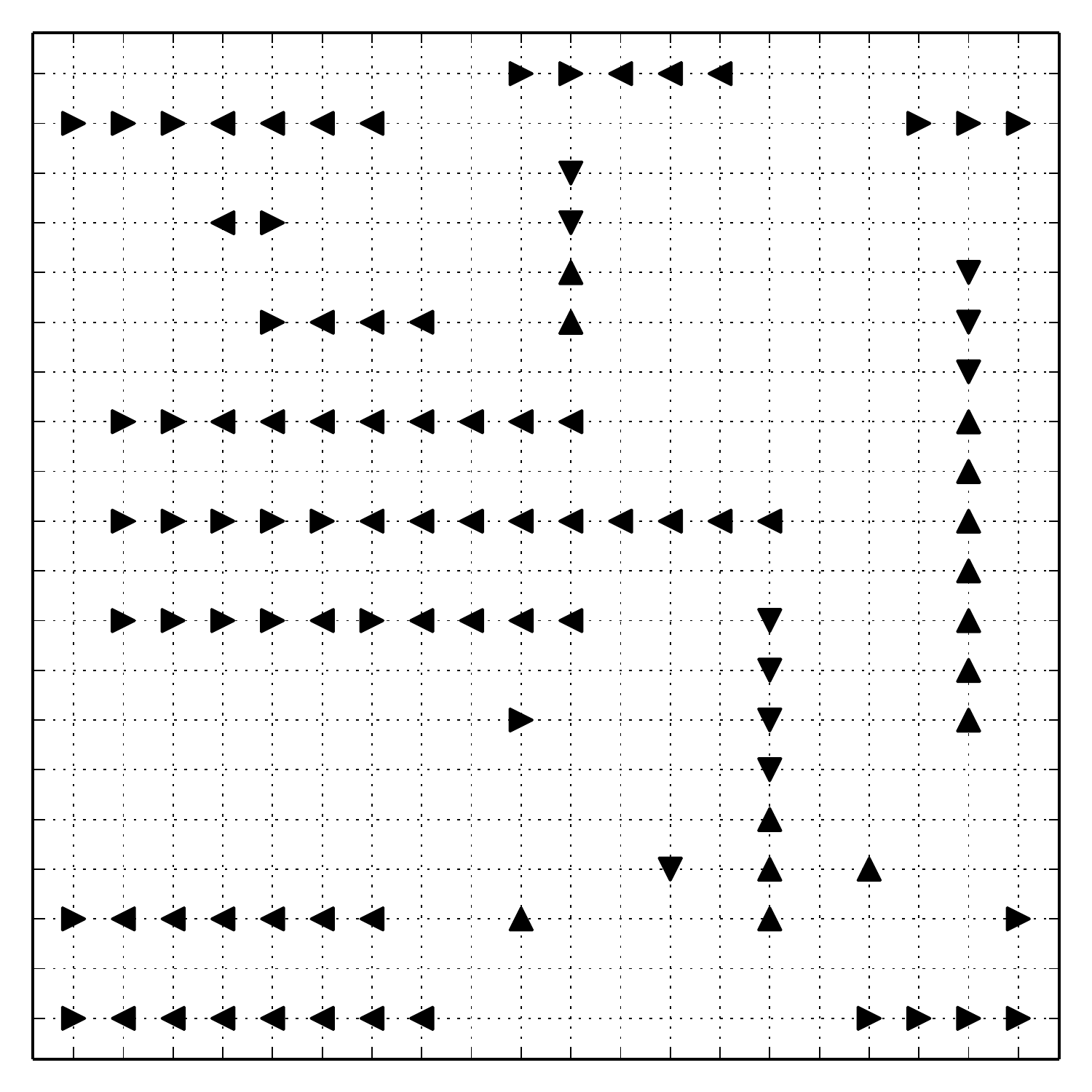}\includegraphics[width=0.25\textwidth]{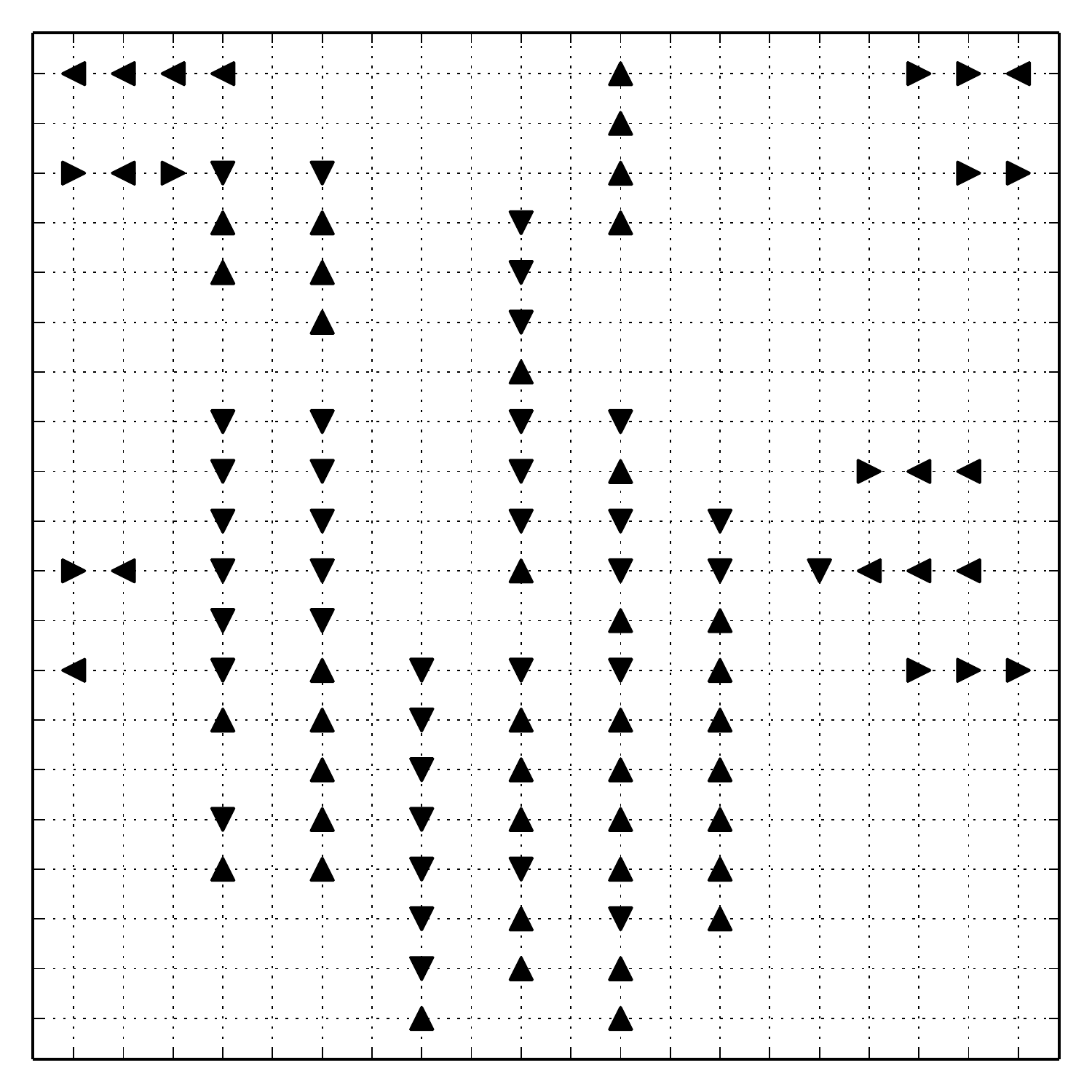}\includegraphics[width=0.25\textwidth]{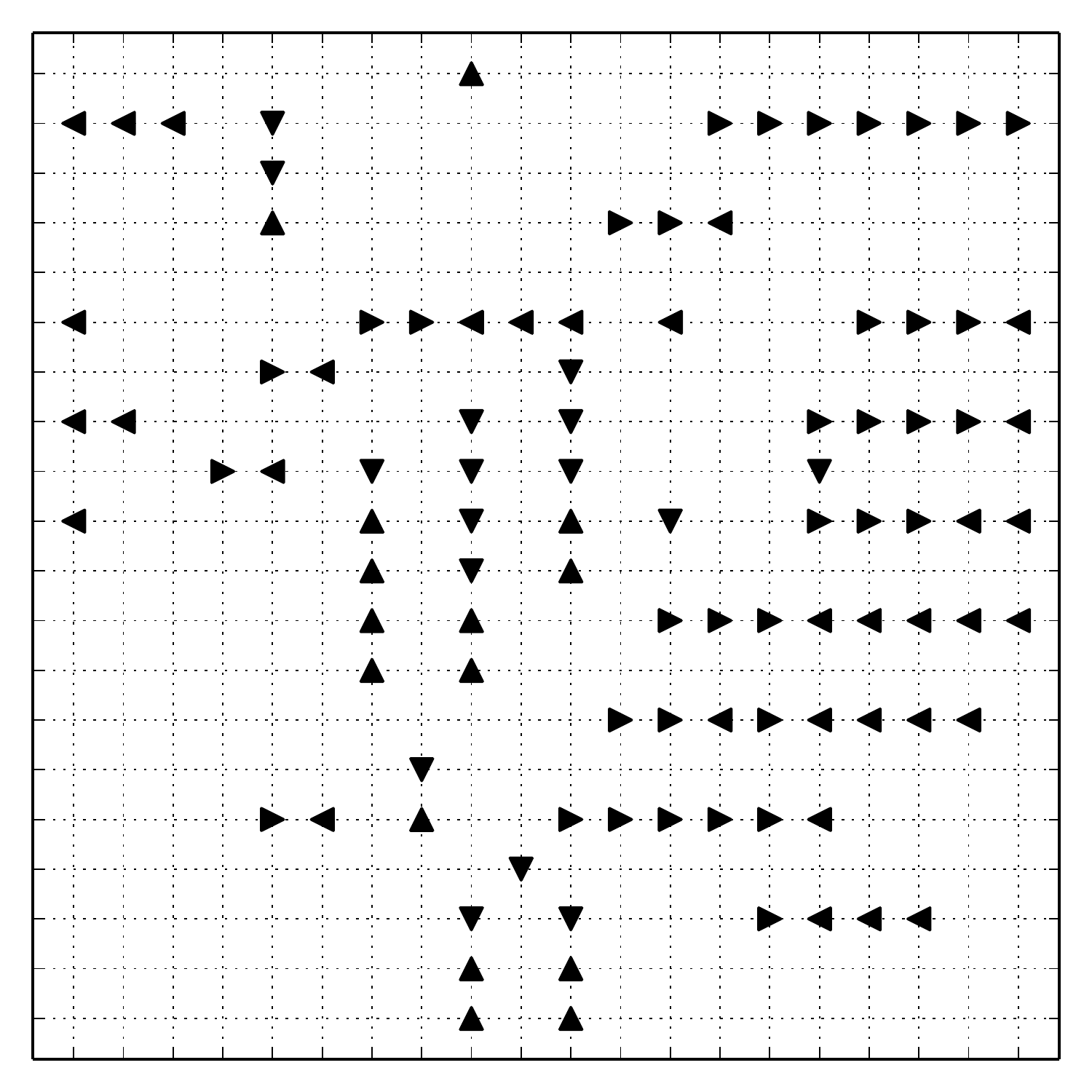}\label{fig:FIG10b}}
    \caption{Resulting structures with predefined predictions. The \textit{triangles} give the robots' headings.
    Mostly horizontal lines, mostly vertical lines, and maze-like line structures (left to right).}
    \label{fig:FIG10}
\end{figure}

We achieve line structures in all runs for both grid sizes (see Fig.~\ref{fig:FIG10}). 
A~median of $66.5\%$ of robots on the $15\times 15$ grid and a median of $89\%$ on the $20\times 20$ grid are within the structure. 
The line structures span over the whole torus on the $15 \times 15$ grid in $68\%$ of the runs (Fig.~\ref{fig:FIG10a}), while this is not observed on the $20 \times 20$ grid.  

Overall, (partially) predefining sensor predictions enables us to bias the evolutionary process, despite the intrinsic driver of minimizing surprise.

\subsubsection{Changing the Geometry of the Environment}

Predefining sensor predictions allows us to evolve a variety of line structures. 
In the next step, we aim to push the emergence towards horizontal or vertical line structures while still predefining all sensor predictions as in Sec.~\ref{sec:PRED}. 
For this purpose, we investigate the effect of the environment on the emerging structures. 
The geometry of our torus environment (i.e., ratio of one diameter to the other) can have an influence on the resulting self-assembly behaviors. 
Thus, we switch from square to rectangular grids. 
We compare the quantity of formed horizontal, vertical, and maze-like line structures in the following experiments with those formed on the $15 \times 15$ and $20 \times 20$ grids. 
Resulting line structures are categorized as mostly horizontal or mostly vertical when more than two thirds of the resulting lines are formed horizontally or vertically, respectively. 
Otherwise the resulting line structure is categorized as maze-like. 

First, we measure the quantity of emergent horizontal, vertical, and maze-like line structures in the runs with predefined predictions on the square grids.
On the $15\times 15$ grid, the best robots self-assemble in mostly horizontal lines and into mostly vertical lines in $28\%$ of the cases each and in $44\%$ in maze-like line structures.
The best robots self-assemble both into mostly horizontal lines and into mostly vertical lines in $22\%$ of the runs and form maze-like line structures in $56\%$ of the runs on the $20\times 20$ grid.
Thus, the ANN pairs evolved on the square torus grid lead to the formation of a variety of line structures.

In the next experiments, we switch to rectangular grids. We use $N=100$ robots on a $25\times 8$ grid which leads to a swarm density of $0.5$.  
The robots can either self-assemble into horizontal lines (along the longer diameter of the torus) or vertical lines (along the shorter diameter) to reach maximum fitness values in this scenario. 
The median best fitness (Eq.~\ref{equ:fitness}) of the last generation is $0.82$ in 50~independent evolutionary runs. 

\begin{figure}[tph]
    \centering
    \subfloat[mostly horizontal lines]{\includegraphics[width=0.45\textwidth]{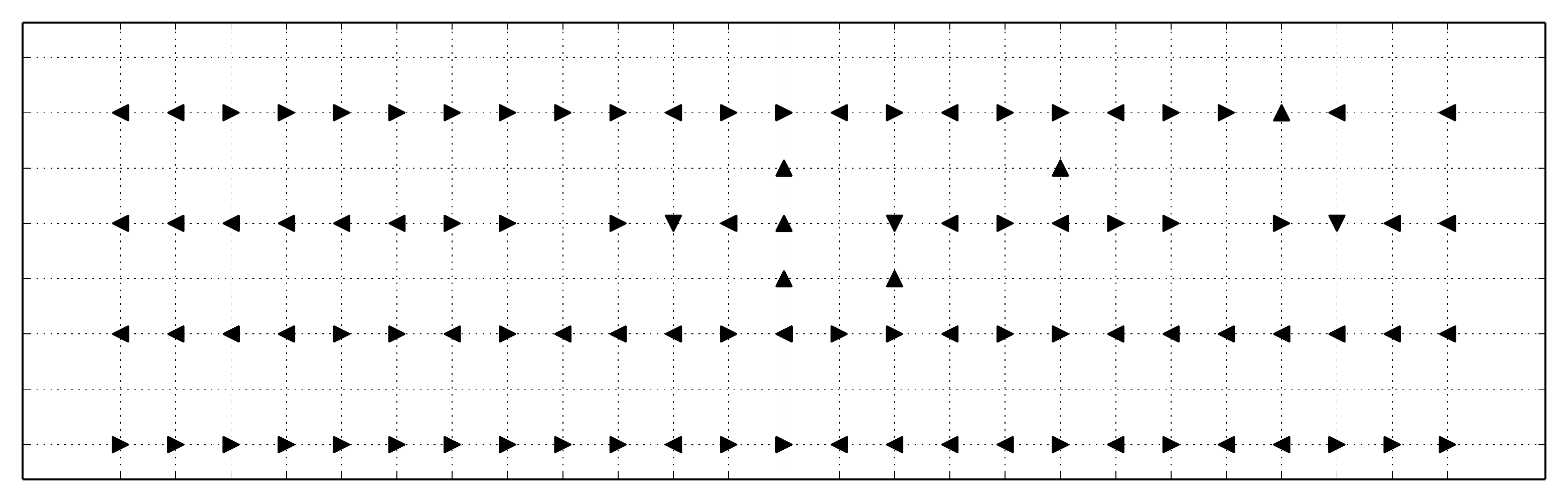}\label{fig:FIG11a}}  
    \subfloat[mostly vertical lines]{\includegraphics[width=0.45\textwidth]{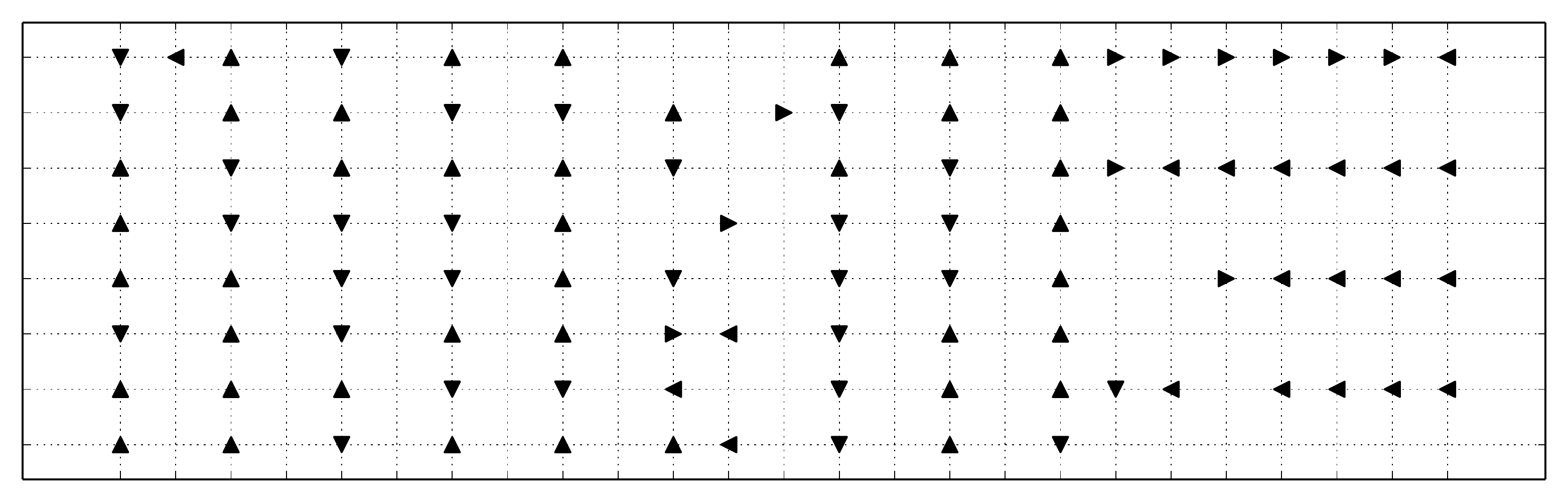}\label{fig:FIG11b}} \\
    \subfloat[maze-like line structures]{\includegraphics[width=0.45\textwidth]{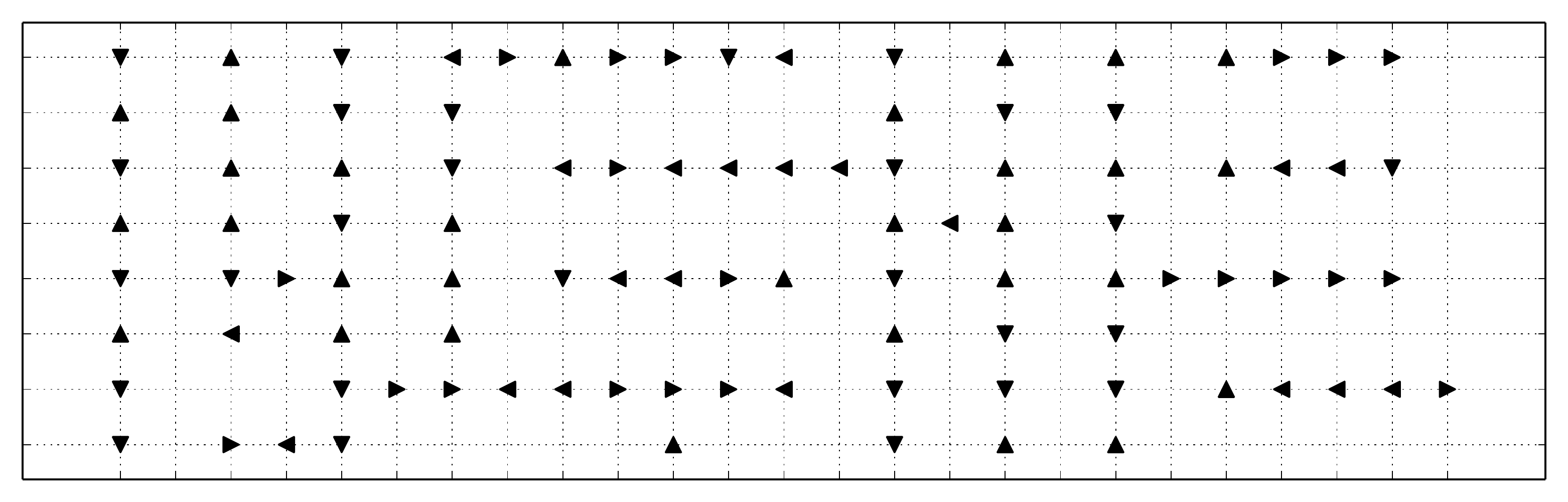}\label{fig:FIG11c}} 
    \caption{Resulting structures on a $25\times 8$ grid with predefined predictions. The \textit{triangles} give the robots' headings.}
    \label{fig:FIG11}
\end{figure}

The best robots self-assemble into mostly horizontal lines in $16\%$ of the runs (Fig.~\ref{fig:FIG11a}), in $32\%$ of the runs into mostly vertical lines (Fig.~\ref{fig:FIG11b}), and in the remaining $52\%$ into maze-like line structures (Fig.~\ref{fig:FIG11c}). 

Compared to the $15\times 15$ grid, which has a similar swarm density, we observe an increase in the formation of vertical lines and a decrease in the formation of horizontal lines on the $25\times 8$ grid.
The overall formation of horizontal and vertical lines decreases slightly from $56\%$ to $48\%$.
But the number of runs in which lines spanning the whole grid length formed increased from $68\%$ on the $15\times 15$ grid to $98\%$ on the $25\times 8$ grid. 
Maybe because one grid length is much smaller, grid-spanning lines can be formed more easily.
Nevertheless, both vertically and horizontally grid-spanning lines were formed. 

We use an $11 \times 18$ grid resulting in a swarm density of approx. $0.51$ to increase the bias towards horizontal lines.
Theoretically, the best possible fitness can be reached if the robots self-assemble into nine lines with 11~robots each and thus, one robot would be left without a proper spot. 
The median best fitness (Eq.~\ref{equ:fitness}) of the last generation is $0.81$. 

\begin{figure}[tph]
    \centering
    \subfloat[horizontal lines]{\includegraphics[width=0.25\textwidth]{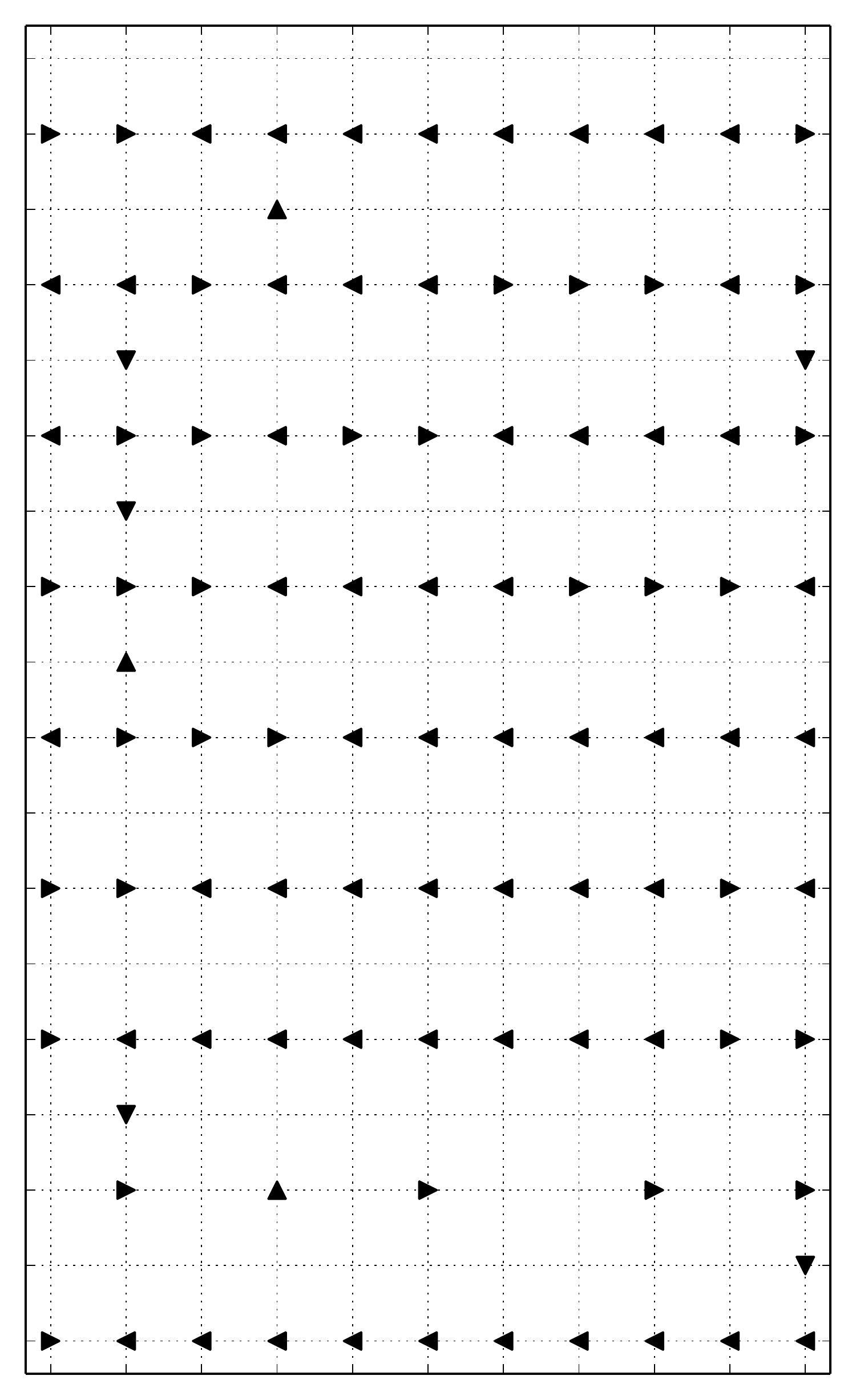}\label{fig:FIG12a}}
    \hspace{4mm}
    \subfloat[vertical lines]{\includegraphics[width=0.25\textwidth]{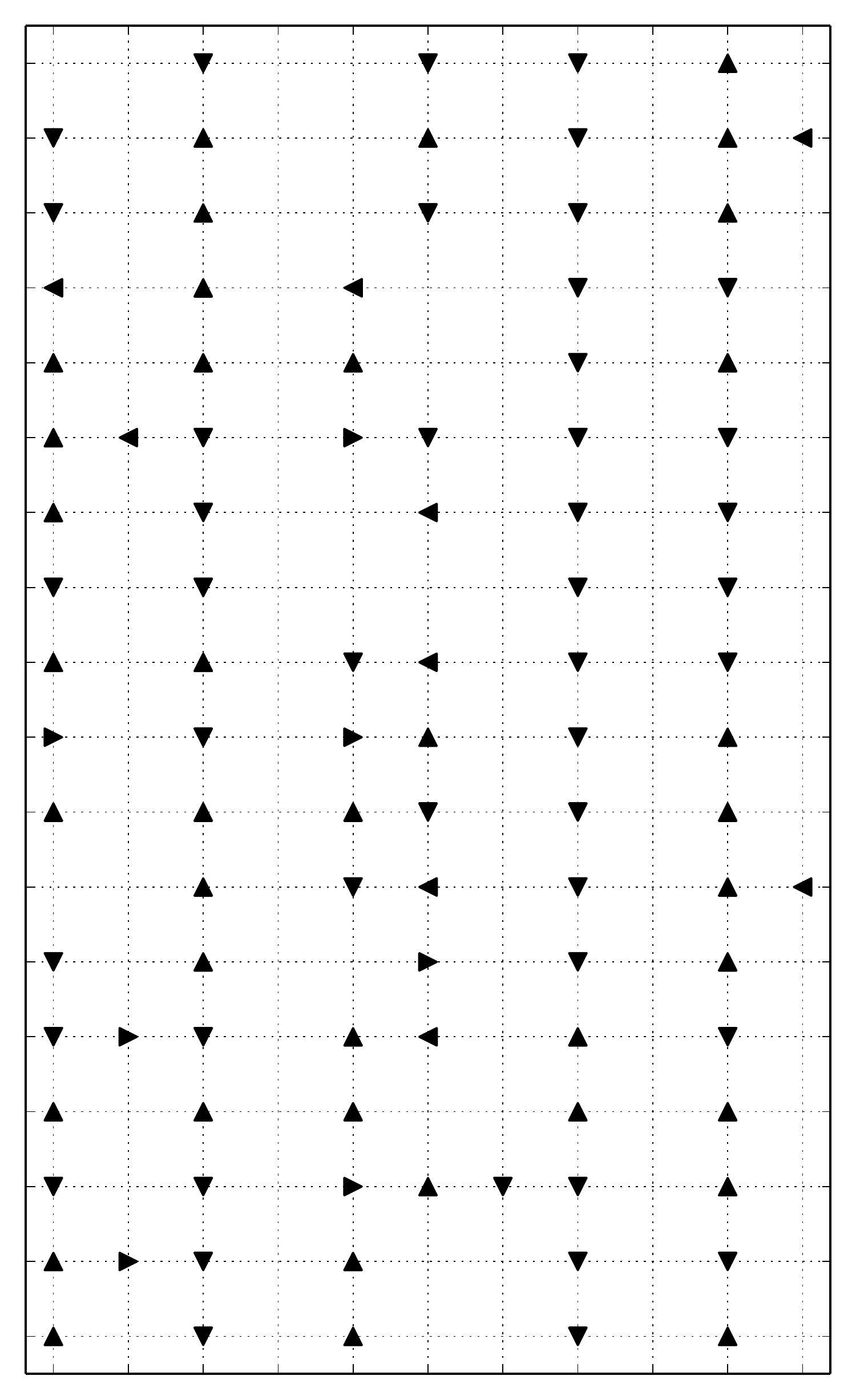}\label{fig:FIG12b}}
    \hspace{4mm}
    \subfloat[maze-like lines]{\includegraphics[width=0.25\textwidth]{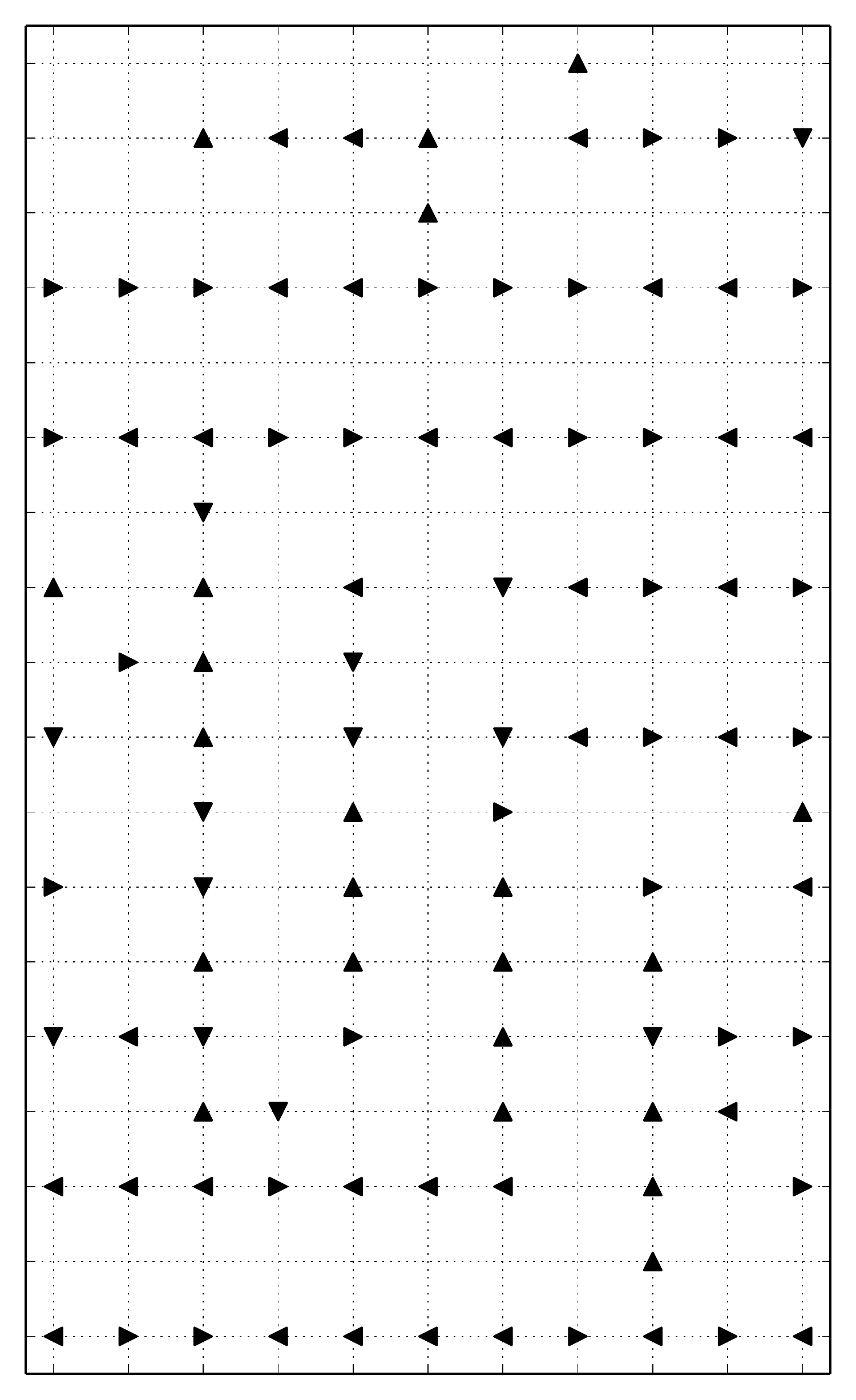}\label{fig:FIG12c}}
    \caption{Resulting structures on an $11\times 18$ grid with predefined predictions. The \textit{triangles} give the robots' headings.}
    \label{fig:FIG12}
\end{figure}

In $42\%$ of the cases, the best evolved individuals result in the emergence of mostly horizontal lines (Fig.~\ref{fig:FIG12a}), of vertical lines in $4\%$ (Fig.~\ref{fig:FIG12b}), and of maze-like line structures in $54\%$ (Fig.~\ref{fig:FIG12c}).  
We have successfully biased evolution towards horizontal lines as they emerge frequently here. 
Lines spanning the whole grid length formed in $88\%$ of the runs.
It includes all runs where mostly horizontal or vertical lines formed. 

Last, we extend our previously presented results~\cite{kaiser18} with a re-evaluation of the best individuals of the evolutionary runs with predefined sensor predictions. 
We use new random starting positions in 20~independent runs per best evolved individual, that is a total of 1000~runs per setting.
This enables us to analyze the influence of the robots' starting positions on the formation of line structures. 
An ANN pair is considered to form vertical, horizontal or maze-like lines, respectively, if more than half of these 20~evaluations with new random initial positions lead to the formation of such line structures. 
Otherwise, the ANN pair is classified as forming diverse line structures. 

\begin{table}[tph]
       \centering
        \caption{Percentage of mostly horizontal, mostly vertical and maze-like line structures when predefining all sensor predictions. Values of the 50~initial runs (1 random starting position) and of the re-evaluations (20 random starting positions/initial run, 1000 runs in total) per scenario are given.  
        \label{tab:TAB6}}
       \pgfplotstabletypeset[normal,
        columns/\space/.style={column name = , 
        column type = l},
       ]{ %
         grid size & \space & vertical & horizontal & maze-like & diverse \\ 
     15$\times$15  & initial run & 28 & 28 & 44 &  - \\ 
          & new initial positions & 2 & 14 & 14 & 70 \\ \hline 
		 20$\times$20  & initial run & 22 & 22 & 56  & -  \\ 
		 & new initial positions & 2 & 2 & 36 & 60 \\ \hline 
		 25$\times$8 & initial run & 32 & 16 & 52 & - \\ 
		 & new initial positions & 38 & 6 & 30 & 26 \\ \hline
	     11$\times$18 & initial run  & 4 & 42 & 52 & - \\
	     & new initial positions & 0 & 52 & 36 & 12 \\ 
       }
\end{table}

Both on the $15\times 15$ grid and on the $20\times 20$ grid, we observe a decrease in the quantity of ANN pairs which lead to the formation of lines in a certain orientation, i.e. vertical, horizontal or maze-like. 
$70\%$ of the evolved ANN pairs on the $15\times 15$ grid and $60\%$ on the $20\times 20$ grid lead to the formation of diverse line structures, cf. Table~\ref{tab:TAB6}.
Thus, the behaviors evolved on the square torus grid do not lead to the formation of specifically oriented line structures. 
In these runs, the initial robot positions influence the formation of the final structure. 

In contrary, on the $25\times 8$ grid the amount of ANN pairs forming vertical and horizontal lines stays similar while the quantity of genomes leading to maze-like line structures drops by $22$~pp. 
In this setting, both the formation of horizontal and of vertical lines leads to maximum fitness.
Thus, the initial position of the robots has a smaller impact on the resulting line structure. The best evolved individuals have a tendency towards vertical lines both in the initial run and in the re-evaluation.
This tendency probably arises because the formation of vertical lines is easier; they are much shorter than grid-spanning horizontal lines.
On the $11\times 18$ grid, the percentage of horizontal lines increases by $10$~pp in the evaluations with new starting positions while only $12\%$ of the ANN pairs lead to the formation of diverse line structures. 
In this setting, maximum fitness can be reached only when robots assemble into horizontal lines. 
This matches the outcome that the majority of best evolved individuals lead to the formation of horizontal lines independent of the initial robot position.
Thus, while the initial positions of the robots influences the formation of the final structure, we observe a stronger influence by the shape of the environment\footnote{Please note that this re-evaluation of the genomes with new random starting positions led to the classification of $0.7\%$ (7 out of 1000 runs) as random dispersion on the $15\times 15$ grid and to pairs in $0.1\%$ (1 out of 1000 runs) on both the $25\times 8$ grid and the $11\times 18$ grid.}.

Overall, changing the environment additionally to predefining sensor predictions enables us to bias the evolutionary process further, despite the intrinsic driver of minimizing surprise.
Thus, we can trigger the emergence of desired self-assembly behaviors by engineering self-organization.

\subsection{Resilience of Self-assembly Behaviors} 
\label{sec:Repair}

Running our approach with complete freedom (i.e., not predefining any sensor predictions), allows evolution to come up with creative solutions. 
The patterns assembled by these emergent behaviors could be vulnerable to damages. 
Thus, we extend our previously presented experiments~\cite{kaiser18} with an investigation of the resilience of the emergent self-assembly behaviors.
We damage the structure by removing or repositioning robots from an area for two of the evolved self-assembly behaviors: the behaviors forming the line structures shown in Fig.~\ref{fig:FIG7e} and the triangular lattices shown in Fig.~\ref{fig:FIG7f}. 
A~video is available online\footnote{\textit{Video3.mp4} -  \url{https://doi.org/10.5281/zenodo.3362285}}.

We evaluate both behaviors using three scenarios. 
First, we rerun the controller and world model pair using new random starting positions showing that structures form independent of initial robot positions. 
In a second step, we damage one part of the structure by removing several robots from the assembled structure (cf. Alg.~\ref{alg:remove}). 
Last, we randomly reposition all robots from a certain area of the structure (cf. Alg.~\ref{alg:reposition}). 

In each scenario, the controller and world model pair is evaluated for another 500~time steps after the damage. 
We run 20~independent runs for each scenario in the first and the last scenario, while we do one run in the second case as each repetition leads to the same result due to our completely deterministic simulation environment. 

We measure the fitness (Eq.~\ref{equ:fitness}, prediction accuracy) and the percentage of robots which are assembled into the pattern at the start and the end of the runs  realizing the above described scenarios (cf. Sec.~\ref{sec:classification}). 
In addition, we measure the similarity~S of the resulting structure in the new run~$C_\textrm{after}$ at the last time step~$T$ of the evaluation to the structure formed in the initial run~$C_\textrm{before}$ at the last time step~$T$ of the evaluation.  
$C$ contains for each occupied grid cell~$i$ the pose~$p_i\in C$ of the respective robot.
We define $S$ as

\begin{equation}
    S(C_\textrm{after}, C_\textrm{before}) = \frac{1}{N} \sum_{p_i\in C_\textrm{after}} m(p_i)
\end{equation}

where $N$ is the swarm size and with match~$m$ defined as

\begin{equation}
    m(p_i) = \left\{
\begin{array}{ll}
1 & \exists q_i\in C_\textrm{before}: p_i=q_i  \, ,\\
0 & \, \textrm{else} \\
\end{array}
\right.
\end{equation}

Only the position is considered for triangular lattices as robots constantly turn to stay on their grid cell, cf. Sec.~\ref{sec:Adaptation}. 
Thus, their headings change in every time step and are irrelevant for the formation of the structure. 
In the case of line structures, the heading of the robots is important as their parallel orientation guarantees the formation of a stable structure. 
The similarity is normalized by the swarm size~$N=100$.
With similarity~$S$ we measure the amount of preserved and reconstructed structure, that is, robots with equal positions (and headings) at the ends of the initial and the new run.

We first analyze the resilience of the line formation behavior shown in Fig.~\ref{fig:FIG7e}. 
A few robots are not part of the structure as they still move. 
The formed structure is not stable, however, a total of $82\%$ of the robots assemble into lines in this run. 

\begin{table}[tph]
       \centering
        \caption{Resilience of line structures: mean fitness (Eq.~\ref{equ:fitness}) in 500~time steps, mean percentage of robots in the structure at the start and the end of the run and mean similarity of the newly assembled structures to the structure formed in the initial run. 
        Median values in brackets.  \label{tab:TAB7}}
       \pgfplotstabletypeset[normal,
        columns/\space/.style={column name = , 
        column type = l},
        columns/Pct/.style={column name = \% of robots in structure (end), column type = F},
        columns/Pct start/.style={column name = \% of robots in structure (start), column type = F}
       ]{ %
         \space & fitness & Pct start & Pct &  similarity \\ 
         initial run & 0.814 & 0 & 82.0 &  1.0 \\ \hline
		 new starting  & 0.855 & 1.3 & 83.3 & 0.099 \\ 
		 position & (0.850) & (0.0) & (86.0) & (0.02) \\ \hline
		 remove area A & 0.958 & 69.3 & 79.5 &  0.76 \\ \hline
		 remove area B & 0.94 & 49.4 & 90.4 & 0.57  \\ \hline 
		 remove area C & 0.939 & 65.2 & 89.1 & 0.72 \\ \hline 
		reposition & 0.915 & 49.4 & 83.5 & 0.588\\
		area A & (0.917) & (54.0) & (86.5) & (0.6)\\ \hline
		reposition  & 0.905 & 32.2 & 83.9 & 0.392 \\ 
		area B & (0.913) & (36.5) & (86.0) & (0.415) \\ \hline 
		reposition & 0.903 & 56.2 & 86.2 & 0.587 \\ 
	    area C  & (0.932) & (60.0) & (87.0) & (0.585) \\
       }
\end{table}

First, we rerun the robots with the respective ANN pair using new random starting positions. 
We measure an increase in fitness (Eq.~\ref{equ:fitness}) and almost all robots form line structures, see Table~\ref{tab:TAB7}. 
The fitness of the reruns is higher as the fitness of the initial run is the minimum of ten repetitions, cf. Sec.~\ref{sec:Approach}. 
For the other scenarios, the median fitness is higher than in the initial run as well. 
In these runs, the robots have the advantage to be mostly positioned in the structure at the beginning of the run compared to random starting positions.
Thus, the predictions of the robots' world model, cf.~\ref{fig:FIG9c}, matches more closely from the start of the run and a higher fitness can be reached.

\begin{figure}[tph]
    \centering
    \subfloat[area A]{\includegraphics[width=0.25\textwidth]{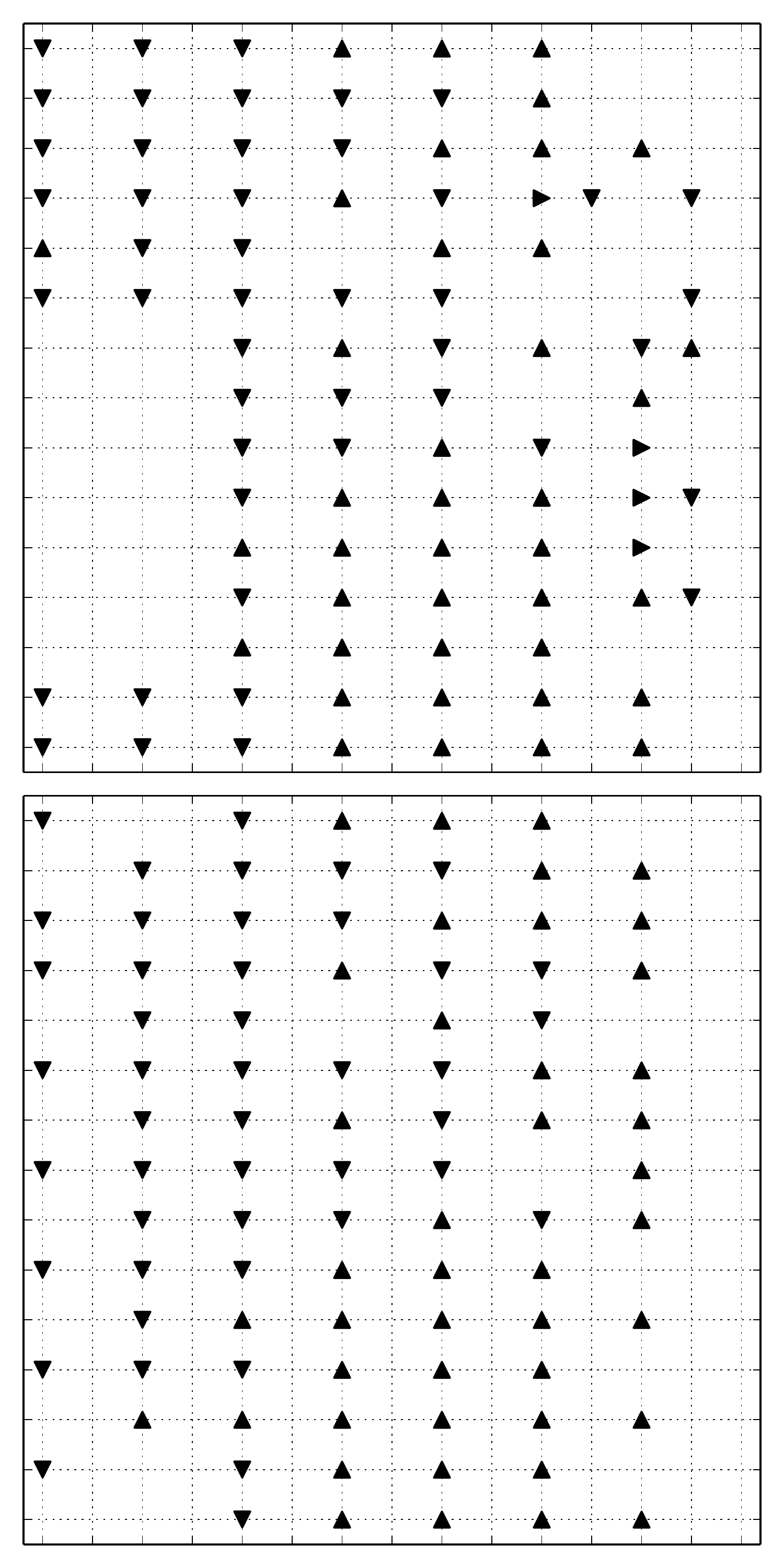}\label{fig:FIG13a}}
    \hspace{3mm}
    \subfloat[area B]{\includegraphics[width=0.25\textwidth]{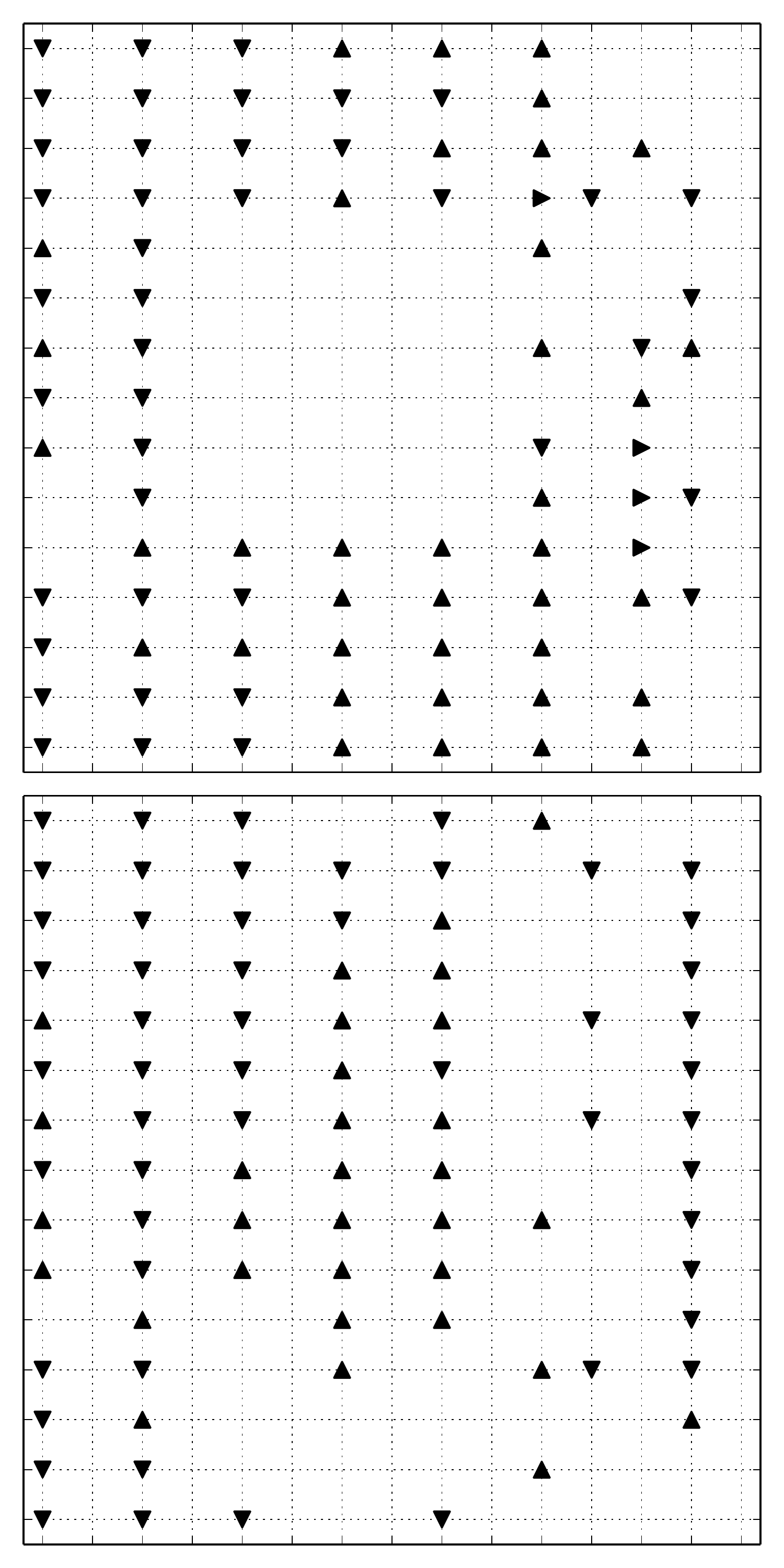}\label{fig:FIG13b}}
    \hspace{3mm}
    \subfloat[area C]{\includegraphics[width=0.25\textwidth]{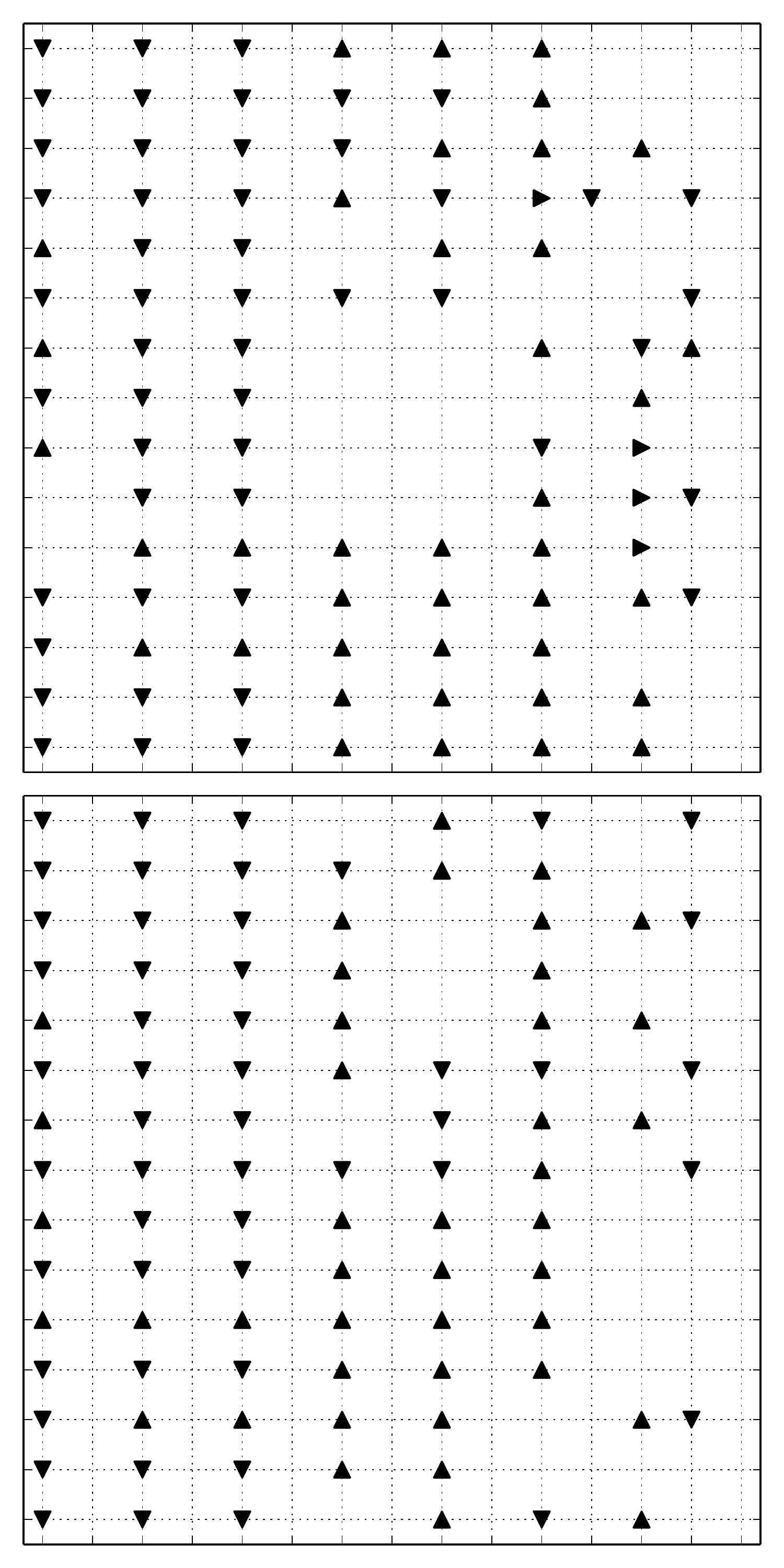}\label{fig:FIG13c}} 
    \caption{Removal of robots from different areas in line structures. 
    Starting positions top, final positions after running the simulation for another 500~time steps bottom.}
    \label{fig:FIG13}
\end{figure}

Next, we completely remove robots from three different areas of the structure, see Fig.~\ref{fig:FIG13}. We remove 12~(see Fig.~\ref{fig:FIG13a}), 17~(see Fig.~\ref{fig:FIG13b}) and eight (see Fig.~\ref{fig:FIG13c}) robots, respectively. 
In all three cases, the percentage of robots within the structure drops due to the removal of robots by at least~$13$~pp.
We run the controller for another 500~time steps and observe that the percentage of robots within the structure increases again.
Even higher levels than in the initial structure are reached in two out of three cases. 
Furthermore, line structures form in the removed area again, see Fig.~\ref{fig:FIG13}, and we find a high resemblance to the original structure in two out of three runs.

Instead of removing robots from the arena, we randomly reposition them outside of the three different areas but within the arena now.
The swarm density stays constant and robots have to coordinate with more other robots than in the tests above. 
The percentage of robots in the structure reduces by at least $25$~pp by randomly repositioning them, cf. Table~\ref{tab:TAB7}. 
In all three scenarios, the percentage of robots within the structure increases again and is in the last time step at least $1.5$~pp higher than in the initial structure.  
However, the similarity to the initial structure does not reach a higher median value than $60\%$. 

\begin{table}[t]
       \centering
       \caption{Resilience of triangular lattices: mean fitness (Eq.~\ref{equ:fitness}) in 500~time steps, mean percentage of robots in the structure at the start and the end of the run and mean  similarity of the newly assembled structure to the structure formed in the initial run. 
        Median values in brackets. \label{tab:TAB8}}
       \pgfplotstabletypeset[normal,
        columns/\space/.style={column name = , 
        column type = l},
         columns/Pct/.style={column name = \% of robots in structure (end), column type = F},
        columns/Pct start/.style={column name = \% of robots in structure (start), column type = F}
       ]{ %
         \space & fitness & Pct start & Pct &  similarity \\ 
        initial run & 0.764 & 0 & 63.0 & 1.0 \\ \hline
		 new starting & 0.78 & 0.0 & 77.0 &  0.468 \\ 
		 position & (0.781) & (0.0) & (78.0) & (0.46) \\ \hline
		remove area A & 0.763 & 60.9 & 58.6 & 0.39 \\ \hline 
		remove area B & 0.761 & 67.1 & 65.9 & 0.52  \\ \hline 
		remove area C & 0.791 & 66.3 & 76.1 & 0.67 \\ \hline 
		reposition  & 0.789 & 21.8 & 75.2 &  0.564\\ 
		area A & (0.788) & (22.5) &(80.5) & (0.62) \\ \hline
		reposition & 0.791 & 16.4 & 78.2 & 0.545   \\
		area B & (0.791) & (18.0) & (80.0) & (0.605) \\ \hline 
		reposition & 0.798 & 38.4 & 79.3 & 0.571 \\
		area C & (0.798) & (39.5) & (80.5) & (0.61) \\ 
       }
\end{table}

As a second example, we show the resilience of the behavior leading to the formation of triangular lattices as illustrated in Fig.~\ref{fig:FIG7f}. 
As before, we observe an increase in the fitness rerunning the controller with new initial random starting positions, see Table~\ref{tab:TAB8}. 
Furthermore, a high percentage of robots is part of the structure. 

\begin{figure}[t]
    \centering
    \subfloat[area A]{\includegraphics[width=0.25\textwidth]{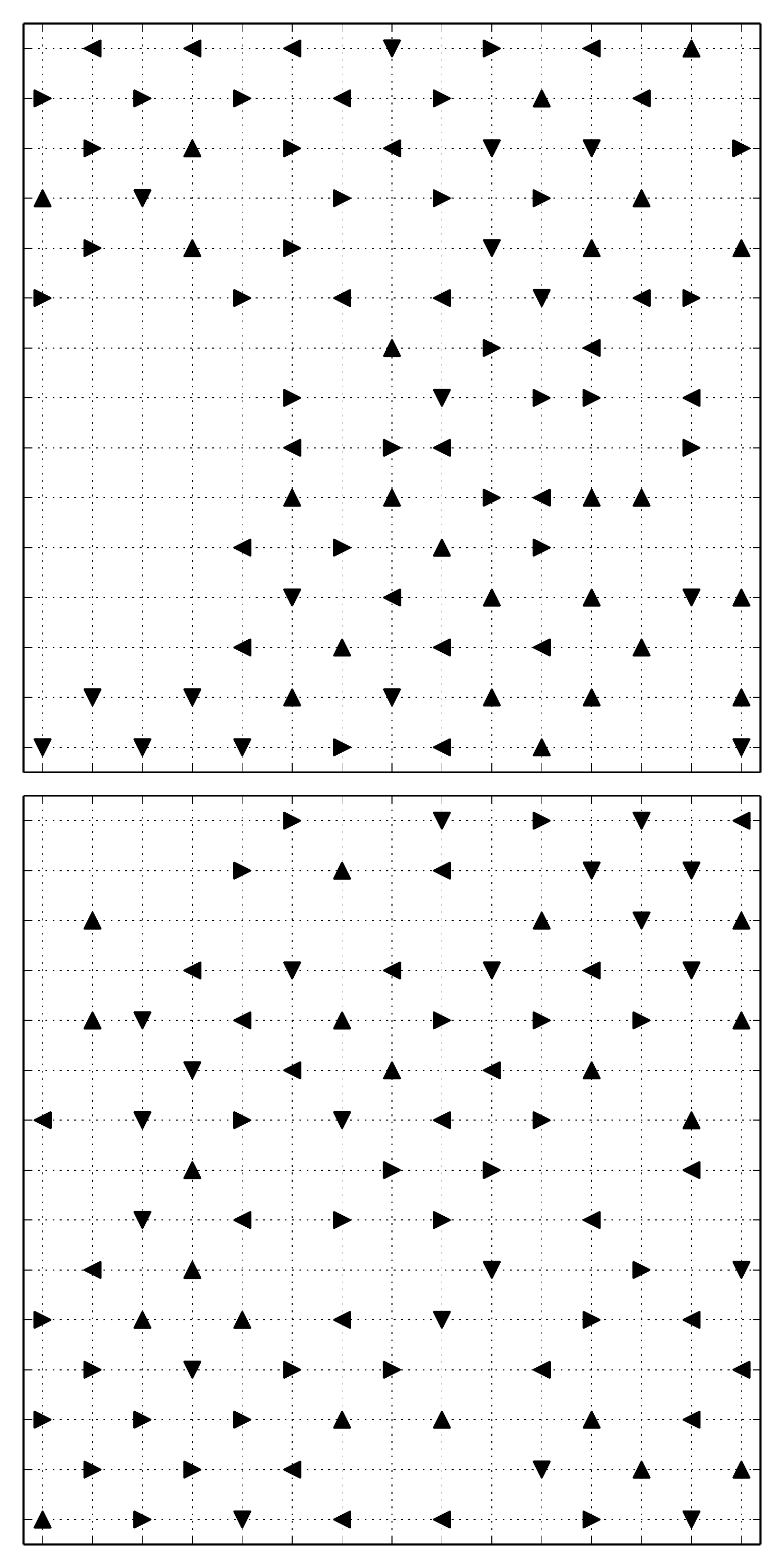}\label{fig:FIG14a}}
    \hspace{3mm}
    \subfloat[area B]{\includegraphics[width=0.25\textwidth]{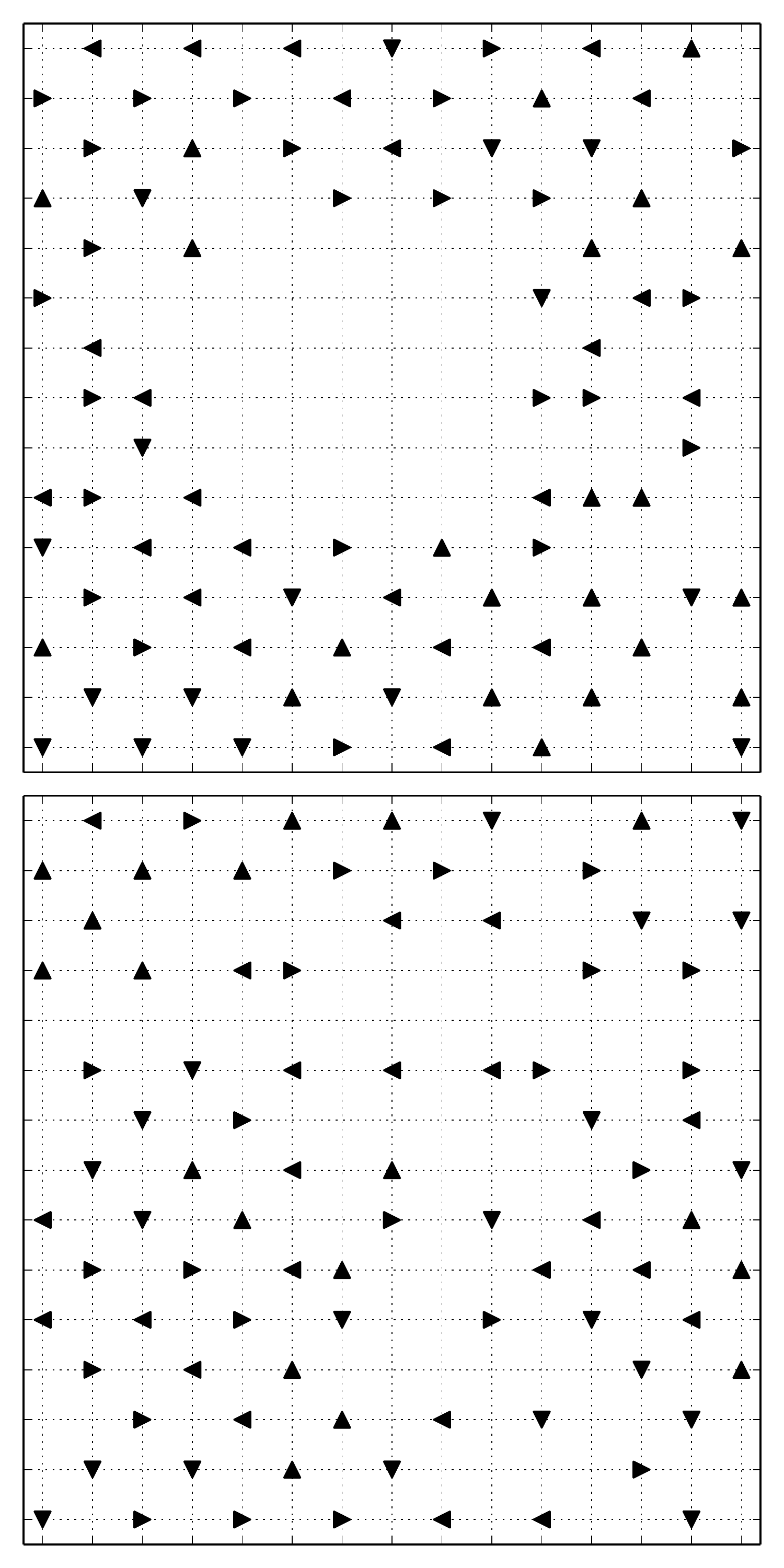}\label{fig:FIG14b}}
    \hspace{3mm}
    \subfloat[area C]{\includegraphics[width=0.25\textwidth]{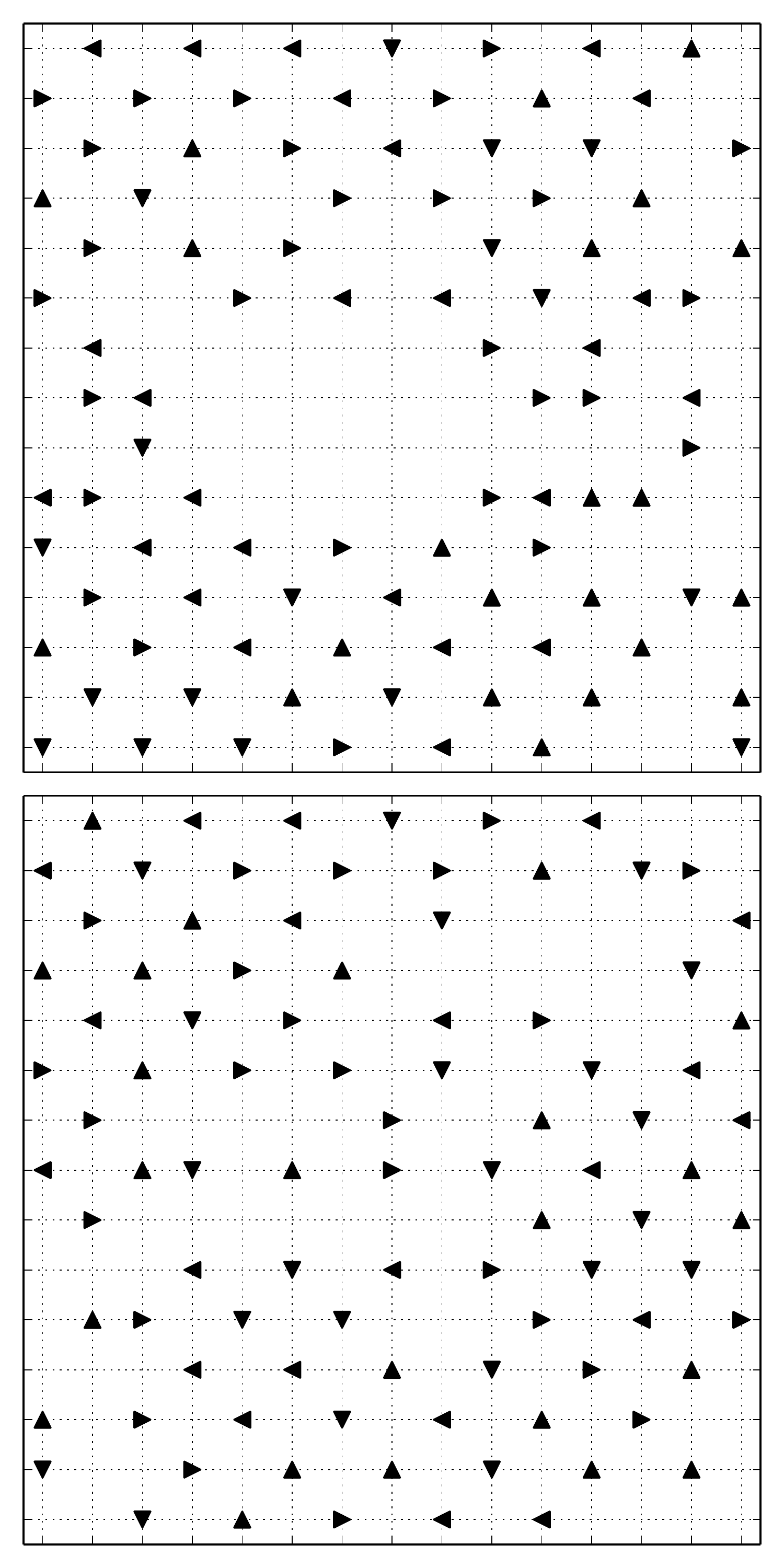}\label{fig:FIG14c}}
    \caption{Removal of robots from different areas in a triangular lattice. Starting positions top, final positions after running the simulation for another 500~time steps bottom.
    }
    \label{fig:FIG14}
\end{figure}

We remove three different areas from the structure, cf. Fig.~\ref{fig:FIG14}. 
Thus, 13 (see Fig.~\ref{fig:FIG14a}), 15 (see Fig.~\ref{fig:FIG14b}) and eight (see Fig.~\ref{fig:FIG14c}) robots, respectively, are removed from the structure. 
This leads to an immediate increase of the percentage of robots within the structure for two out of three scenarios, see Table~\ref{tab:TAB8}. 
At the end of the initial run, $63\%$ of robots are positioned within the structure. 
Then we remove robots from the structure, which reduces the total number of robots, and hence the percentage of robots being positioned in the structure can change. 
When we remove robots from areas~B and~C, we get higher percentages at the start of these new runs than at the end of the initial run: $66.3\%$ and $67.1\%$, respectively.
This indicates that robots which were not part of the structure got removed. 
In one case, this percentage increases after running the controller for another 500~time steps while in the other two cases it even decreases. 
On a $15\times 15$ grid, 112~robots are required to form a repetitive triangular lattice pattern over the whole grid. 
Our original swarm size of $N=100$ is lower, but still close to the ideal number of robots. 
Reducing the number slightly makes the formation of a triangular lattice harder, but it is still possible. 
Removing a larger amount of robots, in our example~13 or~15, leads to a too low swarm density that may prevent the formation of a repetitive pattern. 

We reposition the robots from the three areas in the next step. 
It leads to a major decrease of robots being part of the triangular lattice. 
In all three scenarios, we observe an increase of the number of robots within the structure by at least $12$~pp compared to the initial structure and an intermediate similarity to the initial structure. 
Thus, the robots reassemble to form the same pattern as in the initial run, but do not position themselves on the same grid cells. 

In total, we observe that the evolved self-assembly behaviors are resilient as removing or repositioning robots leads to the reassembly of the pattern. 

\subsection{Sensor Noise} \label{sec:noise} 

Finally, we study the influence of noise on the system. 
By adding sensor noise, we make a first step towards more realistic environments as experienced by real-world robots.
In 20~runs per scenario, we flip each binary sensor value with a probability of $5\%$, $10\%$ or~$15\%$. 
Now, sensors have a non-deterministic element that may impact on what behaviors emerge. 
We run the approach with complete freedom, that is, without predefining any sensor predictions.
Again, we do experiments on a $15\times 15$ grid and on a $20\times 20$ grid.
We compare the noisy runs with runs that have deterministic sensors.
For this purpose, we include the data for sensor model~C (20~runs per scenario) from Sec.~\ref{sec:sensormodel} in Tabs.~\ref{tab:TAB9} and~\ref{tab:TAB10} as the $0\%$~noise case.

\begin{table}[tph]
       \centering
        \caption{Median best fitness (Eq.~\ref{equ:fitness}) of the last generation and percentage of resulting structures on a $15 \times 15$ grid for different noise levels. 20~runs each. }
        \label{tab:TAB9}
       \pgfplotstabletypeset[normal,
        columns/\space/.style={column name = , 
        column type = l},
        columns/sensor noise/.style = { column type = C }, 
        columns/aggre- gation/.style = { column type = C }, 
        columns/loose grouping/.style = { column type = C }, 
        columns/triang. lattice/.style = {column type = C },
        columns/loose disp./.style = {column type = C },
       ]{ %
         sensor noise & fitness & lines & aggre- gation & clustering & loose grouping & triang. lattice \\ 
         0\% & 0.698 & 15 & 10 & 55 & 15 & 5  \\  
         5\% & 0.724 & 10 & 55 & 30 & 5 & 0  \\  
         10\% & 0.694 & 5 & 40 & 30 & 15 & 10  \\  
         15\% & 0.654 & 0 & 50 & 30 & 15 & 5  \\  
       }
        
        \bigskip
        
       \caption{Median best fitness (Eq.~\ref{equ:fitness}) of the last generation and percentage of resulting structures on a $20\times 20$ grid for different noise levels. 20~runs each.}
       \label{tab:TAB10}
       \pgfplotstabletypeset[normal,
        columns/\space/.style={column name = , 
        column type = l},
        columns/sensor noise/.style = { column type = C }, 
        columns/aggre- gation/.style = { column type = C }, 
        columns/loosely grouped/.style = { column type = C }, 
        columns/triang. lattice/.style = {column type = C },
        columns/random dispersion/.style = {column type = E },
       ]{ %
         sensor noise & fitness & pairs & lines & clustering & random dispersion \\ 
         0\% & 0.805 & 27.5 & 22.5 & 10 & 40  \\ 
         5\% & 0.768 & 0 & 15 & 5 & 80  \\ 
         10\% & 0.725 &  0 & 0 & 15 & 85  \\ 
         15\% & 0.699 & 0 & 0 & 0 & 100  \\ 
       }
\end{table} 

On the smaller grid, the median best fitness (Eq.~\ref{equ:fitness}) of the last generation stays on a similar level of about~$70\%$ for noise levels of up to~$10\%$. 
It decreases slightly when adding~$15\%$ noise, see Tab.~\ref{tab:TAB9}. 
The median best fitness on the larger grid decreases with increasing sensor noise, see Tab.~\ref{tab:TAB10}.
The task complexity grows with increasing non-determinism because the noise is inherently unpredictable.

On both grid sizes, the emergence of more easily predictable structures increases when sensor readings are subject to noise. 
On the smaller grid (higher swarm density) grouping behaviors are favored. 
In these structures most robots are stuck at their grid cells due to gridlock. 
This prevents a quick disassembly of structures due to wrongly selected actions based on inaccurate sensor readings. 
The number of robots unaffected by gridlock rises with the size of the structure's border. 
Aggregates are the most robust structures because they have the smallest border. 
On the larger grid, random dispersion behaviors prevail.
Inaccurate sensor readings would either lead to the avoidance of false positive detections of neighbors or to a minor decrease in prediction accuracy (fitness). 
Hence, aggregates and random dispersion behaviors can easily form in non-deterministic environments, too. 

Other behaviors rely on correct positioning and sensing to maintain the formed structure.
Pairs do not emerge when introducing sensor noise and lines form less frequently with increasing noise levels. 
Robots in lines and pairs may not detect the robot in front for a moment due to noise and turn.
Then the structure may disperse because the robots lost their alignment (cf. Sec.~\ref{sec:classification}).
Forming and maintaining pairs and lines gets harder with increasing noise.

As before, triangular lattices emerge rarely and only on the smaller grid independently of sensor noise. 
The measured differences between different levels of noise may be due to variance in the experiments.

Clusters form rarely on the larger grid and for noise levels of up to $10\%$. 
We expect to observe clusters also for $15\%$ sensor noise with an increased number of experiments. 

Overall, the formation of a variety of structures gets harder with increasing noise levels 
while various patterns are still observed on both grids for sensor noise up to $10\%$.
Based on these results, we are confident that the switch to real-world settings is challenging but possible. 

\section{Discussion and Conclusion} 
\label{sec:Discussion}

In our study, we observe different self-assembly behaviors when running our approach without predefined predictions.
The emergent behaviors depend on the swarm density. 
For high densities, we find mostly grouping behaviors (aggregation, clustering, etc.) and for intermediate densities a wide variety of behaviors (lines, dispersion, triangular lattices, etc.). 
Thus, too high and too low swarm densities generate rather trivial collective behaviors which is in line with our previous findings~\cite{hamann14d}. 
The predictions of the robots are complicated by densities in between, which may lead to the emergence of more complex behaviors to still achieve ordered and easy to predict patterns.

The (partially) engineered approach of predefining the prediction network's outputs successfully biases towards desired self-organizing collective behaviors. 
We find different collective self-assembly behaviors based on the density and/or the shape of the environment (here, the relation of the short radius to the long radius of the torus), that is, an adaptation to the environment.  
Furthermore, we find that the evolved self-assembly behaviors are resilient and lead to the reassembly of the pattern when removing or replacing robots. 
Introducing sensor noise complicates pattern formation, but we still find a variety of structures for low noise levels.

In future work, we will analyze the resulting networks in more detail, investigate the influence of heterogeneous swarms on the emerging behaviors, and explore how self-assembly behaviors can be evolved that lead to the formation of more complex patterns while minimizing surprise.
We plan to study non-deterministic environments by adding noise to action execution. The swarm will probably react by forming structures that are depending less on precision while being more robust.

We are confident to switch our grid-based self-assembly simulation to continuous space. 
Dispersion and aggregation behaviors already emerged in previous work applying minimal surprise in a continuous environment~\cite{borkowski17}.
In our experiments, several of the formed structures rely on gridlock. 
It can be exploited for grouping behaviors, such as aggregation, in continuous environments as well because robots can get trapped within a bigger cluster (e.g., as observed for the BEECLUST algorithm~\cite{schmickl_BEECLUST_2010}). 
Other patterns forming on the grid, such as lines or pairs, are harder to accomplish in continuous settings. 
However, variants of the flocking behaviors evolved in a 2D continuous environment in our previous work~\cite{borkowski17} already resemble to moving lines. 
Furthermore, the emergence of lines and pairs can be fostered by the robot shape or hardware. 
Andr\'{e}en et al.~\cite{Andreen2016} shaped their robots such that they can loosely attach to each other, which led to the formation of chains and lines.
Similarly, robots that can physically attach to each other enable the formation of stable self-assemblages and basically discretize the environment (e.g., Swarmbot~\cite{mondada_2004_swarm-bot} and modular robotic systems as used in the SYMBRION and REPLICATOR projects~\cite{levi10}). 
Thus, we are confident to tackle the challenge of transferring our results to real-world settings. 
Evolved self-assembly behaviors can then enable robots to cover and monitor large areas (dispersion in regular patterns), to cross gaps which are too wide for an individual robot or to push heavy objects (lines and pairs).  
In future, we might even be able to evolve more complex assembly behaviors which lead to the formation of objects or tools, that is, the swarm may act as programmable matter.  

In ongoing work, we conduct first experiments to prepare our approach for swarm robot experiments.
Instead of the evolutionary approach, we make use of machine learning techniques (backpropagation). 
The prediction network of a robot can be trained with self-supervised learning using a robot's own predictions and sensor readings~\cite{ha_2018, nolfi_learning_1994}. 
Thus, the approach can be used without the need to label training data or to specify a fitness function.
Additionally, we switch from an action ANN to a simple rule set. 
In initial experiments using a single Thymio~II robot~\cite{riedo13}, we were able to train the prediction network and a wall following behavior emerged. 
This is a promising first step to show that our approach is suitable for real-world scenarios - also because the sensors of the Thymio are continuous. 
Furthermore, we showed in Sec.~\ref{sec:noise} that a variety of behaviors emerges also under a limited amount of sensor noise.
In future work, we aim to learn more complex behaviors on a single robot as well as in robot swarms such that complex multi-robot behaviors can arise on top of a mere self-supervised learning process driven by the purely intrinsic motivation to minimize surprise.   

\renewcommand{\thesection}{A}
\section{Appendix}

In Sec.~\ref{sec:Repair}, robots are removed or replaced in an area of the grid. Here, we provide a formal definition in form of algorithms.

\begin{algorithm}
\caption{Remove Area} \label{alg:remove} 
\begin{algorithmic} 
\STATE{rectangular area defined by ($x_{min}$, $y_{min}$) and ($x_{max}$, $y_{max}$)}
\STATE{set r of all robots with positions ($x$, $y$)}

\FOR{each robot in r} 
    \IF{robot position $x_{max} \geq x \geq x_{min}$ and $y_{max} \geq y \geq y_{min}$}
       \STATE{remove robot from r} 
      \ENDIF 
      \ENDFOR 
\end{algorithmic}
\end{algorithm}

\begin{algorithm}
\caption{Reposition Area} \label{alg:reposition} 
\begin{algorithmic} 
\STATE{rectangular area defined by ($x_{min}$, $y_{min}$) and ($x_{max}$, $y_{max}$)}
\STATE{set r of all robots with positions ($x$, $y$)}

\FOR{each robot in r} 
    \WHILE{robot position $x_{max} \geq x \geq x_{min}$ and $y_{max} \geq y \geq y_{min}$}
      \STATE{generate new random position p ($x$, $y$)}
       \STATE{reposition robot to p} 
      \ENDWHILE
      \ENDFOR 
\end{algorithmic}
\end{algorithm}

\end{document}